\documentclass{article}

\usepackage{microtype}
\usepackage{graphicx}
\usepackage{subfigure}
\usepackage{booktabs} % for professional tables

\usepackage{amsmath,amsfonts,bm}

\def\eqref#1{equation~\ref{#1}}
\def\1{\bm{1}}

\DeclareMathAlphabet{\mathsfit}{\encodingdefault}{\sfdefault}{m}{sl}
\SetMathAlphabet{\mathsfit}{bold}{\encodingdefault}{\sfdefault}{bx}{n}

\DeclareMathOperator*{\argmax}{arg\,max}
\DeclareMathOperator*{\argmin}{arg\,min}

\usepackage{url}
\usepackage{booktabs}
\usepackage{pdflscape} % allow landscape
\usepackage{bm, bbm}
\usepackage{amsmath,amssymb,amsthm}
\usepackage{caption}

\newcommand{\x}{\bm{x}}

\newcommand{\z}{\bm{z}}
\newcommand{\y}{\bm{y}}

\newcommand{\data}{\mathcal{D}}

\usepackage{hyperref}

\usepackage[accepted]{icml2019}

\icmltitlerunning{Are Generative Classifiers More Robust to Adversarial Attacks?}

\begin{document}

\twocolumn[
\icmltitle{Are Generative Classifiers More Robust to Adversarial Attacks?}



\icmlsetsymbol{equal}{*}

\begin{icmlauthorlist}
\icmlauthor{Yingzhen Li}{msr}
\icmlauthor{John Bradshaw}{cam,mpi}
\icmlauthor{Yash Sharma}{tue}
\end{icmlauthorlist}

\icmlaffiliation{msr}{Microsoft Research Cambridge, UK}
\icmlaffiliation{cam}{University of Cambridge, UK}
\icmlaffiliation{mpi}{Max Planck Institute for Intelligent Systems, Germany}
\icmlaffiliation{tue}{Eberhard Karls University of T\"ubingen, Germany}

\icmlcorrespondingauthor{Yingzhen Li}{Yingzhen.Li@microsoft.com}

\icmlkeywords{Machine Learning, ICML}

\vskip 0.3in
]



\printAffiliationsAndNotice{}  % leave blank if no need to mention equal contribution

\begin{abstract}
There is a rising interest in studying the robustness of deep neural network classifiers against adversaries, with both advanced attack and defence techniques being actively developed. However, most recent work focuses on \emph{discriminative} classifiers, which only model the conditional distribution of the labels given the inputs. In this paper, we propose and investigate the \emph{deep Bayes} classifier, which improves classical naive Bayes with conditional deep generative models. We further develop detection methods for adversarial examples, which reject inputs with low likelihood under the generative model. Experimental results suggest that deep Bayes classifiers are more robust than deep discriminative classifiers, and that the proposed detection methods are effective against many recently proposed attacks.
\end{abstract}

\section{Introduction}
\label{sec:intro}

Deep neural networks have been shown to be vulnerable to adversarial examples \citep{szegedy:intriguing2013, goodfellow:explaining2014}. The latest attack techniques can easily fool a deep net with imperceptible perturbations \citep{goodfellow:explaining2014, papernot:limitations2016, carlini:attack2017,kurakin:adversarial2016, madry:towards2018, chen:ead2017}, even in the black-box case, where the attacker does not have access to the network's weights \citep{papernot:practical2017, chen2017zoo, brendel2017decision, athalye:obfuscated2018, alzantot2018genattack, uesato:spsa2018}.
Adversarial attacks are serious security threats to machine learning systems, threatening applications beyond image classification \citep{carlini:audio2018, alzantot2018nlp}.

To address this outstanding security issue, researchers have proposed defence mechanisms against adversarial attacks. 
Adversarial training, which augments the training data with adversarially perturbed inputs, has shown moderate success at defending against recently proposed attack techniques \citep{szegedy:intriguing2013, goodfellow:explaining2014, tramer:ensemble2017, madry:towards2018}. In addition, recent advances in Bayesian neural networks have demonstrated that uncertainty estimates can be used to detect adversarial attacks \citep{li:dropout2017, feinman:detecting2017,louizos:multiplicative2017,smith:understanding2018}. 
Another notable category of defence techniques involves the usage of generative models. For example, \citet{gu:towards2014} used an auto-encoder to denoise the inputs before feeding them to the classifier. This denoising approach has been extensively investigated, and the ``denoisers'' in usage include generative adversarial networks \citep{samangouei:defensegan2018}, PixelCNNs \citep{song:pixeldefend2018} and denoising auto-encoders \citep{kurakin:adversarial2018}.
These developments rely on the \emph{``off-manifold''} conjecture -- adversarial examples are far away from the data manifold, although \citet{gilmer:sphere2018} has challenged this idea with a synthetic ``sphere classification'' example.

Surprisingly, much less recent work has investigated the robustness of \emph{generative classifiers} \citep{ng:discriminative2002} against adversarial attacks, where such classifiers explicitly model the conditional distribution of the inputs given labels. Typically, a generative classifier produces predictions by comparing between the likelihood of the labels for a given input, which is closely related to the ``distance'' of the input to the data manifold associated with a class. Therefore, generative classifiers should be robust to many recently proposed adversarial attacks if the ``off-manifold'' conjecture holds for many real-world applications. 
Unfortunately, many generative classifiers in popular use, including naive Bayes and linear discriminant analysis \citep{fisher:lda1936}, perform poorly on natural image classification tasks, making it difficult to verify the ``off-manifold'' conjecture and the robustness of generative classifiers with these tools. %In recent work, k-nearest neighbors \citep{cover:knn1967}, a method which shares many similarities with generative classifiers, has been significantly improved in handling natural images by leveraging deep feature representations \citep{papernot:deepknn2018}. To the best of our knowledge, an approach which targets a similar contribution has not yet been proposed for generative classifiers. 

Are generative classifiers more robust to recently proposed adversarial attack techniques? To answer this, we improve the naive Bayes algorithm by using deep generative models, and evaluate the conjecture on the proposed generative classifier. 
In summary, our contributions include:
\begin{itemize}
\vspace{-0.05in}
\setlength\itemsep{0em}
\item We propose \emph{deep Bayes} which models the (conditional) distribution of an input by a deep latent variable model (LVM). We learn the LVM with the variational auto-encoder algorithm \citep{kingma:vae2013, rezende:vae2014}, and for classification we approximate Bayes' rule using importance sampling.
\item We propose three detection methods for adversarial attacks. The first two use the learned generative model as a proxy of the data manifold, and reject inputs that are far away from it. The third computes statistics for the classifier's output probability vector, and rejects inputs that lead to under-confident predictions.
\item We evaluate the robustness of the proposed generative classifier on MNIST and a binary classification dataset derived from CIFAR-10. We further show the advantage of generative classifiers over a number of discriminative classifiers, including Bayesian neural networks and \emph{discriminative} LVMs. 
\item We improve the robustness of deep neural networks on CIFAR-10 \emph{multi-class} classification, by fusing discriminatively learned visual features with the proposed generative classifiers. On defending a number of popular $\ell_{\infty}$ attacks, the fusion model outperforms a baseline discriminative VGG16 network \citep{simonyan2014very} by a large margin.

\end{itemize}

\section{Deep Bayes: conditional deep LVM as a generative classifier}

Denote $p_{\data}(\x, \y)$ the data distribution for the input $\x \in \mathbb{R}^D$ and label $\y \in \{ \y_c | c = 1, ..., C\}$, where $\y_c$ is the one-hot encoding vector for class $c$. For a given $\x \in \mathbb{R}^D$ we can define the ground-truth label by 
$\y \sim p_{\data}(\y | \x) \text{ if } \x \in \text{supp}(p_{\data}(\x))$.
%
%We do not assume a ground-truth label for $\x \notin \text{supp}(p_{\data}(\x))$. %which is different from (CITE) that assume an oracle classifier that assign labels to any points in $\mathbb{R}^D$. %Under this assumption we define success of an attack in a way that is slightly different from the conventional one used in the adversarial machine learning community: \emph{an attack is successful if it can fool the classifier without being detected}.
%
We assume the data distribution $p_{\data}(\x, \y)$ follows the \emph{manifold assumption}: for every class $c$, the conditional distribution $p_{\data}(\x | \y_c)$ has its support as a low-dimensional manifold. 
Thus the training data $\data = \{ (\x^{(n)}, \y^{(n)}) \}_{n=1}^N$ is generated as follows:
$\y^{(n)} \sim p_{\data}(\y), \x^{(n)} \sim p_{\data}(\x | \y)$.
%We denote $\hat{p}_{\data}$ as the training data empirical distribution of $p_{\data}$, and assume a balanced data setting, i.e.~$\hat{p}_{\data}(\y) = p_{\data}(\y) = \text{Uniform}(\y)$.

%(VERY STRONG ASSUMPTION, DO WE REALLY NEED THIS?)
%For simplicity we further assume that 
%$$\int p_{\data}(\x) \bm{1}_{\x \in \mathcal{M}_{c_1} \cap \mathcal{M}_{c_2}} d\x = 0, \forall c_1, c_2,$$
%meaning that ambiguous inputs have zero probability to occur. 
%

%\subsection{Deep latent variable model as a generative classifier}

A generative classifier first builds a \emph{generative model} $p(\x, \y) = p(\x | \y) p(\y)$, and then, in prediction time, predicts the label $\y^*$ of a test input $\x^*$ using Bayes' rule,
\begin{equation*}
p(\y^* | \x^*) = \frac{p(\x^* | \y^*) p(\y^*)}{p(\x^*)} = \text{softmax}_{c=1}^C \left[ \log p(\x^*, \y_c) \right],
\end{equation*}
where $\text{softmax}_{c=1}^C$ denotes the softmax operator over the $c$ axis. Here the output probability vector is computed analogously to many discriminative classifiers which use softmax activation in the output layer, so many existing attacks can be tested directly. However, unlike discriminative classifiers, the ``logit'' values prior to softmax activation have a clear meaning here, which is the log joint distribution $\log p(\x^*, \y_c)$ of input $\x^*$ and a given label $\y_c$. Therefore, one can also analyse the logit values to determine whether the unseen pair $(\x^*, \y^*)$ is legitimate, a utility which will be discussed further in section \ref{sec:detection}.

\emph{Naive Bayes} is perhaps the most well-known generative classifier; it assumes a factorised distribution for the conditional generator, i.e. $p(\x | \y) = \prod_{d=1}^D p(x_d | \y)$, which is inappropriate for e.g.~image and speech data. Fortunately, we can leverage the recent advances in generative modelling and apply a deep generative model for the joint distribution $p(\x, \y)$. 
More specifically, we use a deep latent variable model (LVM) $p(\x, \z, \y)$ to construct a generative classifier. Importantly, this leads to a conditional distribution 
$
p(\x | \y) = \frac{\int p(\x, \z, \y) d\z}{\int p(\x, \z, \y) d\z d\x}
$
that is \emph{not} factorised (even when $p(\x | \z, \y)$ is), which is much more powerful than naive Bayes. 
We refer to such generative classifiers that use deep generative models as \emph{deep Bayes} classifiers.

\begin{figure}[t]
\centering
\includegraphics[width=0.85\linewidth]{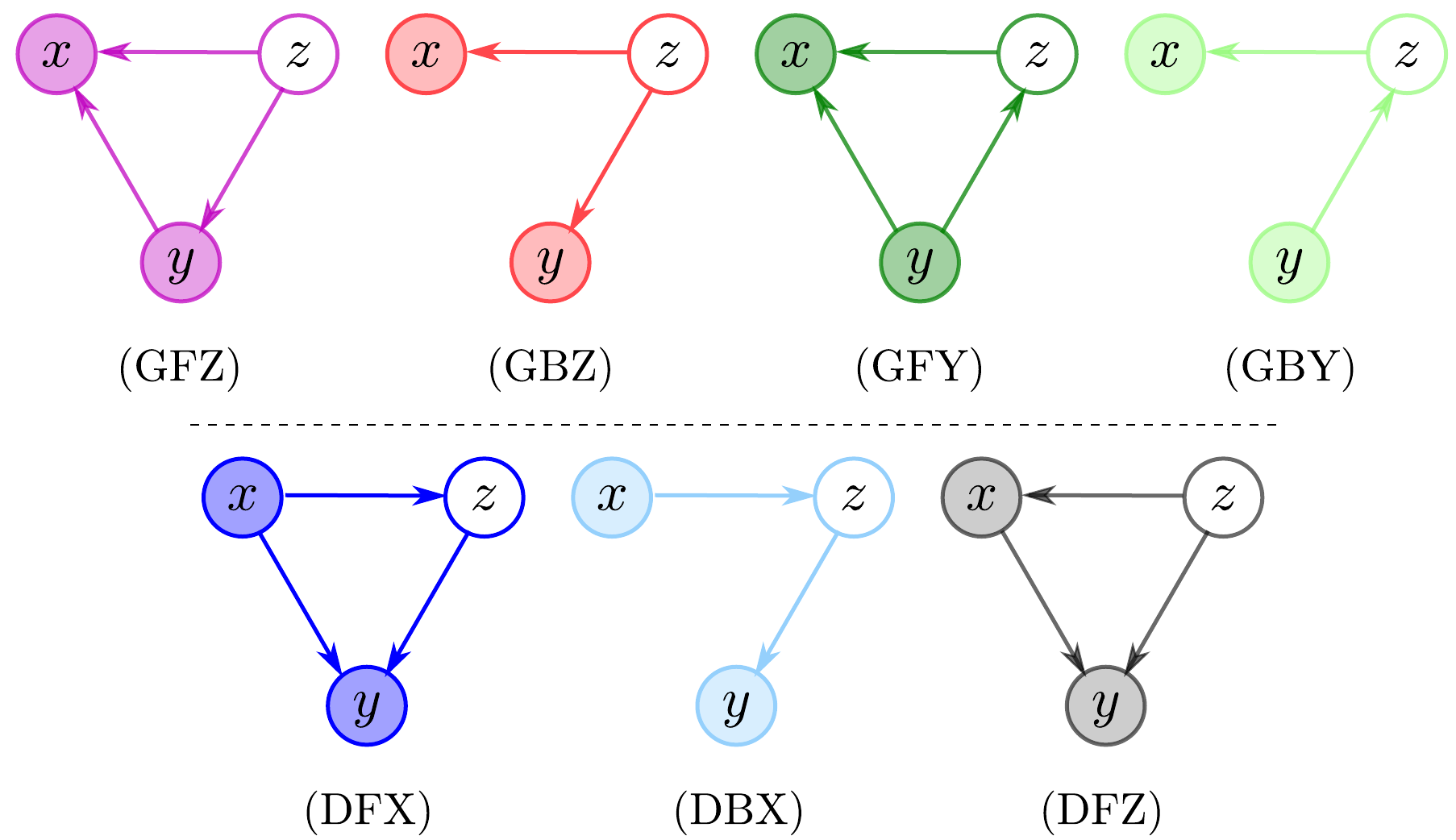}
\vspace{-0.12in}
\caption{A visualisation of the graphical models, including both \textbf{G}enerative (top row) and \textbf{D}iscriminative ones (bottom row), as well as \textbf{F}ully connected and \textbf{B}ottleneck ones. The last character indicates the first node in the topological order of the graph. The graphs are colored in a consistent way as in the result figures.}
\vspace{-0.12in}
\label{fig:graphical_model}
\end{figure}

Depending on the definition of the model $p(\x, \z, \y)$, the resulting classifier can be either generative or discriminative. Thus we evaluate the effect of different factorisation structures on the robustness of the induced classifier from the joint distribution $p(\x, \z, \y)$ (see Figure \ref{fig:graphical_model}). 
%
%\begin{minipage}{1\linewidth}
\begin{align*}
p(\x, \z, \y) &= p(\z) p(\y | \z) p(\x | \z, \y) \tag{GFZ} \label{eq:gfz} \\
p(\x, \z, \y) &= p_{\data}(\y) p(\z | \y) p(\x | \z, \y) \tag{GFY} \label{eq:gfy} \\
p(\x, \z, \y) &= p(\z) p(\y | \z) p(\x | \z) \tag{GBZ} \label{eq:gbz} \\
p(\x, \z, \y) &= p_{\data}(\y) p(\z | \y) p(\x | \z) \tag{GBY} \label{eq:gby} \\
p(\x, \z, \y) &= p_{\data}(\x) p(\z | \x) p(\y | \z, \x) \tag{DFX} \label{eq:dfx} \\
p(\x, \z, \y) &= p(\z) p(\x | \z) p(\y | \z, \x) \tag{DFZ} \label{eq:dfz} \\
p(\x, \z, \y) &= p_{\data}(\x) p(\z | \x) p(\y | \z) \tag{DBX} \label{eq:dbx}
\end{align*}
%\end{minipage}
%
%
We use the initial character ``G'' to denote generative classifiers and ``D'' to denote discriminative classifiers. 
Models with the second character as ``F'' have a \emph{fully connected} graph, while ``B'' models have \emph{bottleneck} structures. 
The last character of the model name indicates the first node in topological order. 
We do not test other architectures under this nomenclature, as either the graph contains directed cycles (e.g.~$\x \rightarrow \y \rightarrow \z \rightarrow \x$), or $\z$ is the last node in topological order (e.g.~$\x \rightarrow \y, (\x, \y) \rightarrow \z$) so that the marginalisation of $\z$ does not affect classification.

Within the generative classifiers in our design, different graphical models impose different assumptions on the data generation process. E.g.~\ref{eq:gfz} and \ref{eq:gbz} assume there is a confounder $\z$ that affects both $\x$ and $\y$, while \ref{eq:gfy} and \ref{eq:gby} assume the distribution over $\z$ is class-dependent. 
The bottleneck models \ref{eq:gbz}, \ref{eq:gby} and \ref{eq:dbx}, when compared with their fully-connected counterparts, enforces the usage of the latent code $\z$ for representation learning. 

For training, we follow \citet{kingma:vae2013} and \citet{rezende:vae2014} to introduce an amortised approximate posterior $q(\z | \x, \y)$, and train both $p$ and $q$ by maximising the \emph{variational lower-bound}:
\begin{equation}
\mathbb{E}_{\mathcal{D}}[  \mathcal{L}_{\text{VI}}(\x, \y) ] = \frac{1}{N} \sum_{n=1}^N \mathbb{E}_{q} \left[ \log \frac{p(\x_n, \z_n, \y_n)}{q(\z_n | \x_n, \y_n)} \right].
\end{equation}
After training, the predicted class probability vector $\y^*$ for a future input $\x^*$ is computed by an approximation to Bayes' rule with importance sampling $\z^k_c \sim q(\z | \x^*, \y_c)$:
\begin{equation}
\hspace{-0.05in}
p(\y^* | \x^*) \approx \text{softmax}_{c=1}^C \left[  \log \frac{1}{K} \sum_{k=1}^K \frac{p(\x^*, \z^k_c, \y_c)}{q(\z^k_c | \x^*, \y_c)} \right].
\label{eq:deep_bayes_vae}
\end{equation}
Therefore, for generative classifiers the probability of $\x$ under the generative model affects the predictions. 
By contrast, \ref{eq:dfx} and \ref{eq:dbx} do not know if $\x$ is close to the data manifold or not, as the $p_{\mathcal{D}}(\x)$ term in these models is cancelled out in eq.~(\ref{eq:deep_bayes_vae}).
Model \ref{eq:dfz} is somewhat intermediate, as it builds a generative model for the inputs $\x$ (thus $p(\x, \z)$ is used in eq.~(\ref{eq:deep_bayes_vae})) but also directly parameterises the conditional distribution $p(\y|\x, \z)$. 

\section{Detecting adversarial attacks with generative classifiers}
\label{sec:detection}
We propose detection methods for adversarial examples using the generative classifier's logit values.
As an illustrating example, consider a labelled dataset of ``cat'' and ``dog'' images. If an adversarial image of a cat $\x_{\text{adv}}$ is incorrectly labelled as ``dog'', then either this image is ambiguous, or, under a \emph{perfect} generative model, the logit $\log p(\x_{\text{adv}}, \text{``dog''})$ will be significantly lower than normal. 
This means we can detect attacks using the logits $\log p(\x^*, \y_c), c = 1, ..., C$ computed on a test input $\x^*$. 
The goal here is to reject both unlabelled input $\x$ that have low probability under $p(\x)$ (as a proxy to $p_{\data}(\x)$), and labelled data $(\x, \y)$ that have low $p(\x, \y)$ values.
Concretely, the proposed detection algorithms are as follows.
\begin{itemize}
\vspace{-0.1in}
\setlength\itemsep{0em}
\item \textbf{Marginal detection}: rejecting inputs that are far away from the manifold. \\
One can select a threshold $\delta$ and reject an input $\x$ if $-\log p(\x) > \delta$. 
%If the log-likelihood is defined using a distance metric (which is the case for CMP), then this approach would eliminate all the inputs outside of the $\delta$-ball $\cup_{c=1}^C \mathcal{B}(\hat{\mathcal{M}}_c, \delta)$. 
To determine the threshold $\delta$, we can compute $\bar{d}_{\data} = \mathbb{E}_{\x \sim \data} [-\log p(\x)]$ and $\sigma_{\data} = \sqrt{ \mathbb{V}_{\x \sim \data} [\log p(\x)] }$, then set $\delta = \bar{d}_{\data} + \alpha \sigma_{\data}$. %It is also possible to compute the statistics $\bar{d}_p, \sigma_p$ on the images generated by the generative model accordingly. 

\item \textbf{Logit detection}: rejecting inputs using joint density. \\
Given a victim model $\y = F(\x)$, one can reject $\x$ if $-\log p(\x, F(\x)) > \delta_{\y}$. We can use the mean and variance statistics $\bar{d}_c, \sigma_c$ computed on $\log p(\x, \y_c)$ and select $\delta_{\y_c} = \bar{d}_c + \alpha \sigma_c$.

%For discriminative classifiers we use the logit values before softmax as a proxy, but keep in mind that these logits cannot be interpreted as (log) joint density values.

%More interestingly, if the manifolds are well separated in terms of distance $d(\cdot, \cdot)$, we can set $\delta = \frac{1}{2} \min_{c_1, c_2} d(\hat{\mathcal{M}}_{c_1}, \hat{\mathcal{M}}_{c_2})$, and this approach will eliminate all adversarial inputs of the victim model that are within the $\delta$-ball $\cup_{c=1}^C \mathcal{B}(\hat{\mathcal{M}}_c, \delta)$. This is because an input $\x$ with ground-truth label $\y_c$ and $\x \in \mathcal{B}(\hat{\mathcal{M}}_{c}, \delta)$ has $d(\x, \hat{\mathcal{M}}_{c'}) > \delta, \forall c' \neq c$, which immediately implies $-\log p(\x | \y_{c'}) > \delta, \forall c' \neq c$ under the CMP model. 
%
\item \textbf{Divergence detection}: rejecting inputs with over- and/or under-confident predictions. \\
Denote $\mathbf{p}(\x)$ as a $C$-dimensional probability vector outputted by the classifier. For each class $c$, we first collect the \emph{mean classification probability vector} $\mathbf{p}_c = \mathbb{E}_{(\x, \y_c) \in \data}[\mathbf{p}(\x)]$, then compute the mean $\bar{d}_c$ and standard deviation $\sigma_c$ on a selected divergence/distance measure $\mathrm{D}[\mathbf{p}_c || \mathbf{p}(\x)]$ for all $(\x, \y_c) \in \data$. A test input $\x^*$ with prediction label $c^* = \argmax \mathbf{p}(\x^*)$ is rejected if $\mathrm{D}[\mathbf{p}_{c^*} || \mathbf{p}(\x^*)] > \bar{d}_{c^*} + \alpha \sigma_{c^*}$. Therefore, an example $\x^*$ can be rejected if the probability vector $\mathbf{p}(\x^*)$ is very different to the ones seen in training. 

When $\mathrm{D}$ is the KL-divergence, we call this method \emph{KL detection}. Other divergence/distance measures such as total variation (TV) can also be used.
\vspace{-0.1in}
\end{itemize}

\begin{figure}
\includegraphics[width=1.0\linewidth]{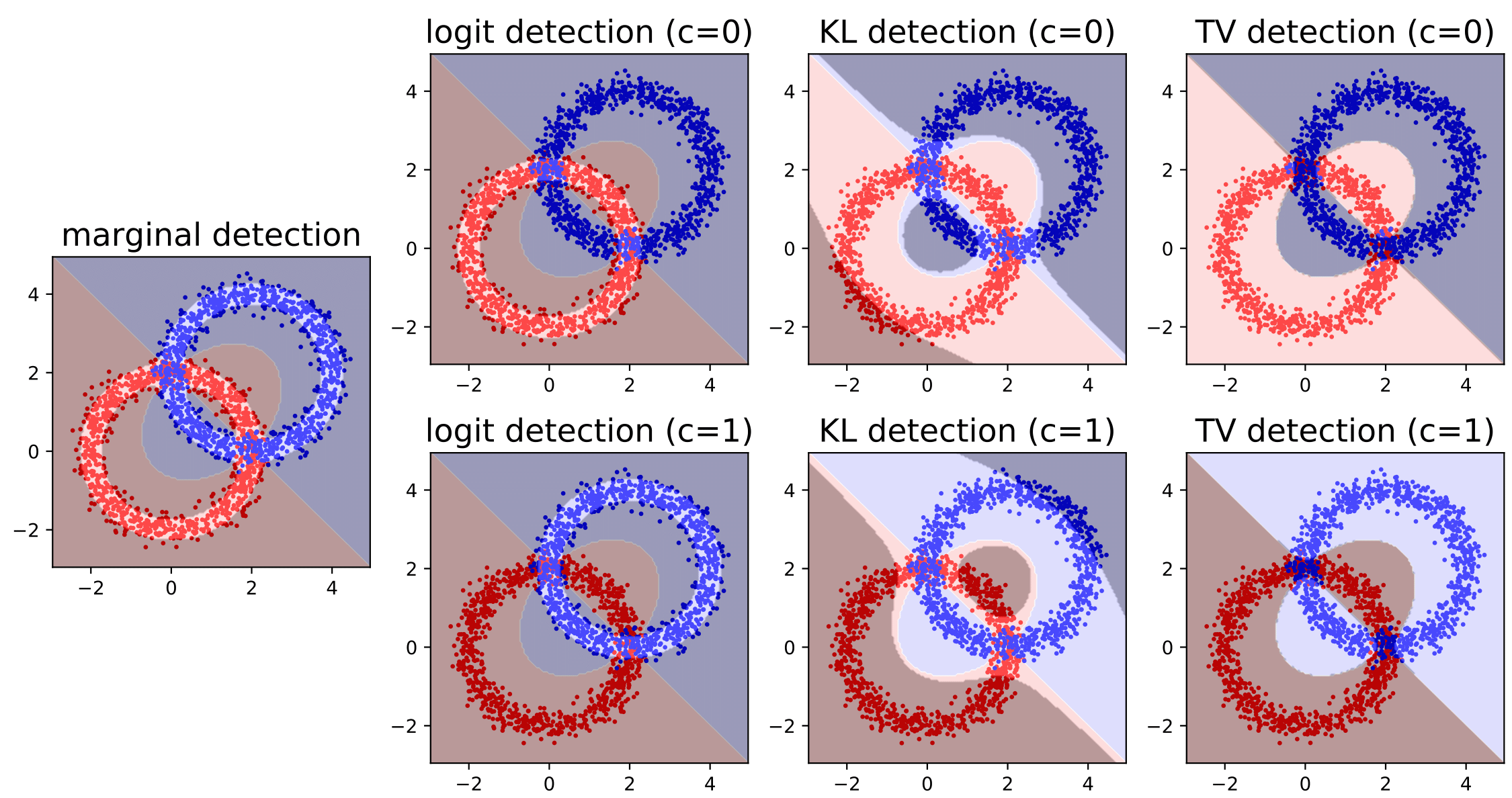}
\vspace{-0.25in}
\caption{Visualising detection mechanisms. The scattered dots are training data points, with different classes shown in different colours (red for $c=0$ and blue for $c=1$). Same labels are manually assigned for inputs when the detection method requires $\y$. Decision regions are shown in the corresponding colours. Input points in the shaded area are rejected by detection methods.}
\label{fig:detection_toy}
\vspace{-0.10in}
\end{figure}

For better intuition, we visualise the detection mechanisms in Figure \ref{fig:detection_toy} with a synthetic ``two rings'' binary classification example. In this case we sample $2\times 1000$ training data points by
%\begin{equation*}
%\centering
%\begin{aligned}
%    \y \sim \text{Bern}(0.5), \quad \theta \sim \text{Uniform}(0, 2\pi), \\
%    \x | \y \sim \mathcal{N}(\x; \bm{c}_{\y} + r_{\y} [cos(\theta), sin(\theta)]^{\text{T}}, \sigma^2 \mathbf{I}).
%\end{aligned}
%\end{equation*}
$\y \sim \text{Bern}(0.5), \theta \sim \text{Uniform}(0, 2\pi), 
    \x | \y \sim \mathcal{N}(\x; \bm{c}_{\y} + r_{\y} [cos(\theta), sin(\theta)]^{\text{T}}, \sigma^2 \mathbf{I}).$
We consider a generative classifier 
%\begin{equation*}
%\centering
%\begin{aligned}
%p(\x, \y) &= p(\x|\y) p_{\data}(\y) = \mathcal{N}(\x; \bm{\mu}_{\y}, \sigma^2 \mathbf{I}) \text{Bern}(0.5), \\
%\bm{\mu}_{\y} &= \text{proj}(\x, \text{ring}_{\y}) = \argmin_{|| \hat{\x} - \bm{c}_{\y} ||_2 = r_{\y}} || \x - \hat{\x} ||_2.
%\end{aligned}
%\end{equation*}
$
p(\x, \y) = p(\x|\y) p_{\data}(\y), p(\x|\y) = \mathcal{N}(\x; \bm{\mu}_{\y}, \sigma^2 \mathbf{I}),
\bm{\mu}_{\y} = \argmin_{|| \hat{\x} - \bm{c}_{\y} ||_2 = r_{\y}} || \x - \hat{\x} ||_2.
$
The $\delta$ thresholds are selected to achieve $10\%$ false positive rates on training data.
From the visualisations we see that inputs that are far away from the model manifold are rejected by marginal/logit detection. At the same time, logit detection rejects data points that are not on the manifold of the given class. KL/TV detection does not construct manifold-aware acceptance regions, which is as expected since the proposed divergence detection method does not require the classifier to be generative. However, both detection methods have some success in rejecting uncertain predictions, especially for TV, which also rejects ambiguous inputs (see the ring-cross regions in the last two plots). Combining all three methods, we see that the rejected inputs are either far away from the manifold, or are ambiguous. 

Detection methods using logit values have been used in e.g.~\citet{li:dropout2017, feinman:detecting2017}, but it is unclear whether the logits values in discriminative classifiers have a semantic meaning.  
\citet{song:pixeldefend2018,samangouei:defensegan2018,kurakin:adversarial2018} trained a \emph{separate} generative model for denoising/detection. But the features of the generator and the discriminative classifier can be very different, hence the generator cannot detect the ``manifold attack'' against the classifier \citep{gilmer:sphere2018}.
Unlike these approaches, we highlight three critical properties of generative classifiers and the accompanied detection methods:
\begin{itemize}
\vspace{-0.1in}
\setlength\itemsep{0em}
    \item[1.] the representations of the data manifold are the same for both the classifier and the detector (since they share the same generative model $p(\x, \y) \approx p_{\data}(\x, \y)$;
    \item[2.] the logit values in generative classifiers have a clear semantic meaning: the log probability $\log p(\x, \y)$ of generating the input $\x$ given the class label $\y = \y_c$; also note that $F(\x) = \argmax_{\y_c} \log p(\x, \y_c)$;
    \item[3.] the marginal/logit detection aims at rejecting $\x$ that has low $\log p(\x)$ and/or $\log p(\x, \y)$. These probabilities are obtained as part of the classification procedure and so do not require the running of any extra models.
\vspace{-0.1in}
\end{itemize}
Assuming an accurate approximation $p(\x, \y) \approx p_{\data}(\x, \y)$, a new input $\x^*$ is accepted by marginal/logit detection only if $\x^*$ is close to the manifold of $F(\x^*)$ (thus $p(\x^*, \y=F(\x^*))$ is high).\footnote{\citet{nalisnick2018do} showed that existing generative models fail to detect all possible outliers. A solution to counter this issue in logit/marginal detection is to add a lower-bounding threshold so that an input-output pair $(\x, \y)$ will be rejected if $-\log p(\x, \y) < \zeta_{\y}$. We leave this investigation to future work.} 
So $F(\x^*)$ should be equal to the ground truth label $\y^*$ if $\x^*$ is not an ambiguous input (which will be detected by KL/TV detection).  
Therefore the ``off-manifold'' conjecture should hold for a powerful generative classifier, and below we present an empirical study to validate the assumptions of this conjecture.

\section{Experiments}

We carry out a number of tests on the deep Bayes classifiers, our implementation is available at \url{https://github.com/deepgenerativeclassifier/DeepBayes}.

The distributions $q(\z | \cdot)$ and $p(\z | \cdot)$ are factorised Gaussians, and the conditional probability $p(\x | \cdot)$, if required, is parameterised by an $\ell_2$ loss. 
Besides the LVM-based classifiers, we further train a Bayesian neural network (BNN) with Bernoulli dropout (dropout rate 0.3), as it has been shown in \citet{li:dropout2017} and \citet{feinman:detecting2017} that BNNs are more robust than their deterministic counterparts. The constructed BNN has 2x more channels than LVM-based classifiers, making the comparison slightly ``unfair'', as the BNN layers have more capacity. We use $K=10$ Monte Carlo samples for all the classifiers.

The adversarial attacks are taken from the CleverHans 2.0 library \citep{papernot:cleverhans2017}.
Three metrics are reported: \emph{accuracy} of the classifier on crafted adversarial examples, mean \emph{minimum perturbation distance} computed on adversarial examples that have successfully fooled the classifier, and \emph{detection rate} on successful attacks. This detection rate is defined as the true positive (TP) rate of finding an adversarial example, and the detection threshold is selected to achieve a $5\%$ false positive rate on clean training data.

The experiments are performed under various threat model settings. In the main text we present gradient-based attacks and gradient masking sanity checks, and provide a brief summary of further experiments presented in the appendix. Full table results can also be found in appendix \ref{sec:appendix_full_results}. 

\subsection{Gradient-based attacks}

We first evaluate the robustness of generative classifiers against gradient-based attacks. These attacks are performed under a white-box setting  \emph{against the classifier} \citep{carlini:bypass2017}: the attacker can differentiate through the classifier, but is not aware of the existence of the detector. We report results on two datasets (MNIST and CIFAR-binary) and two classes of attacks ($\ell_{\infty}$ and $\ell_2$ attacks).

\paragraph{Datasets}

For MNIST tests, we use $\text{dim}(\z) = 64$ for the LVM-based classifiers. All classifiers achieve $>98\%$ clean test accuracy, therefore we apply attacks on the whole test dataset (10,000 datapoints).  
Since the robustness properties of MNIST classifiers might not extend to natural images \citep{carlini:bypass2017}, we further consider the same set of evaluations on \emph{CIFAR-binary}, a binary classification dataset containing ``airplane'' and ``frog'' images from CIFAR-10. We choose to work with this simpler dataset because (1) \emph{fully} generative classifiers are less satisfactory for classifying clean CIFAR-10 images\footnote{The clean test accuracies for \ref{eq:gfz} \& \ref{eq:gfy} on CIFAR-10 are all $<50\%$; a conditional PixelCNN++ \citep{salimans:pixelcnn++2017} (with much deeper networks) achieves $72.4\%$ clean test accuracy.}, and (2) we want to evaluate whether the robustness properties of \emph{fully} generative classifiers hold on natural images \citep[c.f.][]{carlini:bypass2017}.  On CIFAR-binary, the generative classifiers use $\text{dim}(\z) = 128$ and obtain $>90\%$ clean test accuracy (see appendix). The attacks are then performed on the test images that all models initially correctly classify, leading to a test set of 1,577 instances for CIFAR-binary. For both datasets the images are scaled to [0, 1].

\begin{figure*}[t]
\centering
\includegraphics[width=1\linewidth]{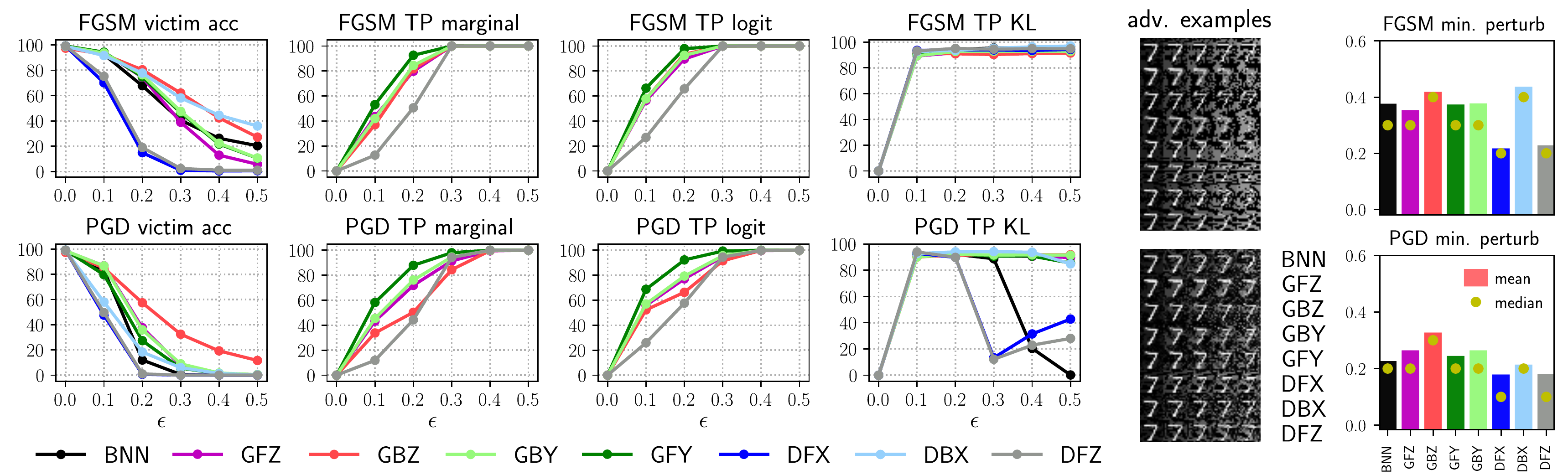}
\vspace{-0.25in}
\captionof{figure}{Victim accuracy, detection rates  and minimum $\ell_{\inf}$ perturbation against \textbf{white-box FGSM attacks} on MNIST. The higher the better. The visualised adversarial examples (not necessarily successful) are crafted with with $\ell_{\infty}$ distortion $\epsilon$ growing from 0.1 to 0.5.}
\label{fig:mnist_white_box}
\vspace{0.1in}
\includegraphics[width=1\linewidth]{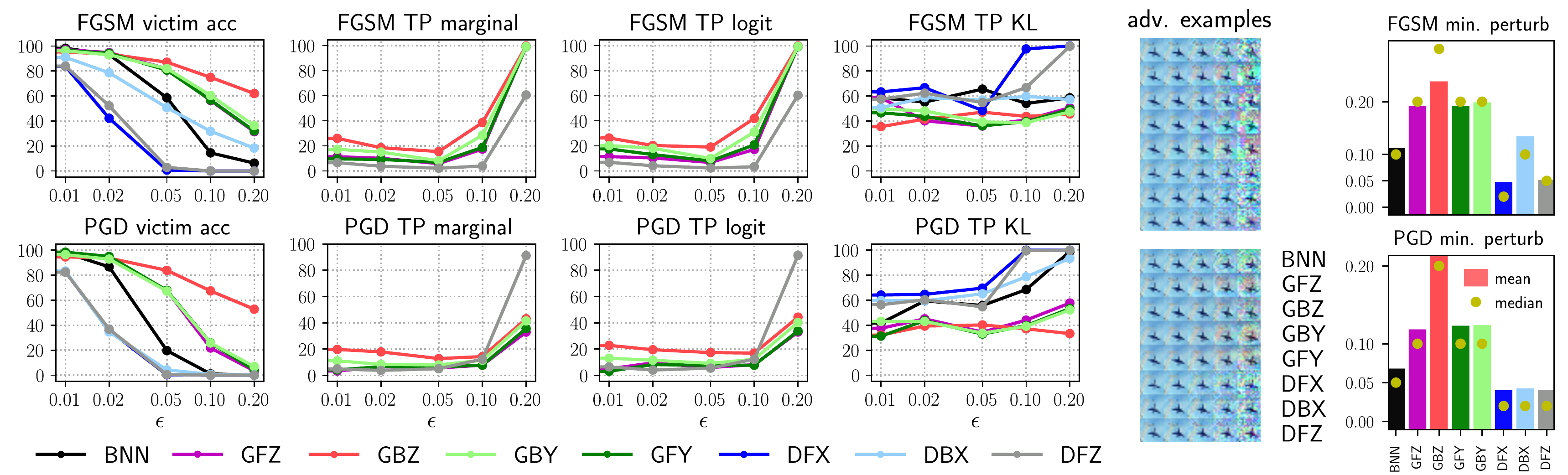}
\vspace{-0.25in}
\captionof{figure}{Victim accuracy, detection rates and minimum $\ell_{\inf}$ perturbation against \textbf{white-box FGSM attacks} on CIFAR-binary. The higher the better. The visualised adversarial examples (not necessarily successful) are crafted with $\ell_{\infty}$ distortion $\epsilon$ growing from 0.01 to 0.2.}
\vspace{-0.1in}
\label{fig:cifar_white_box}
\end{figure*}

\paragraph{$\ell_{\infty}$ attacks}
The attacks in test are: fast gradient sign method \citep[FGSM,][]{goodfellow:explaining2014}, projected gradient descent \citep[PGD,][]{madry:towards2018} and momentum iterative method \citep[MIM,][]{dong:mim2017}.\footnote{See appendix \ref{sec:appendix_additional_to_main} for MIM results with similar observations as in the main text.} We use the distortion strengths as $\epsilon \in \{0.1, 0.2, 0.3, 0.4, 0.5 \}$ for MNIST, and $\epsilon \in \{0.01, 0.02, 0.05, 0.1, 0.2 \}$ for CIFAR-binary.

Results are reported in Figure \ref{fig:mnist_white_box} for MNIST and Figure \ref{fig:cifar_white_box} for CIFAR-binary, respectively. 
For both datasets, generative classifiers perform generally better in terms of victim accuracy,
especially \ref{eq:gbz} is significantly more robust than the others on CIFAR-binary.  
By contrast, discriminative VAE-based classifiers are less robust, e.g.~on MNIST, \ref{eq:dfx} \& \ref{eq:dfz} are not robust to the weakest attack (FGSM) even when $\epsilon = 0.2$ (where the distorted inputs are still visually close to the original digit ``7'').
Interestingly, \ref{eq:dbx} is relatively robust against FGSM on both datasets, which agrees with the preliminary tests in \citet{alemi:deepvib2017}. Further investigations in appendix \ref{sec:appendix_further_exp_bottleneck} show that the bottleneck structure might be beneficial for defending certain attacks. %Classifier \ref{eq:dbx} is less robust to PGD though.
%
%BNN is sometimes better than other discriminative LVM-based classifiers (especially on CIFAR-binary), presumably due to higher randomness.

In terms of minimum perturbation which is computed across all $\epsilon$ settings,\footnote{We manually assign the minimum perturbation of an input as $\epsilon_{\text{max}} + 0.1$ if none of the attacks is successful.} quantitatively the amount of distortion required to fool generative classifiers is much higher than that for discriminative ones. Indeed on both datasets, the visual distortion of the adversarial examples on generative classifiers is more significant.

For detection, generative classifiers successfully detect the adversarial examples with $\epsilon \geq 0.3$ on MNIST, which is reasonable as the visual distortion is already significant.
Detection results on CIFAR-binary are less satisfactory though: marginal/logit detection fail to detect attacks with $\epsilon = 0.1$ (which attain both high success rate and induce visually perceptible distortion). These results suggest that the per-pixel $\ell_2$ loss might not be best suited for modelling natural images \citep[c.f.][]{larsen:vaegan2016, oord:pixel2016}, indeed we present improved robustness results in section \ref{sec:experiments_fusion}, where the generative classifiers use an alternative likelihood function that is closely related to the \emph{perceptual loss} \citep{dosovitskiy2016generating,johnson2016perceptual}.

As a side note, \ref{eq:dfz}, as an intermediate between generative and discriminative classifiers, has worse robustness results, but has good detection performance for the marginal and logit metrics. This is because with softmax activation, the marginal distribution $p(\x)$ is dropped, but in marginal/logit detection $p(\x)$ is still in use.
KL detection works well for all classifiers, and on CIFAR-binary, discriminative classifiers start to dominate in this metric as $\epsilon$ increases. Inspecting the visualised adversarial examples on generative classifiers (Figure \ref{fig:cifar_white_box}), we see that for large $\epsilon$ values, these inputs are significantly distorted, thus ``not ambiguous'', indeed they are detected by marginal/logit detection methods.

%%% CW %%%

\begin{figure}[t]
\centering
\includegraphics[width=0.95\linewidth]{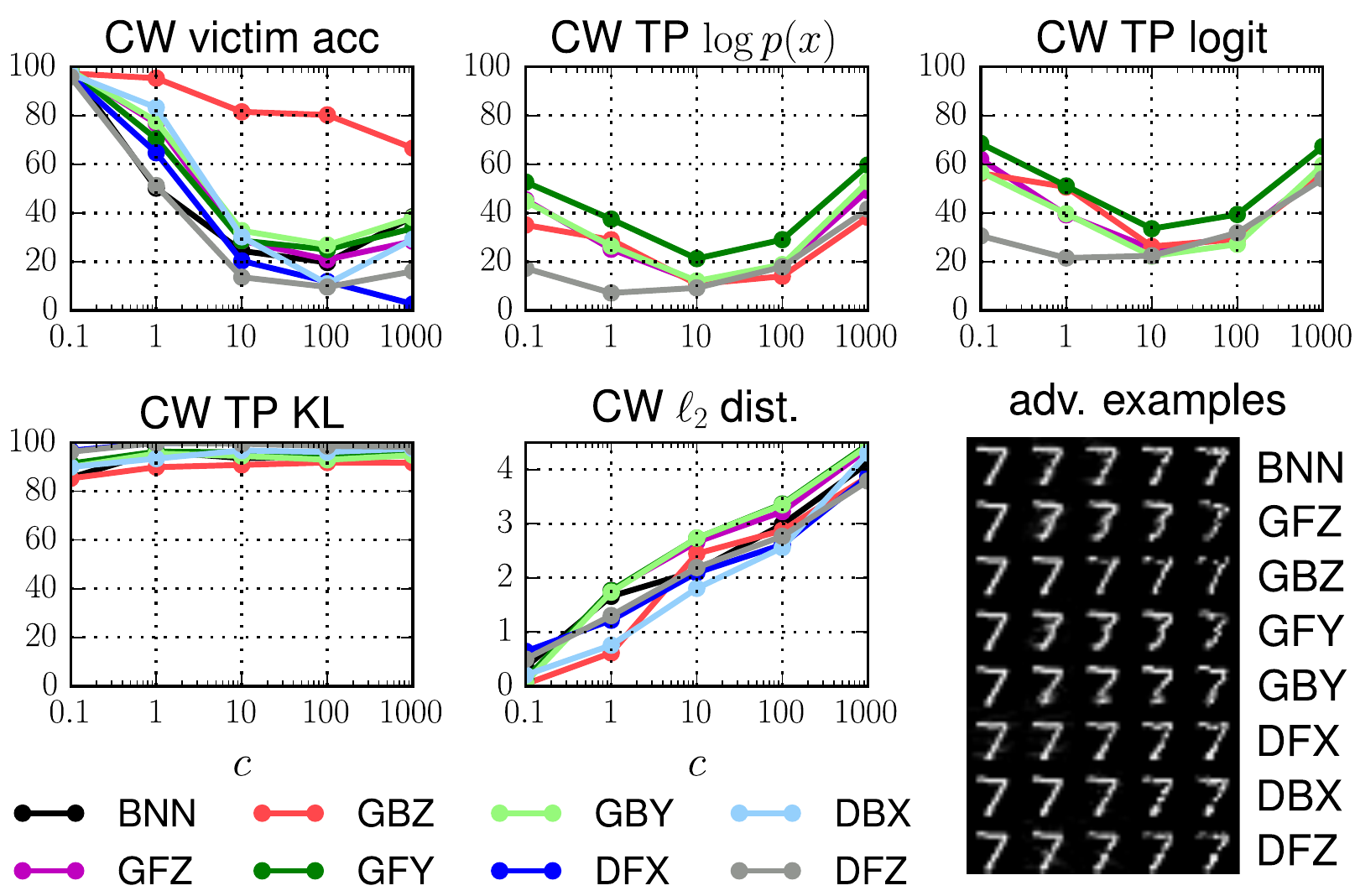}
\vspace{-0.05in}
\captionof{figure}{Accuracy, $\ell_2$ distortion, and detection rates against \textbf{white-box CW attacks} on MNIST. }
\label{fig:mnist_white_box_cw}
\vspace{0.05in}
\subfigure[clean inputs]{
\includegraphics[width=0.4\linewidth]{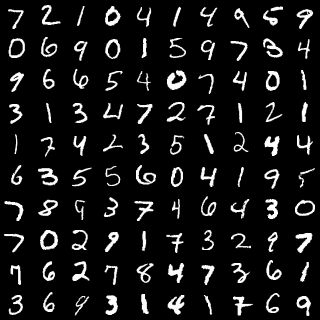}}
\hspace{0.15in}
\subfigure[CW adv.~inputs]{
\includegraphics[width=0.4\linewidth]{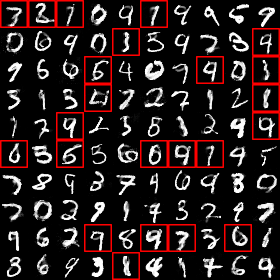}}
\vspace{-0.1in}
\captionof{figure}{Clean inputs and CW adversarial examples ($c=10$) crafted on \ref{eq:gfz}, digits in red rectangles show significant ambiguity.}
\label{fig:cw_attack_input}
\vspace{-0.12in}
\end{figure}

\paragraph{$\ell_{2}$ attack} We perform the Carlini \& Wagner (CW) $\ell_2$ attack \citep{carlini:attack2017}, with a loss-balancing parameter $c \in \{0.1, 1, 10, 100, 1000 \}$, so that the $\ell_2$ distortion regulariser in CW's loss function decreases with larger $c$. 

MNIST results are reported in Figure \ref{fig:mnist_white_box_cw}. Here \ref{eq:gbz} is a clear winner: for a given level of $\ell_2$ distortion, \ref{eq:gbz} is significantly more robust than the others. Other classifiers perform similarly on MNIST, however, these attack successes on generative classifiers is mainly due to the ambiguity of the crafted adversarial images. As visualised in Figure \ref{fig:cw_attack_input} (and Figure \ref{fig:cw_attack_appendix} in appendix), the induced distortion from CW leads to ambiguous digits which sit at the perceptual boundary between the original and the adversarial classes.
%
%The generative classifiers are generally better than discriminative ones on CIFAR-binary.

%
Interestingly, the detection rates on MNIST adversarial inputs do not grow monotonically as $c$ increases. Combined with the victim accuracy results, this means $c=10$ is the sweet-spot parameter that achieves the best success rates against both the classifier and the marginal/logit detection methods.
KL detection achieves $> 95\%$ detection rates on all $c$ and all the classifiers. This is as expected as the CW attack generates adversarial examples that lead to minimal difference between the logit values of the most and the second most probable classes. In this case the adversarial examples might correspond to ambiguous inputs (Figure \ref{fig:cw_attack_input}).

\begin{figure}[t]
\centering
\includegraphics[width=0.95\linewidth]{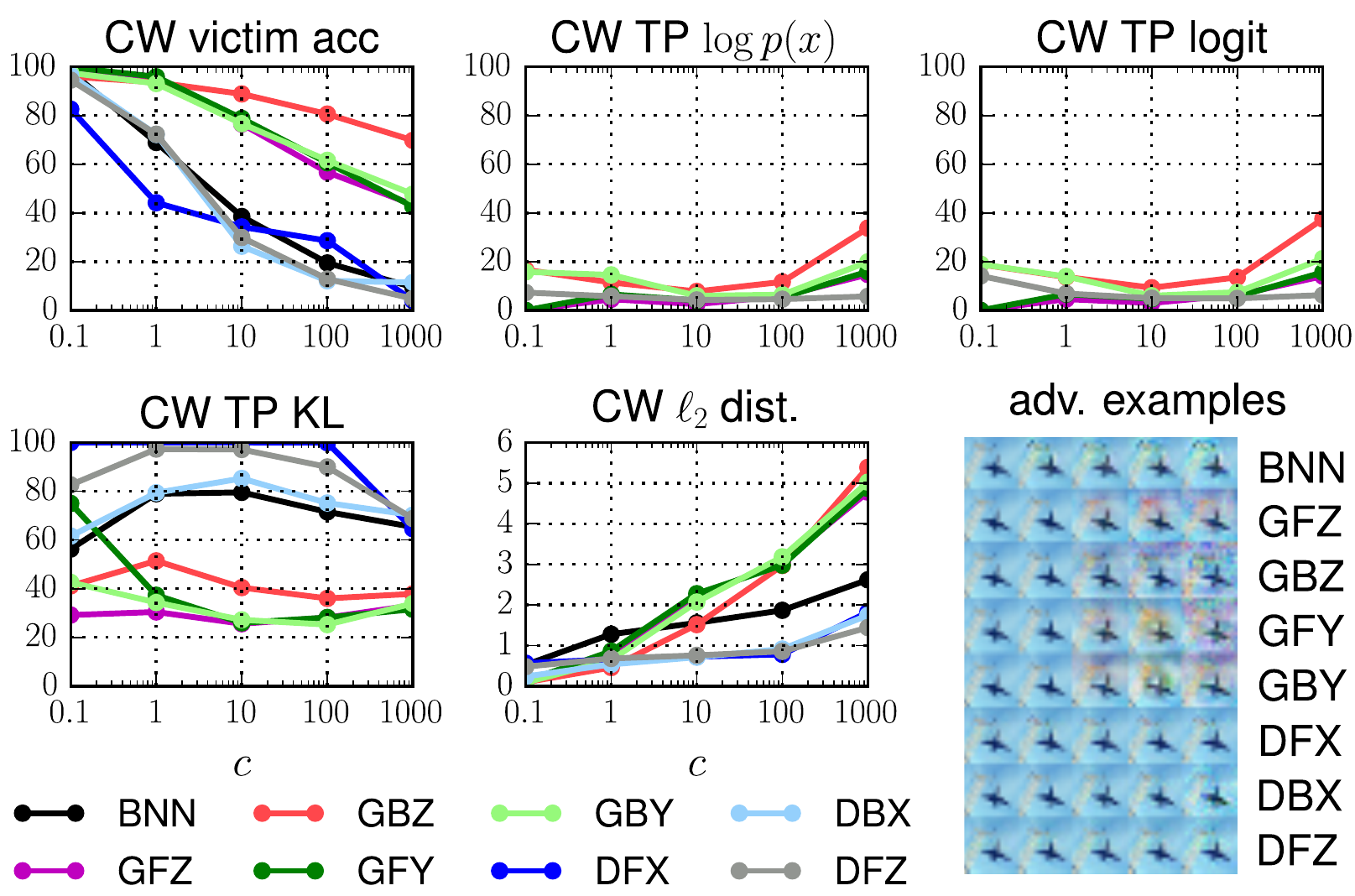}
\vspace{-0.05in}
\caption{Accuracy, $\ell_2$ distortion, and detection rates against \textbf{white-box CW attacks} on CIFAR-binary. }
\vspace{-0.15in}
\label{fig:cifar_white_box_cw}
\end{figure}

Figure \ref{fig:cifar_white_box_cw} presents the robustness and detection results for CIFAR-binary. Here the generative classifiers are significantly more robust than the others (with the best being \ref{eq:gbz}), and the mean $\ell_2$ distortions computed on successful attacks are also significantly higher. The TP rates are low for marginal/logit detection when $c$ is small, which is reasonable as the crafted images are visually similar to the clean ones. Note that the distortion for the attacks on generative classifiers is perceptible, and the logit TP rates also increase as $c$ increases. These results indicate that this CW attack is ineffective when attacking generative classifiers.

\paragraph{Summary}
The tested generative classifiers are generally more robust than the discriminative ones against white-box gradient-based attacks.
In particular, the generative classifier's victim accuracy decreases as the distortion increases, but at the same time the TP rates for marginal/logit detection also increase. Therefore the two attacks in test fail to find near-manifold adversarial examples that fool both the classifier and the detector components of generative classifiers.
%

%%%%%%%%%%% gradient masking

\subsection{Sanity checks on gradient masking}
If a successful defence against white-box gradient-based attacks is due to gradient masking, then this defence is likely to be less effective against attacks that do not differentiate through the victim classifier and the defence \citep{papernot:practical2017, athalye:obfuscated2018}.
Therefore, here we present two types of attacks as sanity checks on gradient masking.

\paragraph{Distillation-based attacks}
We perform two attacks based on distilling the victim classifier using a ``student'' CNN. The two attacks differ in their threat models: in the \textbf{grey-box} setting the attacker has access to both the training data and the output probability vectors of the classifiers on the training set, while in the \textbf{black-box} setting the attacker only has access to queried labels on a given input. For the latter black-box setting, we follow \citet{papernot:practical2017} to train a substitute CNN using Jacobian-based dataset augmentation (see appendix \ref{sec:appendix_black_box_distillation_algorithm}). We then craft adversarial examples on the grey-/black-box substitutes using PGD and MIN, and transfer them to the victim classifiers. As a sanity check, these attacks with reasonably small $\epsilon$ values achieve $\sim 100\%$ success rates on fooling the substitutes ($\epsilon \geq 0.2$ for MNIST and $\epsilon \geq 0.1$ for CIFAR-binary, see appendix \ref{sec:appendix_full_results}).

We report the results for these transferred attacks in Table \ref{table:mnist_min_dist} for MNIST and Table \ref{table:cifar_min_dist} for CIFAR-binary, respectively, with a comparison to white-box attack results taken from the last section.
It is clear that the adversarial examples crafted on substitute models do not transfer very well to the generative classifiers. Importantly, for a fixed $\epsilon$ setting, the white-box attacks achieve significantly higher success rates than their grey-/black-box counterparts, and the gap is at least $>20\%$ for MNIST (with $\epsilon \leq 0.3$) and $>30\%$ for CIFAR-binary (with $\epsilon \leq 0.1$). 
Furthermore, the mean minimum $\ell_{\infty}$ perturbation obtained by grey-/black-box attacks is significantly higher than those obtained by white-box attacks.

\begin{figure}[t]
\vspace{-0.1in}
%\begin{table}
\captionof{table}{Mean minimum $\ell_{\infty}$ perturbation (in red, computed on $\epsilon \in \{0.1, 0.2, 0.3, 0.4, 0.5 \}$) and victim accuracy (in blue, for $\epsilon \leq 0.3$) for $\ell_{\infty}$ attacks on MNIST.} %We manually assign the min.~perturbation $\epsilon=0.6$ to inputs that all attacks failed to find adversarial perturbations.}
\label{table:mnist_min_dist}
\centering
\scriptsize
\setlength{\tabcolsep}{4pt}
\begin{tabular}{l|cccc}\toprule
 Attack & GFZ & GBZ & GFY & GBY \\
\toprule
PGD (white) & \color{red}{0.23} \color{black}{/} \color{blue}{7.71\%} & \color{red}{0.30} \color{black}{/} \color{blue}{30.78\%} & \color{red}{0.21} \color{black}{/} \color{blue}{5.52\%} & \color{red}{0.23} \color{black}{/} \color{blue}{8.89\%} \\
MIM (white) & \color{red}{0.24} \color{black}{/} \color{blue}{9.02\%} & \color{red}{0.21} \color{black}{/} \color{blue}{4.97\%} & \color{red}{0.22} \color{black}{/} \color{blue}{6.72\%} & \color{red}{0.21} \color{black}{/} \color{blue}{1.54\%} \\
\midrule
PGD (grey) & \color{red}{0.37} \color{black}{/} \color{blue}{51.08\%} & \color{red}{0.36} \color{black}{/} \color{blue}{50.64\%} & \color{red}{0.38} \color{black}{/} \color{blue}{53.29\%} & \color{red}{0.36} \color{black}{/} \color{blue}{48.66\%} \\
MIM (grey) & \color{red}{0.34} \color{black}{/} \color{blue}{43.00\%} & \color{red}{0.33} \color{black}{/} \color{blue}{40.94\%} & \color{red}{0.34} \color{black}{/} \color{blue}{46.64\%} & \color{red}{0.33} \color{black}{/} \color{blue}{40.06\%} \\
\midrule
PGD (black) & \color{red}{0.40} \color{black}{/} \color{blue}{61.93\%} & \color{red}{0.42} \color{black}{/} \color{blue}{66.75\%} & \color{red}{0.38} \color{black}{/} \color{blue}{56.35\%} & \color{red}{0.43} \color{black}{/} \color{blue}{68.50\%} \\
MIM (black) & \color{red}{0.36} \color{black}{/} \color{blue}{50.44\%} & \color{red}{0.38} \color{black}{/} \color{blue}{59.86\%} & \color{red}{0.36} \color{black}{/} \color{blue}{48.07\%} & \color{red}{0.39} \color{black}{/} \color{blue}{61.78\%} \\
\bottomrule
\end{tabular}
%\end{table}
\vspace{-0.15in}
%
%\begin{table}
\captionof{table}{Mean minimum $\ell_{\infty}$ perturbation (in red, computed on $\epsilon \in \{0.01, 0.02, 0.05, 0.1, 0.2 \}$) and victim accuracy (in blue, for $\epsilon \leq 0.1$) for $\ell_{\infty}$ attacks on CIFAR-binary. }%We manually assign the min.~perturbation $\epsilon=0.3$ to inputs that all attacks failed to find adversarial perturbations.}
\label{table:cifar_min_dist}
\centering
\scriptsize
\setlength{\tabcolsep}{4pt}
\begin{tabular}{l|cccc}\toprule
 Attack & GFZ & GBZ & GFY & GBY  \\
\toprule
PGD (white) & \color{red}{0.11} \color{black}{/} \color{blue}{21.81\%} & \color{red}{0.20} \color{black}{/} \color{blue}{65.63\%} & \color{red}{0.11} \color{black}{/} \color{blue}{25.81\%} & \color{red}{0.11} \color{black}{/} \color{blue}{25.24\%} \\
MIM (white) & \color{red}{0.09} \color{black}{/} \color{blue}{15.22\%} & \color{red}{0.13} \color{black}{/} \color{blue}{37.60\%} & \color{red}{0.10} \color{black}{/} \color{blue}{16.4\%9} & \color{red}{0.09} \color{black}{/} \color{blue}{14.39\%} \\
\midrule
PGD (grey) & \color{red}{0.15} \color{black}{/} \color{blue}{50.48\%} & \color{red}{0.23} \color{black}{/} \color{blue}{77.30\%} & \color{red}{0.16} \color{black}{/} \color{blue}{54.66\%} & \color{red}{0.17} \color{black}{/} \color{blue}{57.96\%} \\
MIM (grey) & \color{red}{0.15} \color{black}{/} \color{blue}{47.62\%} & \color{red}{0.21} \color{black}{/} \color{blue}{75.71\%} & \color{red}{0.15} \color{black}{/} \color{blue}{51.11\%} & \color{red}{0.16} \color{black}{/} \color{blue}{53.84\%} \\
\midrule
PGD (black) & \color{red}{0.19} \color{black}{/} \color{blue}{68.36\%} & \color{red}{0.23} \color{black}{/} \color{blue}{79.45\%} & \color{red}{0.20} \color{black}{/} \color{blue}{70.13\%} & \color{red}{0.19} \color{black}{/} \color{blue}{67.98\%} \\
MIM (black) & \color{red}{0.18} \color{black}{/} \color{blue}{66.39\%} & \color{red}{0.23} \color{black}{/} \color{blue}{78.38\%} & \color{red}{0.19} \color{black}{/} \color{blue}{68.42\%} & \color{red}{0.19} \color{black}{/} \color{blue}{66.52\%} \\
\bottomrule
\end{tabular}
%\end{table}

\vspace{-0.1in}
\end{figure}

%%%%%%%% SPSA %%%%%%%%%%%%

\begin{table}
\vspace{-0.1in}
    \centering
    \caption{Victim accuracy results against \textbf{white-box PGD} and \textbf{black-box SPSA attack}. We use $\epsilon = 0.3$ for MNIST and $\epsilon=0.05$ for CIFAR-binary. }
    \label{tab:spsa_all}
    \small
    \begin{tabular}{c|cc|cc}\toprule 
          & \multicolumn{2}{c}{MNIST} & \multicolumn{2}{c}{CIFAR-binary} \\
          \midrule
          & PGD & SPSA & PGD & SPSA \\ 
          \midrule
          GFZ & 4.0\% & 68.2\% & 67.7\% & 96.4\% \\
          GBZ & 29.7\% & 79.0\% & 83.9\% & 95.2\% \\
          GBY & 7.4\% & 71.0\% & 67.9\% & 96.4\% \\
          GFY & 2.3\% & 55.9\% & 67.3\% & 96.3\% \\
          \bottomrule
    \end{tabular}
\vspace{-0.1in}
\end{table}

\paragraph{SPSA (evolutionary strategies)}
We consider another black-box setting that only assumes access to the logit values of the prediction given an input. We use the SPSA $\ell_{\infty}$ attack \citep{uesato:spsa2018}, which is similar to the white-box attacks, except that gradients are numerically estimated using the logit values from the victim classifier. 
Results are presented in Table \ref{tab:spsa_all} for MNIST (using 1,000 randomly sampled test datapoints) and CIFAR-binary. Both experiments clearly show that SPSA performs much worse on generative classifiers when compared to \emph{white-box} PGD.

\paragraph{Summary} Both distillation-based attacks and score-based attack (SPSA) failed to obtain higher success rate than gradient-based attacks against generative classifiers. Therefore, gradient masking is unlikely to be responsible for the improved robustness of generative classifiers, as utilising the exact gradient yielded better success rates for the attacks. 
%

%%%%%%%%%%%%%%%%%%%%%%%%%%%%%%%%

\subsection{Combining deep Bayes and discriminative features}
\label{sec:experiments_fusion}

This experiment examines the robustness of CIFAR-10 \emph{multi-class} classifiers, with the generative classifiers trained on \emph{discriminative} visual features. To do this, we download a pretrained VGG16 network \citep{simonyan2014very} on CIFAR-10 ($93.59\%$ test accuracy),\footnote{\url{https://github.com/geifmany/cifar-vgg}} and use its features $\bm{\phi}(\x)$ as the input to the VAE-based classifiers: $p(\y|\x) = p(\y | \bm{\phi}(\x))$. This means $p(\x | \cdot)$ of the generative classifiers is defined by a perceptual loss \citep{dosovitskiy2016generating,johnson2016perceptual}, see appendix \ref{sec:appendix_further_discussion} for a discussion.
The classifiers in test include \ref{eq:gbz}, \ref{eq:gby} and \ref{eq:dbx}. We use fully-connected neural networks for these classifiers, and select from VGG16 the $9^{\text{th}}$ convolution layer (\textsc{conv9}) and the first fully connected layer after convolution (\textsc{fc1}) as the feature layers to ensure $\sim 90\%$ test accuracy.

\begin{figure}[t]
\centering
\includegraphics[width=0.95\linewidth]{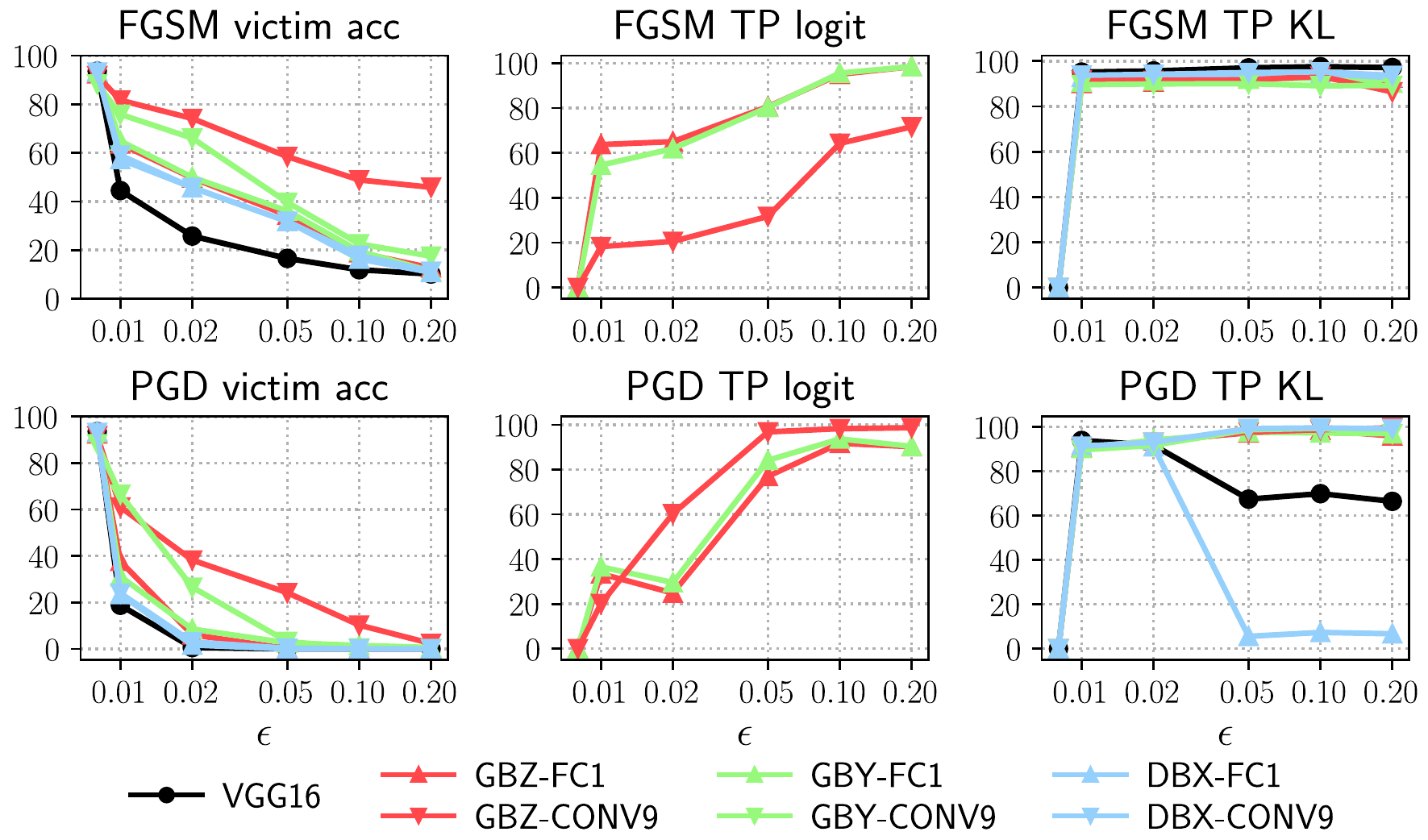}
\vspace{-0.1in}
\caption{Accuracy and detection rates against \textbf{white-box} $\ell_{\infty}$ attacks on CIFAR-10. The higher the better. Note that results for DBX, GBZ-\textsc{fc1} and GBY-\textsc{fc1} are almost identical.}
\vspace{-0.12in}
\label{fig:vgg_white_box}
\end{figure}

Results on white-box $\ell_{\infty}$ attacks are visualised in Figure \ref{fig:vgg_white_box}. For all LVM-based classifiers we see clear improvements in robustness and detection over the VGG16 baseline. In particular, \ref{eq:gbz} and \ref{eq:gby} with \textsc{conv9} features are overall better than \ref{eq:dbx}. More importantly, generative classifiers based on \textsc{conv9} features are significantly more robust than those based on \textsc{fc1} features. By contrast, for \ref{eq:dbx}, which is \emph{discriminative}, the robustness results are very similar. This indicates that the level of feature representation has little effect for \ref{eq:dbx}, presumably \ref{eq:dbx}-\textsc{conv9} has learned high-level features that resembles \textsc{fc1}. Also the logit detection method works much better on the fusion models when compared with the generative classifiers using $\ell_2$ likelihood (c.f.~Figure \ref{fig:cifar_white_box}). These results suggest that one can achieve both high clean accuracy and better robustness/detection rates against adversaries by combining discriminatively learned visual features and generative classifiers.

%%%%%%%%%%%% appendix summary %%%%%%%%%%%

\subsection{Summary of additional studies}
We present in appendix \ref{sec:appendix_further_exp} further studies on the robustness properties of generative and discriminative classifiers. 
\begin{itemize}
\vspace{-0.1in}
\setlength\itemsep{0em}
    \item In appendix \ref{sec:appendix_further_exp_pk}, we designed a white-box attack targeting both the classifier and the detector, which also considers the usage of random $\z$ samples by the LVM-based classifiers \citep{biggio2013evasion, carlini:bypass2017}. In results, although this attack can reduce detection levels, it comes with the trade-off of increasing accuracy, suggesting that this adversary cannot break both the classifier and detector working in tandem.
    \item In appendix \ref{sec:appendix_further_exp_bottleneck}, we quantified the bottleneck effect by varying dim($\z$) in \textbf{B}ottleneck classifiers. Results indicate that using a small bottleneck improves the robustness of the classifier against $\ell_{\infty}$ attacks.
    \item In appendix \ref{sec:appendix_further_exp_transfer}, we evaluated the transferability of adversarial examples accross different LVM-based classifiers. We found that these transferred attacks are relatively effective between generative classifiers, but not from generative to discriminative (and vice versa).
\end{itemize}

\section{Discussion}
We have proposed \emph{deep Bayes} as a generative classifier that uses deep LVMs to model the joint distribution of input-output pairs. 
We have given evidence that generative classifiers are more robust to many recent adversarial attacks than discriminative classifiers.
Furthermore, the logit in generative classifiers has a well-defined meaning and can be used to detect attacks, even when the classifier is fooled.

Concurrent to us, \citet{schott:abs2018} also demonstrated the robustness of generative classifiers on MNIST, in which their graphical model corresponds to \ref{eq:gfy} in our design, and the logits are computed by a tempered version of the variational lower-bound. However, their approach requires thousands of random $\z$ samples and tens of optimisation steps to approximate $\log p(\x|\y)$ for every input-output pair $(\x, \y)$, making it much less scalable than our importance sampling technique. Indeed, we have scaled our approach to CIFAR-10, a natural image dataset, and the robustness results are consistent with those on MNIST.

Importantly, the graphical model structure has a significant impact on robustness, which is not mentioned in \citet{schott:abs2018}. Our study shows that deep LVM-based generative classifiers generally outperform the (randomised) discriminative ones, and the bottleneck is useful for defending against $\ell_{\infty}$ attacks.
Our results corroborate with the Bayesian neural network literature, in particular \citet{li:dropout2017, feinman:detecting2017, carlini:bypass2017}, in showing that modelling unobserved variables are effective for defending against adversarial attacks.%\footnote{In Bayesian neural networks, the network weights are treated as unobserved/latent variables.} 

While we have given strong evidence to suggest that generative classifiers are more robust to current adversarial attacks, we do not wish to claim that these models will be robust to \emph{all} possible attacks. Aside from many recent attacks being designed specifically for discriminative neural networks, there is also evidence for the fragility of generative models; e.g.~naive Bayes as a standard approach for spam filtering is well-known to be fragile \citep{dalvi:adversarial2004,huang:adversarial2011}, and recently \citet{tabacof:adversarial2016,kos:adversarial2017,creswell:latentpoison2017} also designed attacks for (unconditional) VAEs.
However, generative classifiers can be made more robust too, to counter these weaknesses. \citet{dalvi:adversarial2004} have shown that generative classifiers can be made more secure if aware of the attack strategy, and \citet{biggio:robustness2011,biggio:security2014} further improved naive Bayes' robustness by modelling the conditional distribution of the adversarial inputs.
These approaches are similar to the adversarial training of discriminative classifiers \citep{tramer:ensemble2017, madry:towards2018}, efficient ways for doing so with generative classifiers can be an interesting research direction.

But even with this note of caution, we believe this work offers exciting avenues for future work. Using generative classifiers offers an interesting way to evaluate generative models and can drive improvements in their ability to tackle high-dimensional datasets, where traditionally generative classifiers have been less accurate than discriminative classifiers \citep{efron:efficiency1975, ng:discriminative2002}.
In addition, the combination of generative and discriminative models is a compelling direction for future research.
Overall, we believe that progress on generative classifiers can inspire better designs of attack, defence and detection techniques. 

\subsection*{Acknowledgements}
John Bradshaw acknowledges support from an EPSRC studentship.


\bibliography{references}
\bibliographystyle{icml2019}

\clearpage
\onecolumn
\appendix

\onecolumn

%%%%%%%%%%%%%%%%% more exps %%%%%%%%%%%%%%%%

\section{Further experiments}
\label{sec:appendix_further_exp}
% to make the counter like C.1
\setcounter{table}{0}
\setcounter{figure}{0}
\renewcommand*\thetable{\Alph{section}.\arabic{table}}
\renewcommand*\thefigure{\Alph{section}.\arabic{figure}}

\subsection{A white-box attack against both the classifier and the detection mechanism}
\label{sec:appendix_further_exp_pk}

\begin{figure}[t]
\centering
\includegraphics[width=1\linewidth]{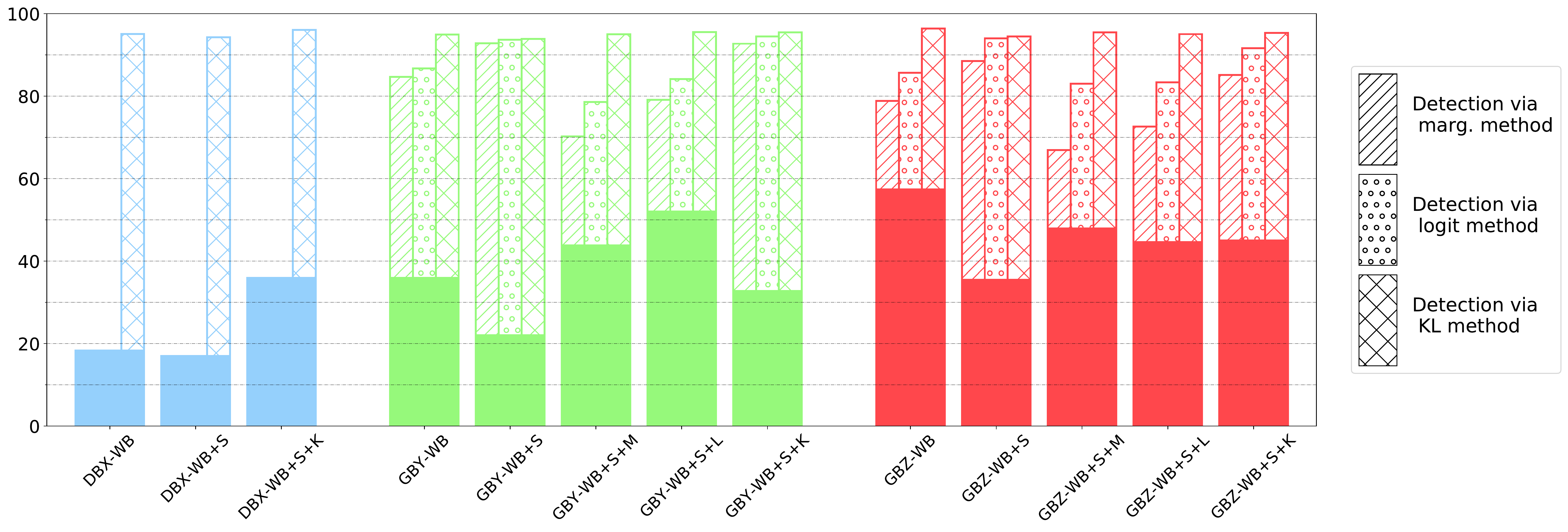}
\vspace{-0.3in}
\captionof{figure}{Accuracy and detection rates of DBX, GBY, and GBZ against PGD-based white-box (WB) attacks ($\epsilon = 0.2$, $\lambda_{\text{detect}} = 0.1$) on MNIST. The solid area denotes accuracy and the hatched area denotes the proportion of detected successful attacks with each considered detector. See text for the descriptions of the labels.
}
\vspace{-0.1in}
\label{fig:mnist_superwhite}
\end{figure}
We design a white-box attack against both the classifier and the detection mechanism, where the attacker knows everything about the victim system: it has access to the training data, can differentiate through both the classifier and the detector, and knows the usage of random $\z$ samples by the VAE-based classifiers \citep{biggio2013evasion, carlini:bypass2017}.
This PGD-based $\ell_{\infty}$ attack is designed following \citet{carlini:bypass2017}: we construct an (approximate) Bayes classifier $p_k(\y | \x)$ using (\ref{eq:deep_bayes_vae}) for each set of samples $\{\z_{c}^k\}_{c=1}^C$, and minimize the following with PGD:
\begin{equation}
\mathcal{L}(\bm{\eta}) = \sum_{k=1}^K \log p_k(\y | \x + \bm{\eta}) + \lambda_{\textrm{detect}} \max(0,  \Phi(\x + \bm{\eta}, \y) - \delta), \quad \bm{p}_k(\y | \x) = \text{softmax}_{c=1}^C \left[\log \frac{p(\x, \y = \y_c, \z^k_c)}{q(\z_c^k | \x, \y=\y_c)} \right].
\label{eq:pk-eqn}
\end{equation}
The detection statistic $\Phi(\x + \bm{\eta}, \y)$ is $-\log p(\x + \bm{\eta})$ for marginal detection,
and $\delta$ is the corresponding threshold computed on training data. For logit/KL detection, the detection statistics and thresholds are constructed accordingly. 

We refer the attack that considers the random sampling of $\z$ in the classifier only as the "white-box+sampling (WB+S)" attack, which corresponds to the case that $\lambda_{\text{detect}}=0$. When $\lambda_{\text{detect}}$ is non-zero and the marginal log probability is used as the detection statistic, the corresponding attack is labelled as "white-box+sampling+marginal detection (WB+S+M)". Similarly we also label the attacks for logit and KL detections as WB+S+L and WB+S+K, respectively. The white-box attack presented in the main text did not consider either randomness or detection, and we label this attack as WB.

Results are visualised in Figure \ref{fig:mnist_superwhite} for MNIST ($\epsilon=0.2$, $\lambda_{\text{detect}}=0.1$). 
Here we consider two metrics: the accuracy of the classifier against the attack (shown by solid bars), and the detection rates of \emph{successful attacks} (shown by hatched bars, as the absolute percentage of \emph{detected successful attacks} in all tested inputs). 
We see that although the attacker can reduce detection levels, this comes with the trade-off of increasing accuracy, suggesting that an adversary cannot break both the classifier and detector working in tandem. 

\begin{table}
\caption{WB+S+L attacks on MNIST with $\epsilon=0.2$ and $\lambda_{\text{detect}} \in \{0.0, 0.1, 1.0, 10.0 \}$ (for attacking logit detection). The $\lambda_{\text{detect}}$ values are shown in parentheses. The WB+S attack uses $\lambda_{\text{detect}}=0$. 
The white-box (WB) attack (against classifier only) results in the main text are included for reference.}
\centering
\label{tab:mnist_pk_vary_lambda}
\small
\setlength{\tabcolsep}{10pt}
\begin{tabular}{c|c|c|c|ccc}
\toprule
model & metric & WB & WB+S ($0$) & WB+S+L ($0.1$) & WB+S+L ($1.0$) & WB+S+L ($10.0$) \\
\midrule
GBZ & victim acc. & 57.4 & 35.5 & 44.6 & 71.2 & 90.5 \\
 & detect rate & 66.3 & 90.7 & 69.9 & 22.4 & 11.5 \\
\hline
GBY & victim acc. & 35.9 & 22.0 & 52.0 & 71.2 & 93.3 \\
 & detect rate & 79.3 & 91.9 & 66.9 & 22.4 & 17.2 \\
 \bottomrule
\end{tabular}
\end{table}

To further understand how the generative model's classifier and detector components interact, we tune the $\lambda_{\text{detect}}$ parameters to trade-off between the classification loss and the detection loss. With larger $\lambda_{\text{detect}}$ values the attack focuses on fooling the detection algorithm, on the other hand with small $\lambda_{\text{detect}}$ values the attack focuses on making the generative classifier predict wrong class labels. More specifically when $\lambda_{\text{detect}}=0$, the corresponding WB+S attack only focuses on fooling the classifier and thus it should achieve the highest success rate. Indeed this is shown in Table \ref{tab:mnist_pk_vary_lambda} for attacks on MNIST: WB+S achieves the lowest victim accuracy when compared with other WB+S+L attacks with non-zero $\lambda_{\text{detect}}$ values, also it out-performs the WB attack by a large margin. However the success in fooling the classifier comes with the price of increased logit detection rates: the detection rate increases when the victim accuracy decreases. Therefore these results provide evidence that the designed attack cannot break both the classifier and the detector simultaneously.

\begin{figure}
\includegraphics[width=1\linewidth]{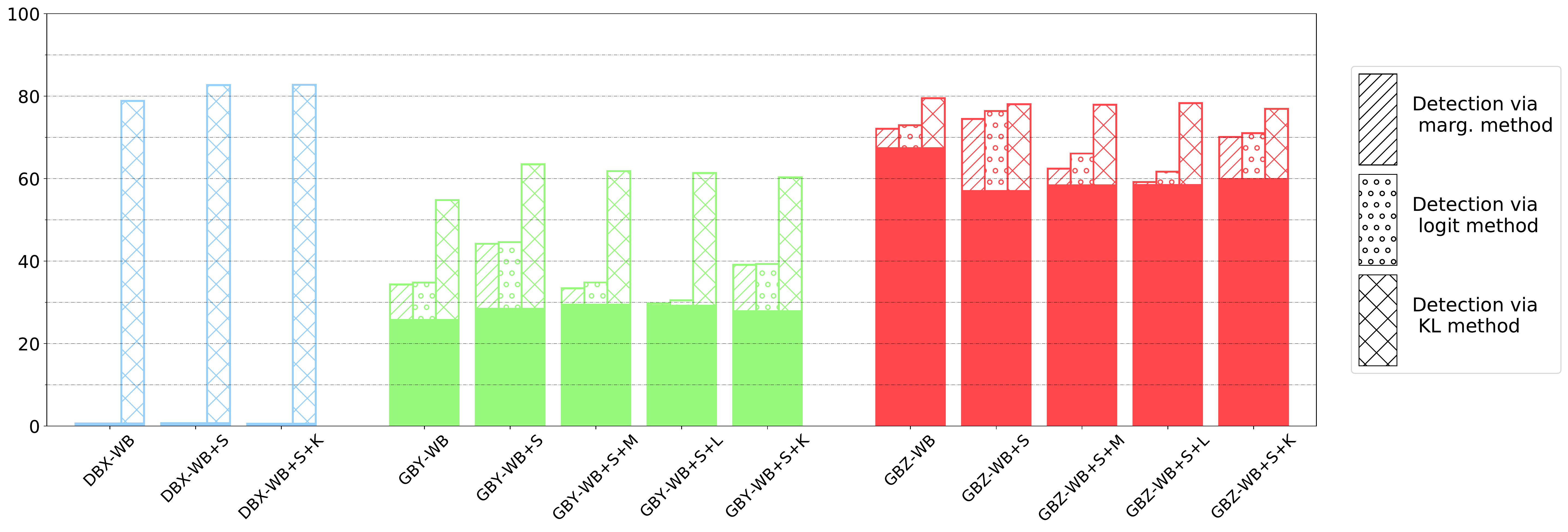}
\vspace{-0.3in}
\captionof{figure}{Accuracy and detection rates of DBX, GBY, and GBZ against PGD-based white-box (WB) attacks ($\epsilon = 0.1$, $\lambda_{\text{detect}} = 1.0$) on CIFAR binary task. The solid area denotes accuracy and the hatched area denotes the proportion of detected successful attacks with each considered detector. See text for the descriptions of the labels.
}
\label{fig:cifar_superwhite}
\end{figure}

Figure \ref{fig:cifar_superwhite} shows the WB+S+(M/L/K) attacks on CIFAR-binary ($\epsilon=0.1$, $\lambda_{\text{detect}}=1.0$). Again we see that on \ref{eq:gbz}, although the attack is effective for the detection schemes, it comes with the price of decreased mis-classification rates. Interestingly \ref{eq:gby} seems to be robust to this attack, where the accuracy on the crafted adversarial examples increase. Another surprising finding is that the attack results seem to be insensitive to the $\lambda_{\text{detect}}$ values in use. E.g.~for WB+S+L attacks we tuned $\lambda_{\text{detect}} \in \{0.1, 1.0, 10.0, 100.0, 1000.0, 10000.0 \}$, and the results are almost the same as reported in Figure \ref{fig:cifar_superwhite}: victim accuracy results are around $60\% / 30\%$ for \ref{eq:gbz}/\ref{eq:gby}, and logit detection rates are around $6\% / 1\%$. As the low detection rates indicate that the adversarial examples are close the generative model's manifold, we conjecture that the reason for this insensitivity is that the adversary has found the same ``hole'' of model density near the model manifold for all $\lambda_{\text{detect}}$ settings.
This also indicates that the generative model's manifold might not be a good approximation to the data manifold, due to the use of $\ell_2$ likelihood function in pixel space which is sub-optimal for natural images. See discussions in the main text and also section \ref{sec:appendix_further_discussion}. 

\clearpage

%%%%%%%%%% bottleneck %%%%%%%%%
\begin{table}[!h]
\caption{Clean test accuracy on MNIST classification (with varied bottleneck layer sizes).}
\centering
\label{tab:mnist_bottleneck_clean_acc}
\begin{tabular}{c|cccc}\hline 
 & $\text{dim}(\z) = 16$ & $\text{dim}(\z) = 32$ & $\text{dim}(\z) = 64$ & $\text{dim}(\z) = 128$ \\
  \hline
\ref{eq:dbx} & $99.11\%$ & $99.01\%$ & $98.98\%$ & $98.91\%$ \\
\ref{eq:gbz} & $97.11\%$ & $97.08\%$ & $97.45\%$ & $96.62\%$ \\
\ref{eq:gby} & $98.82\%$ & $98.95\%$ & $98.72\%$ & $98.75\%$ \\
\hline 
\end{tabular} 
\end{table}

\subsection{Quantifying the effect of the bottleneck layer}
\label{sec:appendix_further_exp_bottleneck}

We see from the main text that classifiers with bottleneck structure may be preferred for resisting adversarial examples. To quantify this bottleneck effect, we train on MNIST models \ref{eq:dbx}, \ref{eq:gbz} and \ref{eq:gby} with $\z$ dimensions in $\{16, 32, 64, 128\}$ (the main text experiments use $\text{dim}(\z) = 64$). The clean test accuracy is shown in Table \ref{tab:mnist_bottleneck_clean_acc}, showing that all models in test perform reasonably well.

\begin{figure*}[t]
\centering
\includegraphics[width=0.85\linewidth]{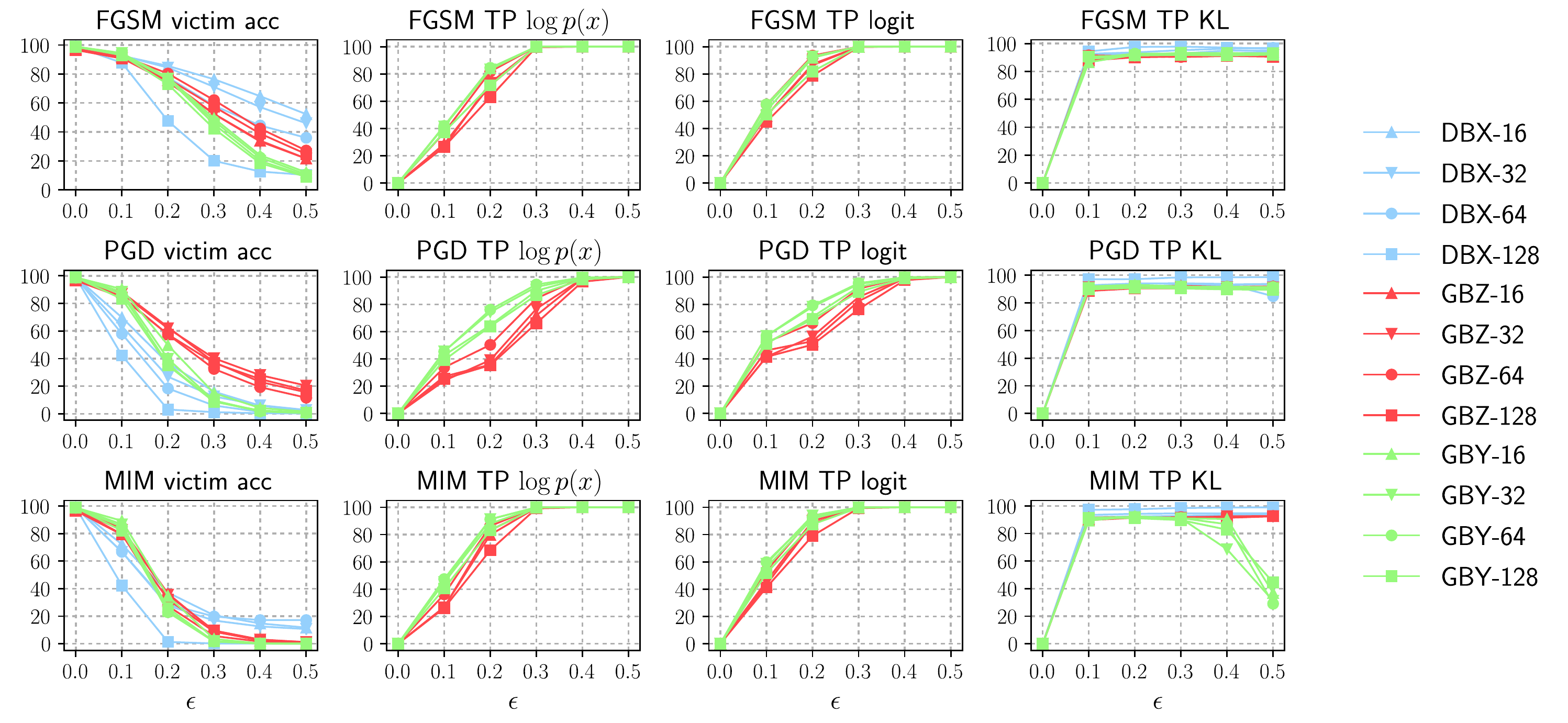}
\caption{Accuracy and detection rates against white-box $\ell_{\infty}$ attacks on MNIST, with varied bottleneck layer sizes.}
\label{fig:mnist_bottleneck}
\end{figure*}

We repeat the same white-box $\ell_{\infty}$ attack experiments as done in the main text, where results are presented in Figure \ref{fig:mnist_bottleneck}. It is clear that for discriminative classifiers, \ref{eq:dbx}, the models become less robust as the bottleneck dimension $\text{dim}(\z)$ increases. Interestingly \ref{eq:dbx} classifiers seem to be very robust against FGSM attacks, which agrees with the results in \citet{alemi:deepvib2017}. For the generative ones, we also observe similar trends (although less significant) of decreased robustness for \ref{eq:gby} classifiers, and for \ref{eq:gbz} the trend is unclear, presumably due to local optimum issues in optimisation. In summary, \ref{eq:gbz} classifiers are generally more robust compared to \ref{eq:gby} classifiers. More importantly, when the accuracy of generative classifiers on adversarial images decreases to zero, the detection rates with marginal/logit detection increases to $100\%$. This clearly shows that the three attacks tested here cannot fool the generative classifiers without being detected.

\clearpage

\subsection{Cross-model attack transferability}
\label{sec:appendix_further_exp_transfer}

\citet{papernot:transferability2016} has shown that adversarial examples transfer well between classifiers that have similar decision boundaries. Therefore we consider the cross-model transferability of the attacks crafted on generative models to discriminative classifiers (and vise versa), in order to understand whether the difference between them are significant.
Here we take adversarial examples crafted in the white-box setting with PGD and MIM on one classifier, and transfer \emph{successful} attacks to other classifiers.

\begin{figure}
\centering
\includegraphics[width=1.0\linewidth]{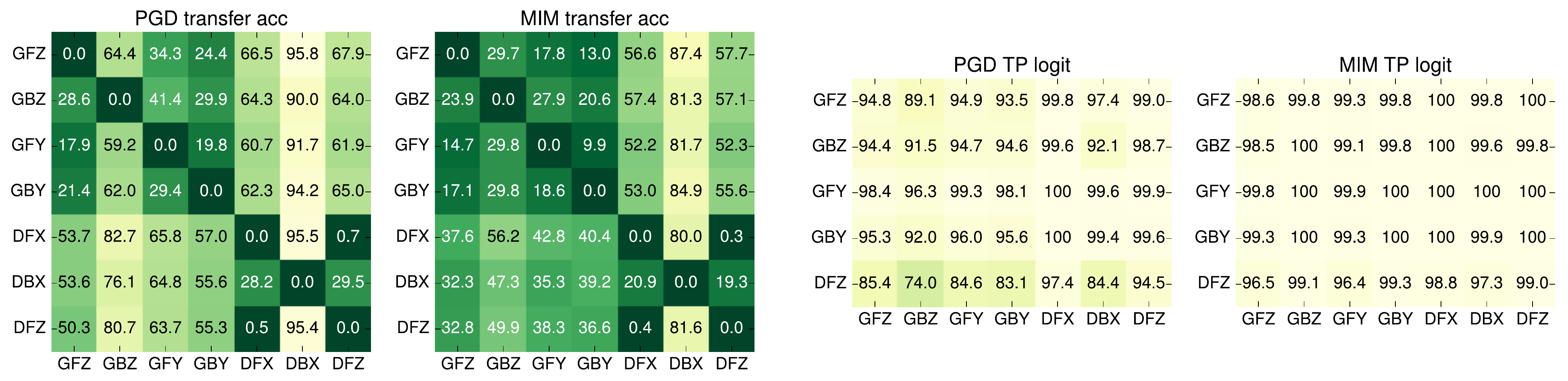}
\captionof{figure}{Results on \textbf{cross-model transfer} attacks on MNIST. Here we select $\epsilon=0.3$. The horizontal axis corresponds to the source victim that the adversarial examples are crafted on, and the vertical axis corresponds to the target victim that the attacks are transferred to. The higher (i.e.~the lighter) the better.}
\label{fig:mnist_transfer}
\vspace{0.1in}
\includegraphics[width=1.0\linewidth]{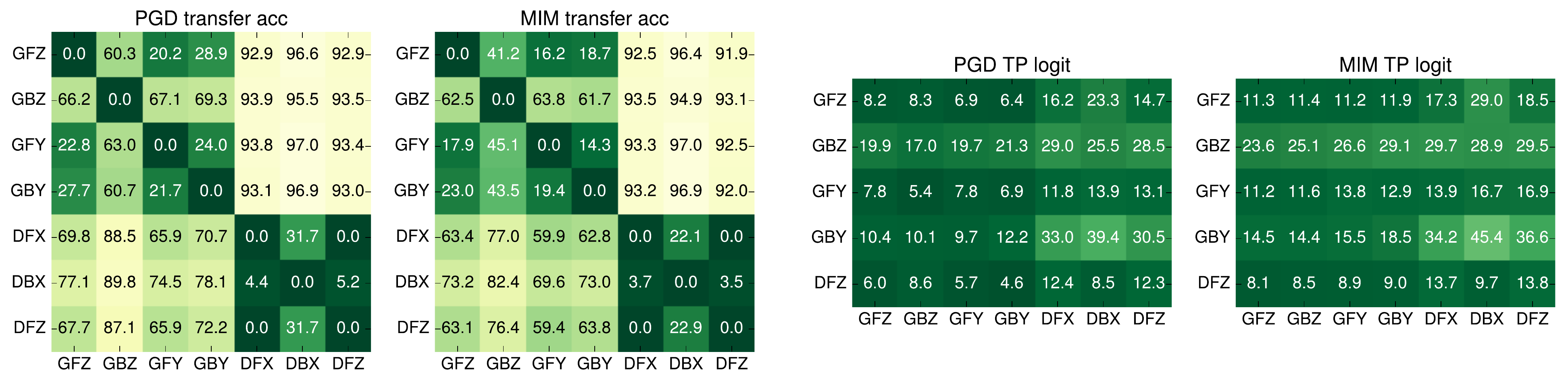}
\captionof{figure}{Results on \textbf{cross-model transfer} attacks on CIFAR-binary. Here we select $\epsilon=0.1$. The horizontal axis corresponds to the source victim that the adversarial examples are crafted on, and the vertical axis corresponds to the target victim that the attacks are transferred to. The higher (i.e.~the lighter) the better.}
\label{fig:cifar_transfer}
\end{figure}

We report in Figure \ref{fig:mnist_transfer} and Figure \ref{fig:cifar_transfer} the transferability results of the crafted adversarial examples between different models. 
We see that in both MNIST and CIFAR-binary experiments, adversarial example transfer is relatively effective between generative classifiers but not from generative to discriminative (and vice versa). Within the class of generative classifiers, \ref{eq:gbz} is the most robust one against transferred attacks. Meanwhile, \ref{eq:gbz}'s adversarial examples transfer less well to other generative classifiers. This means the decision boundary of \ref{eq:gbz} might be different from the other three generative classifiers, which potentially explains \ref{eq:gbz}'s best robustness performance in white-box attack experiments (see main text). On the other hand, the attacks crafted on \ref{eq:dbx} do not transfer in general, while at the same time, \ref{eq:dbx} is the least robust model in this case. 

For detection, the generative classifiers obtain very high detection rates on all transferred attacks on MNIST ($>95\%$). However, on CIFAR-binary the TP rates for logit detection are significantly lower than in the MNIST case, which is similar to the observations in white-box attack experiments (see Figure \ref{fig:cifar_white_box}). Nevertheless, the detection rates for the ``discriminative to generative'' transfer are considerably higher.

In summary, the transferrability test indicate that generative and discriminative classifiers are very diferrent in terms of the decision boundaries. Also generative classifiers are more robust against the tested transfer attacks across different models. 

%%%%%% additional results %%%%%%%%
\newpage

\section{Additional results for main text experiments}
\label{sec:appendix_additional_to_main}
\setcounter{table}{0}
\setcounter{figure}{0}
\renewcommand*\thetable{\Alph{section}.\arabic{table}}
\renewcommand*\thefigure{\Alph{section}.\arabic{figure}}

\begin{figure}
\centering
\includegraphics[width=0.95\linewidth]{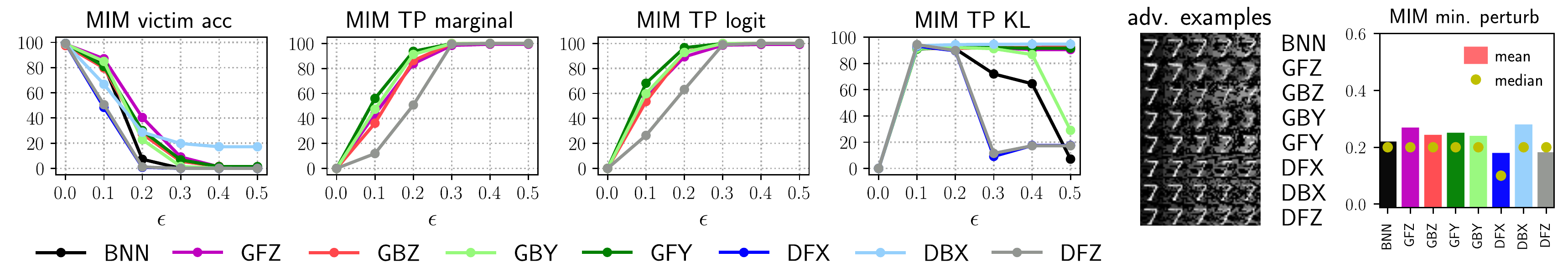}
\captionof{figure}{Victim accuracy, detection rates and minimum $\ell_{\infty}$ perturbation against \textbf{white-box MIM attack} on MNIST. The higher the better. The visualised adversarial examples are crafted with $\ell_{\infty}$ distortion $\epsilon$ growing from 0.1 to 0.5.}
\label{fig:mnist_mim_result}
\vspace{0.1in}
\includegraphics[width=0.95\linewidth]{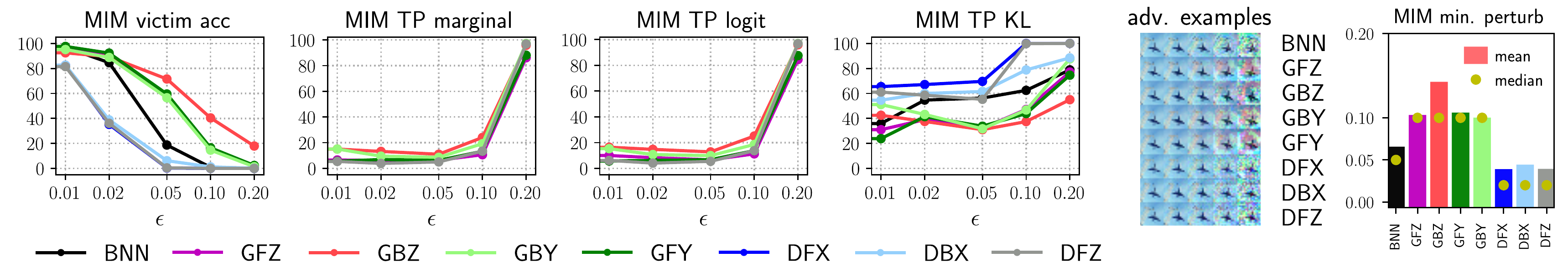}
\captionof{figure}{Victim accuracy, detection rates and minimum $\ell_{\infty}$ perturbation against \textbf{white-box MIM attack} on CIFAR-binary. The higher the better. The visualised adversarial examples are crafted with $\ell_{\infty}$ distortion $\epsilon$ growing from 0.01 to 0.2.}
\label{fig:cifar_mim_result}
\end{figure}

\subsection{White-box MIM attack results}

We visualise the white-box MIM attack results in Figure \ref{fig:mnist_mim_result} for MNIST and Figure \ref{fig:cifar_mim_result} for CIFAR-binary, respectively. Again the generative classifiers are generally more robust than the discriminative ones. Similarly the logit/marginal detection methods successfully detect the adversarial examples when $\epsilon$ increases. 

Interestingly \ref{eq:dbx} seems to be robust to MIM on MNIST (but not on CIFAR-binary). This robustness is also indicated by the minimum perturbation figures, where on MNIST the mean minimum perturbation on \ref{eq:dbx} is the highest. Since MIM is an iterative optimisation version of FGSM, this result seems to agree with FGSM results on MNIST (see main text), as well as the observations in \citet{alemi:deepvib2017}. Furthermore, a sanity check shows that MIM achieves $100\%$ success rate on \ref{eq:dbx} when $\epsilon=0.9$, therefore gradient masking is unlikely to explain the success of the bottleneck effect on MNIST classifiers \citep{athalye:obfuscated2018}. 

\subsection{Clean test accuracy on CIFAR-binary \& CIFAR-10}

We present in Table \ref{tab:plane_frog_clean_acc} the clean accuracy on CIFAR-binary test images (2000 in total).

We present in Table \ref{tab:cifar10_clean_acc} the clean accuracy for the fusion models on CIFAR-10 test images.

\begin{figure}
%\begin{table}
\captionof{table}{Clean test accuracy on CIFAR plane-vs-frog classification.}
\centering
\label{tab:plane_frog_clean_acc}
\begin{tabular}{cccccccc}\hline 
BNN & \ref{eq:gfz} & \ref{eq:gfy} & \ref{eq:dfz} & \ref{eq:dfx} & \ref{eq:dbx} & \ref{eq:gbz} & \ref{eq:gby} \\
  \hline
$97.00\%$ & $91.60\%$ & $91.20\%$ & $94.85\%$ & $95.65\%$ & $96.00\%$ & $89.35\%$ & $90.65$\\
\hline 
\end{tabular} 
%\end{table}
%
\vspace{0.1in}
%
%\begin{table}
\captionof{table}{Clean test accuracy on CIFAR-10 classification.}
\centering
\label{tab:cifar10_clean_acc}
\begin{tabular}{ccccccc}\hline 
VGG16 & \ref{eq:gbz}-FC1 & \ref{eq:gby}-FC1 & \ref{eq:dbx}-FC1 & \ref{eq:gbz}-CONV9 & \ref{eq:gby}-CONV9 & \ref{eq:dbx}-CONV9 \\
  \hline
$93.59\%$ & $92.55\%$ & $93.21\%$ & $93.49\%$ & $91.76\%$ & $88.33\%$ & $93.21\%$\\
\hline 
\end{tabular} 
%\end{table}
\end{figure}

\subsection{Full results for the fusion model experiments}
\begin{figure}
    \centering
    \includegraphics[width=0.7\linewidth]{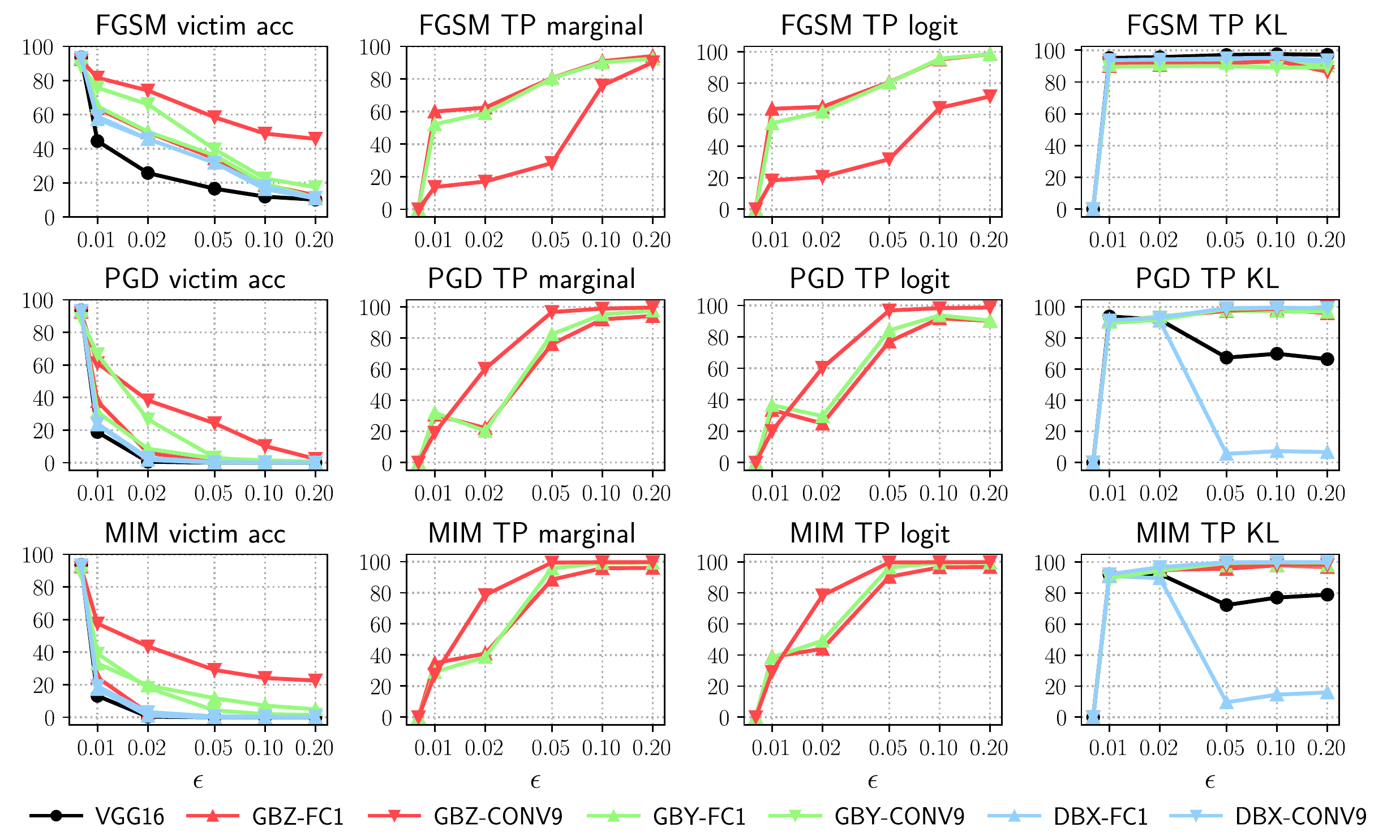}
\captionof{figure}{Victim accuracy and detection rates against \textbf{white-box attacks} on CIFAR-10. The higher the better.}
\label{fig:vgg_white_box_full}
\end{figure}

We present in \ref{fig:vgg_white_box_full} the full results of the CIFAR-10 experiments. The observations in MIM experiments are similar to those in FGSM \& PGD experiments, specifically for \ref{eq:gbz}, using \textsc{conv9} features returns significantly improved robustness results. 

\subsection{Visualising CW-$\ell_2$ adversarial examples}

\begin{figure*}
\centering
\subfigure[clean inputs]{
\includegraphics[width=0.3\linewidth]{figs/data_clean.png}}
\hspace{0.1in}
\subfigure[adv.~inputs (BNN)]{
\includegraphics[width=0.3\linewidth]{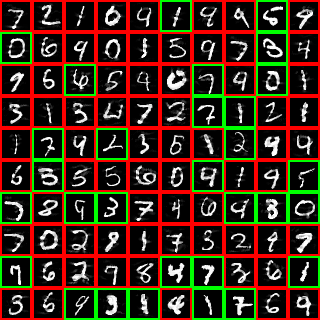}}
\hspace{0.1in}
\subfigure[adv.~inputs (\ref{eq:gfz})]{
\includegraphics[width=0.3\linewidth]{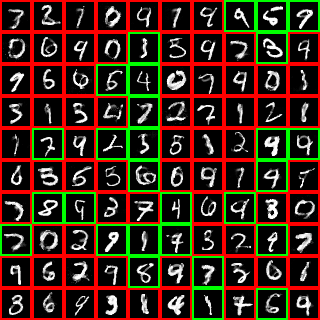}}

\subfigure[adv.~inputs (\ref{eq:gfy})]{
\includegraphics[width=0.3\linewidth]{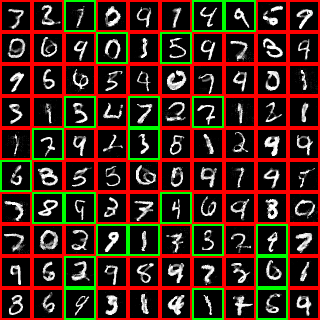}}
\hspace{0.1in}
\subfigure[adv.~inputs (\ref{eq:dfz})]{
\includegraphics[width=0.3\linewidth]{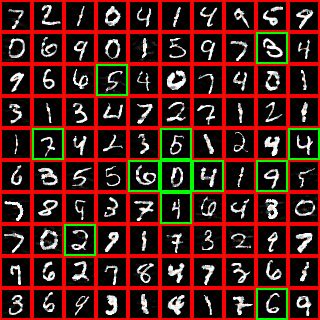}}
\hspace{0.1in}
\subfigure[adv.~inputs (\ref{eq:dfx})]{
\includegraphics[width=0.3\linewidth]{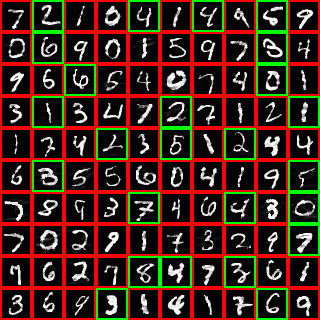}}

\subfigure[adv.~inputs (\ref{eq:dbx})]{
\includegraphics[width=0.3\linewidth]{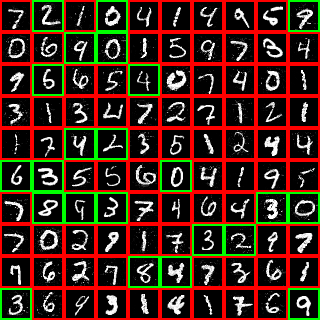}}
\hspace{0.1in}
\subfigure[adv.~inputs (\ref{eq:gbz})]{
\includegraphics[width=0.3\linewidth]{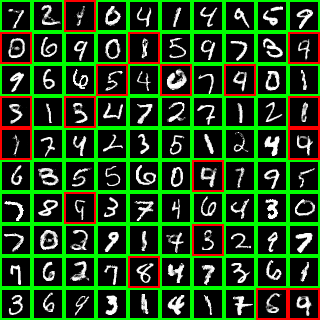}}
\hspace{0.1in}
\subfigure[adv.~inputs (\ref{eq:gby})]{
\includegraphics[width=0.3\linewidth]{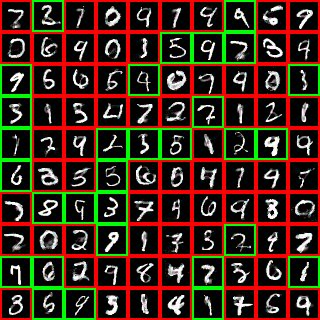}}

\vspace{-0.1in}
\caption{Visualising the clean inputs of MNIST and the CW ($c=10$) adversarial examples crafted on all the classifiers. Digits in \textbf{red} rectangles are \textbf{successful} attacks, and digits in \textbf{green} rectangles are \textbf{unsuccessful} attacks.}
\label{fig:cw_attack_appendix}
\end{figure*}

We visualise in Figure \ref{fig:cw_attack_appendix} the crafted adversarial images using white-box CW attack, where successful attacks are in red rectangles. We clearly see that many of the successful adversarial examples crafted on the generative classifiers sit at the boundary of two classes (thus ambiguous). For example, many digit ``4'' clean images are distorted to resemble digit ''9''. Similarly many digit ``1'' clean images are distorted to resemble digits ``7'' and (very thin) ``3'' and ``8''. On the other hand, we see less ambiguity from the successful attacks on discriminative classifiers. Therefore we conclude that the perceptual distortion of CW attacks on generative and discriminative classifiers are very different.

\clearpage

%%%%%%%% further discussions %%%%%%%%%

\section{Further discusssions}
\label{sec:appendix_further_discussion}

%\subsection{Comparing generative and discriminative classifiers: the adversarial sphere example}

\subsection{The fusion model for vision tasks: connections to perceptual loss}

The CIFAR-binary experiments in the main text indicate that likelihood functions based on per-pixel $\ell_2$ loss in the observation space are less suitable for modelling natural images. This observation has also been made in the deep generative models literature, and in particular, the generative adversarial network approach \citep[GAN][]{goodfellow:gan2014} can be viewed as evaluating the quality of ``perceptually realistic image generations'' using \emph{discriminative} features. Following this principle, researchers have scaled GAN-based approaches to generate high resolution images \citep{karras2018progressive}. 

Similarly, the computer vision community has discovered that ``distances'' defined on the features of a \emph{discriminatively} trained deep CNN work surprisingly well to \emph{perceptually} measure the similarity of two images. Indeed, recent research proposed the \emph{perceptual loss} \citep{dosovitskiy2016generating, johnson2016perceptual} as the $\ell_2$ distance between convolutional features extracted from a very deep CNN (e.g.~VGG):
$$\ell_{\text{perceptual}}(\x_1, \x_2) := ||\bm{\phi}(\x_1) - \bm{\phi}(\x_2) ||_2, \quad \bm{\phi}(\x) = \text{CNN-conv-layer}(\x).$$
This perceptual loss has been successfully applied to neural style transfer, super-resolution and conditional image synthesis \citep{gatys2016image, dosovitskiy2016generating,ledig2017photo,johnson2016perceptual}. 

Critically, we highlight an empirical study from \citet{zhang2018perceptual}: when comparing the ``perceptual similarity'' between two images, decisions based on the perceptual loss are well-aligned with human judgements. 
This alignment of perceptual loss to human vision also explains the success of the fusion model presented in the main text. Here the discriminative VGG features are used to construct the perceptual loss, which is used by the generative classifiers to measure the closeness of a new input $\x^*$ to the manifold of clean images (estimated from training data). By contrast, discriminative classifiers (e.g.~the original VGG classifier), when making decisions, do not explicitly take into account the ``perceptual similarity'' of the input to the images in the predicted catergory. Therefore an adversarial image containing a ``cat'' can easily fool the discriminative classifier to predict a ``dog'' class label. 

Still we note that the fusion model in practice is unlikely to be robust to \emph{all} attacks. The alignment of human vision and the perceptual loss with current deep CNNs is not perfect, therefore under the white-box setting against the whole system, an attacker might be able to craft an adversarial example that has minimum $\ell_2$, $\ell_{\infty}$ or even $\ell_0$ distortion, but at the same time has minimum ``perceptual distance'' to the image manifold of the \emph{incorrect} class. Much future work is to be done on improving representation learning for perceptual losses, as well as on investigating the adversarial robustness of the fusion model under different threat models.

\clearpage

%%%%%%%% attack & model settings %%%%%%%%

\section{Model architectures}
\paragraph{MNIST experiments} The VAEs are constructed with convolutional encoders and deconvolutional generators. More specifically, the encoder network for $q(\z | \x, \y)$ is the same across all VAE-based classifiers. It starts with a 3-layer convolutional neural network with $5 \times 5$ filters and 64 channels, with a max-pooling operation after each convolution. Then, the convolutional network is followed by a MLP with 2 hidden layers, each with 500 units, to produce the mean and variance parameters of $q$. The label $\y$ is injected into the MLP at the first hidden layer, as a one hot encoding (i.e.~for MNIST, the first hidden layer has 500+10 units). The latent dimension is $\text{dim}(\z) = 64$.

The $p$ models' architectures are the following:
\begin{itemize}
\item \ref{eq:gfz}: For $p(\y|\z)$ we use a MLP with 1 hidden layer composed of 500 units. For $p(\x | \y, \z)$ we used an MLP with 2 hidden layers, each with 500 units, and $4 \times 4 \times 64$ dimension output, followed by a 3-layer deconvolutional network with $5 \times 5$ kernel size, stride 2 and [64, 64, 1] channels.
\item \ref{eq:gfy}:  We use an MLP with 1 hidden layer composed of 500 units for $p(\z|\y)$, and the same architecture as \ref{eq:gfz} for $p(\x | \y, \z)$.
\item \ref{eq:dfz}:  We use almost the same deconvolutional network architecture for $p(\x|\z)$ as \ref{eq:gfz}'s $p(\x | \y, \z)$ network, except that the input is $\z$ only. For $p(\y|\x, \z)$ we use almost the same architecture as $q(\z | \x, \y)$ except that the injected input to the MLP is $\z$ and the MLP output is the set of logit values for $\y$.
\item \ref{eq:dfx}:  We use the same architecture as G3 for $p(\y|\x, \z)$. The network for $p(\z | \x)$ is almost identical except that there is no injected input to the MLP, and the network returns the mean and variance parameters for $q(\z | \x)$.
\item \ref{eq:dbx}: We use \ref{eq:gfz}'s architecture for $p(\y|\z)$ and \ref{eq:dfx}'s architecture for $p(\z | \x)$.
\item \ref{eq:gby}: We use \ref{eq:gfy}'s architecture for $p(\z|\y)$ and \ref{eq:dfz}'s architecture for $p(\x | \z)$.
\item \ref{eq:gbz}:  We use \ref{eq:gfz}'s architecture for $p(\y|\z)$ and \ref{eq:dfz}'s architecture for $p(\x | \z)$.
\end{itemize}

The BNN has almost the same architecture as the encoder network $q$, except that it uses 2x the hidden units/channels, and the last layer is 10 dimensions. Note that here we used dropout as it is convenient to implement, and we expect better approximate inference methods (such as stochastic gradient MCMC) to return better results for robustness and detection.

\paragraph{CIFAR-binary experiments}
The model architectures are almost the same as used in MNIST experiments, except that the hidden layer dimensions for the MLP layers are increased to 1000. For the encoder $q$, the channels are increased to [64, 128, 256]. For the $p$ models, the deconvolutional networks have different channel values, [128, 64, 3], and the MLP before the deconvolution outputs a $4 \times 4 \times 256$ vector (before reshaping). The BNN has 2x the channels but still uses 1000 hidden units.

\paragraph{CIFAR-10 experiments}
The pre-trained VGG16 network is downloaded from \url{https://github.com/geifmany/cifar-vgg}, where the \textsc{conv9} and \textsc{fc1} layers correspond to:
\begin{itemize}
    \item \textsc{conv9}: \url{https://github.com/geifmany/cifar-vgg/blob/master/cifar10vgg.py#L82}
    \item \textsc{fc1}: \url{https://github.com/geifmany/cifar-vgg/blob/master/cifar10vgg.py#L109}
\end{itemize}
The VAE-based classifiers build fully connected networks on top of the extracted features, and use $\text{dim}(\z) = 128$ for bottleneck. The encoder $q(\z | \bm{\phi}(\x), \y)$ has the network architectures [dim($\bm{\phi}(\x)$) + dim($\y$), 1000, 1000, dim($\z$) $\times$ 2], and we use the same encoder architecture across all classifiers.
The decoder architectures are as follows:
\begin{itemize}
\item \ref{eq:dbx}: We use an MLP of layers [dim($\z$), 1000, dim($\y$)] for $p(\y | \z)$ and an MLP of layers [dim($\bm{\phi}(\x)$), 1000, 1000, dim($\z$) $\times$ 2] for $p(\z | \bm{\phi}(\x))$.
\item \ref{eq:gbz}: We use an MLP of layers [dim($\z$), 1000, 1000, dim($\y$)] for $p(\y | \z)$ and an MLP of layers [dim($\z$), 1000, 1000, dim($\bm{\phi}(\x)$)] for $p(\bm{\phi}(\x) | \z)$.
\item \ref{eq:gby}: We use an MLP of layers [dim($\y$), 1000, dim($\z$) $\times$ 2] for $p(\z | \y)$ and \ref{eq:gbz}'s architecture for $p(\bm{\phi}(\x) | \z)$.
\end{itemize}

\clearpage

\section{Attack settings}
We use the Cleverhans package to perform attacks. We use the default hyper-parameters, if not specifically stated.

\textbf{PGD:} We perform the attack for 40 iterations with step-size 0.01.

\textbf{MIM:} We perform the attack for 40 iterations with step-size 0.01 and decay factor 1.0.

\textbf{CW-$\ell_2$:} We use learning rate 0.01 for $c=0.1, 1, 10$, learning rate 0.03 for $c=100$, and learning rate 0.1 for $c=1000$. We set the confidence parameter to 0, and we optimise the loss for 1000 iterations.

\textbf{SPSA:} We use almost the same hyper-parameters as in \citet{uesato:spsa2018} except for the number of samples for gradient estimates. In detail, we perform the attack for 100 iterations with perturbation size 0.01, Adam learning rate 0.01, stopping threshold -5.0 and 2000 samples for each gradient estimate.

\subsection{Jacobian-based dataset augmentation}
\label{sec:appendix_black_box_distillation_algorithm}
The black-box distillation attack is based on \citet{papernot:practical2017}, which trains a substitute CNN using Jacobian-based dataset augmentation. Assume $\y = F(\x)$ is the output one-hot vector of the victim, and $\bm{p}(\x)$ is the probability vector output of the substitute model, then at the $t^{\text{th}}$ outer-loop, we train the substitute CNN on dataset $\data_t = \{ (\x_n, \y_n) \}$ with queried $\y_n$ for 10 epochs, and augment the dataset by
\begin{equation}
\data_{t+1} = \data_t \cup \{ (\hat{\x}, F(\hat{\x})) \ | \ \hat{\x} = \x + \lambda \nabla_{\x} \bm{p}(\x)^{\text{T}} \y,  \ (\x, \y) \in \data_t \}.
\end{equation}
We initialise $\data_1$ with $200 \times 10$ datapoints from the MNIST test set, select $\lambda = 0.1$, and run the algorithm for 6 outer-loops. On MNIST, this results in $64,000$ queried inputs, and $\sim 96\%$ accuracy of the substitute model on test data. On CIFAR binary classification, we use $200 \times 2$ datapoints for the inital query set $\mathcal{D}_1$, resulting in $12,800$ queries in total. The substitutes achieved almost the same accuracy as their corresponding victim models on clean test datapoints.

%%%%%%%%%%%%%%%% tables %%%%%%%%%%%%%%%%%

\clearpage

\section{Results in tables}
\label{sec:appendix_full_results}
\setcounter{table}{0}
\setcounter{figure}{0}
\renewcommand*\thetable{\Alph{section}.\arabic{table}}
\renewcommand*\thefigure{\Alph{section}.\arabic{figure}}

We present in tables the full results of the experiments.

See Tables F.1 to F.10 for the white-box attacks.

See Tables F.11 to F.16 for the grey-box attacks.

See Tables F.17 to F.22 for the black-box attacks.

See Tables F.23 to F.25 for CIFAR-10 results with VGG-based classifiers.

See Tables F.26 to F.28 for bottleneck effect quantification results.

\begin{landscape}

% white box

% mnist
\begin{table}
\caption{FGSM white-box attack results on MNIST. }
\centering
\small
\setlength{\tabcolsep}{3pt}
\begin{tabular}{c|ccccc|ccccc|ccccc|ccccc}\hline
 & \multicolumn{5}{c|}{acc.~(adv)} & \multicolumn{5}{c|}{TP marginal} & \multicolumn{5}{c|}{TP logit} & \multicolumn{5}{c}{TP KL} \\
\hline
$\epsilon$ & 0.10 & 0.20 & 0.30 & 0.40 & 0.50 & 0.10 & 0.20 & 0.30 & 0.40 & 0.50 & 0.10 & 0.20 & 0.30 & 0.40 & 0.50 & 0.10 & 0.20 & 0.30 & 0.40 & 0.50 \\
\hline
BNN & $ 92.4 $ &$ 67.8 $ &$ 40.5 $ &$ 26.2 $ &$ 20.4 $ & N/A & N/A & N/A & N/A & N/A & N/A & N/A & N/A & N/A & N/A &$ 93.0 $ &$ 94.2 $ &$ 95.4 $ &$ 95.5 $ &$ 96.5 $  \\
GFZ & $ 94.2 $ &$ 74.5 $ &$ 38.9 $ &$ 12.9 $ &$ 5.7 $ &$ 43.6 $ &$ 79.8 $ &$ 100.0 $ &$ 100.0 $ &$ 100.0 $ &$ 56.4 $ &$ 89.6 $ &$ 99.9 $ &$ 100.0 $ &$ 100.0 $ &$ 89.2 $ &$ 91.7 $ &$ 92.2 $ &$ 92.5 $ &$ 92.2 $  \\
GBZ & $ 92.5 $ &$ 80.3 $ &$ 62.0 $ &$ 42.4 $ &$ 27.2 $ &$ 37.0 $ &$ 81.7 $ &$ 100.0 $ &$ 100.0 $ &$ 100.0 $ &$ 57.6 $ &$ 93.5 $ &$ 100.0 $ &$ 100.0 $ &$ 100.0 $ &$ 91.6 $ &$ 90.9 $ &$ 90.3 $ &$ 91.1 $ &$ 91.5 $  \\
GFY & $ 94.3 $ &$ 74.8 $ &$ 46.5 $ &$ 21.7 $ &$ 10.5 $ &$ 53.1 $ &$ 92.6 $ &$ 100.0 $ &$ 100.0 $ &$ 100.0 $ &$ 66.2 $ &$ 97.9 $ &$ 100.0 $ &$ 100.0 $ &$ 100.0 $ &$ 90.8 $ &$ 93.1 $ &$ 93.8 $ &$ 94.2 $ &$ 94.2 $  \\
GBY & $ 93.6 $ &$ 76.4 $ &$ 47.5 $ &$ 22.3 $ &$ 10.7 $ &$ 41.9 $ &$ 84.5 $ &$ 100.0 $ &$ 100.0 $ &$ 100.0 $ &$ 57.7 $ &$ 92.5 $ &$ 100.0 $ &$ 100.0 $ &$ 100.0 $ &$ 89.3 $ &$ 92.7 $ &$ 92.8 $ &$ 92.9 $ &$ 92.9 $  \\
DFX & $ 70.1 $ &$ 14.8 $ &$ 1.0 $ &$ 0.4 $ &$ 0.6 $ & N/A & N/A & N/A & N/A & N/A & N/A & N/A & N/A & N/A & N/A &$ 93.6 $ &$ 93.8 $ &$ 93.8 $ &$ 93.5 $ &$ 94.3 $  \\
DBX & $ 91.6 $ &$ 77.8 $ &$ 58.1 $ &$ 44.5 $ &$ 36.0 $ & N/A & N/A & N/A & N/A & N/A & N/A & N/A & N/A & N/A & N/A &$ 92.6 $ &$ 93.6 $ &$ 95.2 $ &$ 96.3 $ &$ 97.0 $  \\
DFZ & $ 75.2 $ &$ 19.0 $ &$ 2.4 $ &$ 1.1 $ &$ 1.0 $ &$ 12.6 $ &$ 50.6 $ &$ 100.0 $ &$ 100.0 $ &$ 100.0 $ &$ 26.9 $ &$ 65.7 $ &$ 100.0 $ &$ 100.0 $ &$ 100.0 $ &$ 93.1 $ &$ 95.1 $ &$ 94.5 $ &$ 94.9 $ &$ 94.8 $  \\
\hline
\end{tabular}
\end{table}

\begin{table}
\caption{PGD white-box attack results on MNIST. }
\centering
\small
\setlength{\tabcolsep}{3pt}
\begin{tabular}{c|ccccc|ccccc|ccccc|ccccc}\hline
 & \multicolumn{5}{c|}{acc.~(adv)} & \multicolumn{5}{c|}{TP marginal} & \multicolumn{5}{c|}{TP logit} & \multicolumn{5}{c}{TP KL} \\
\hline
$\epsilon$ & 0.10 & 0.20 & 0.30 & 0.40 & 0.50 & 0.10 & 0.20 & 0.30 & 0.40 & 0.50 & 0.10 & 0.20 & 0.30 & 0.40 & 0.50 & 0.10 & 0.20 & 0.30 & 0.40 & 0.50 \\
\hline
BNN & $ 83.2 $ &$ 12.0 $ &$ 0.5 $ &$ 0.0 $ &$ 0.0 $ & N/A & N/A & N/A & N/A & N/A & N/A & N/A & N/A & N/A & N/A &$ 92.1 $ &$ 92.1 $ &$ 88.7 $ &$ 20.4 $ &$ 0.2 $  \\
GFZ & $ 86.7 $ &$ 37.7 $ &$ 7.7 $ &$ 1.2 $ &$ 0.3 $ &$ 43.2 $ &$ 71.8 $ &$ 91.4 $ &$ 99.4 $ &$ 100.0 $ &$ 55.8 $ &$ 77.1 $ &$ 94.8 $ &$ 99.6 $ &$ 100.0 $ &$ 90.3 $ &$ 92.1 $ &$ 90.9 $ &$ 90.4 $ &$ 89.7 $  \\
GBZ & $ 85.1 $ &$ 57.4 $ &$ 32.5 $ &$ 19.3 $ &$ 11.6 $ &$ 33.7 $ &$ 50.3 $ &$ 84.4 $ &$ 99.7 $ &$ 100.0 $ &$ 52.2 $ &$ 66.3 $ &$ 91.5 $ &$ 99.8 $ &$ 100.0 $ &$ 90.0 $ &$ 91.5 $ &$ 91.5 $ &$ 91.7 $ &$ 91.8 $  \\
GFY & $ 79.7 $ &$ 27.4 $ &$ 5.6 $ &$ 1.2 $ &$ 0.3 $ &$ 58.1 $ &$ 87.9 $ &$ 98.0 $ &$ 100.0 $ &$ 100.0 $ &$ 68.7 $ &$ 92.2 $ &$ 99.3 $ &$ 100.0 $ &$ 100.0 $ &$ 92.6 $ &$ 92.5 $ &$ 90.7 $ &$ 90.7 $ &$ 85.4 $  \\
GBY & $ 86.7 $ &$ 35.9 $ &$ 9.0 $ &$ 1.7 $ &$ 0.4 $ &$ 45.3 $ &$ 76.1 $ &$ 94.6 $ &$ 99.7 $ &$ 100.0 $ &$ 56.9 $ &$ 79.3 $ &$ 95.6 $ &$ 99.9 $ &$ 100.0 $ &$ 90.1 $ &$ 92.1 $ &$ 91.6 $ &$ 91.7 $ &$ 91.4 $  \\
DFX & $ 47.6 $ &$ 0.7 $ &$ 0.0 $ &$ 0.0 $ &$ 0.0 $ & N/A & N/A & N/A & N/A & N/A & N/A & N/A & N/A & N/A & N/A &$ 92.8 $ &$ 89.8 $ &$ 13.3 $ &$ 31.4 $ &$ 42.8 $  \\
DBX & $ 58.0 $ &$ 18.3 $ &$ 6.0 $ &$ 1.3 $ &$ 0.2 $ & N/A & N/A & N/A & N/A & N/A & N/A & N/A & N/A & N/A & N/A &$ 92.7 $ &$ 94.0 $ &$ 94.1 $ &$ 93.7 $ &$ 84.8 $  \\
DFZ & $ 49.6 $ &$ 1.0 $ &$ 0.0 $ &$ 0.0 $ &$ 0.0 $ &$ 11.8 $ &$ 44.3 $ &$ 94.1 $ &$ 100.0 $ &$ 100.0 $ &$ 25.9 $ &$ 57.6 $ &$ 94.5 $ &$ 99.9 $ &$ 100.0 $ &$ 93.9 $ &$ 89.8 $ &$ 12.1 $ &$ 22.9 $ &$ 28.0 $  \\
\hline
\end{tabular}
\end{table}

\begin{table}
\caption{MIM white-box attack results on MNIST. }
\centering
\small
\setlength{\tabcolsep}{3pt}
\begin{tabular}{c|ccccc|ccccc|ccccc|ccccc}\hline
 & \multicolumn{5}{c|}{acc.~(adv)} & \multicolumn{5}{c|}{TP marginal} & \multicolumn{5}{c|}{TP logit} & \multicolumn{5}{c}{TP KL} \\
\hline
$\epsilon$ & 0.10 & 0.20 & 0.30 & 0.40 & 0.50 & 0.10 & 0.20 & 0.30 & 0.40 & 0.50 & 0.10 & 0.20 & 0.30 & 0.40 & 0.50 & 0.10 & 0.20 & 0.30 & 0.40 & 0.50 \\
\hline
BNN & $ 82.0 $ &$ 7.2 $ &$ 0.1 $ &$ 0.0 $ &$ 0.0 $ & N/A & N/A & N/A & N/A & N/A & N/A & N/A & N/A & N/A & N/A &$ 92.2 $ &$ 91.1 $ &$ 72.0 $ &$ 64.5 $ &$ 7.1 $  \\
GFZ & $ 87.0 $ &$ 40.4 $ &$ 9.0 $ &$ 1.2 $ &$ 1.2 $ &$ 43.5 $ &$ 83.8 $ &$ 98.4 $ &$ 99.4 $ &$ 99.4 $ &$ 57.6 $ &$ 89.3 $ &$ 98.6 $ &$ 99.2 $ &$ 99.2 $ &$ 91.5 $ &$ 92.3 $ &$ 92.0 $ &$ 90.6 $ &$ 90.6 $  \\
GBZ & $ 79.6 $ &$ 27.4 $ &$ 5.6 $ &$ 1.5 $ &$ 0.5 $ &$ 36.1 $ &$ 86.3 $ &$ 100.0 $ &$ 100.0 $ &$ 100.0 $ &$ 53.5 $ &$ 93.1 $ &$ 99.9 $ &$ 100.0 $ &$ 100.0 $ &$ 91.3 $ &$ 91.4 $ &$ 91.9 $ &$ 92.2 $ &$ 92.7 $  \\
GFY & $ 80.8 $ &$ 30.1 $ &$ 6.8 $ &$ 1.4 $ &$ 1.4 $ &$ 56.0 $ &$ 93.7 $ &$ 99.8 $ &$ 100.0 $ &$ 100.0 $ &$ 68.1 $ &$ 96.6 $ &$ 99.9 $ &$ 100.0 $ &$ 100.0 $ &$ 90.9 $ &$ 92.4 $ &$ 92.0 $ &$ 91.6 $ &$ 91.6 $  \\
GBY & $ 84.9 $ &$ 22.9 $ &$ 1.5 $ &$ 0.1 $ &$ 0.0 $ &$ 47.4 $ &$ 91.2 $ &$ 99.9 $ &$ 100.0 $ &$ 100.0 $ &$ 59.8 $ &$ 92.8 $ &$ 99.9 $ &$ 100.0 $ &$ 100.0 $ &$ 91.4 $ &$ 92.1 $ &$ 91.2 $ &$ 86.8 $ &$ 29.0 $  \\
DFX & $ 48.4 $ &$ 0.8 $ &$ 0.0 $ &$ 0.0 $ &$ 0.0 $ & N/A & N/A & N/A & N/A & N/A & N/A & N/A & N/A & N/A & N/A &$ 93.2 $ &$ 89.6 $ &$ 9.2 $ &$ 17.5 $ &$ 17.5 $  \\
DBX & $ 66.7 $ &$ 28.7 $ &$ 19.7 $ &$ 17.2 $ &$ 17.2 $ & N/A & N/A & N/A & N/A & N/A & N/A & N/A & N/A & N/A & N/A &$ 93.5 $ &$ 94.3 $ &$ 94.6 $ &$ 94.7 $ &$ 94.7 $  \\
DFZ & $ 50.5 $ &$ 1.2 $ &$ 0.0 $ &$ 0.0 $ &$ 0.0 $ &$ 11.9 $ &$ 50.7 $ &$ 99.4 $ &$ 100.0 $ &$ 100.0 $ &$ 26.4 $ &$ 63.2 $ &$ 99.0 $ &$ 100.0 $ &$ 100.0 $ &$ 94.3 $ &$ 89.8 $ &$ 11.5 $ &$ 17.3 $ &$ 17.3 $  \\
\hline
\end{tabular}
\end{table}

\begin{table}
\caption{CW white-box attack results on MNIST. }
\centering
\small
\setlength{\tabcolsep}{3pt}
\begin{tabular}{c|ccccc|ccccc|ccccc|ccccc}\hline
 & \multicolumn{5}{c|}{acc.~(adv)} & \multicolumn{5}{c|}{TP marginal} & \multicolumn{5}{c|}{TP logit} & \multicolumn{5}{c}{TP KL} \\
\hline
$\epsilon$ & 0.10 & 1.00 & 10.00 & 100.00 & 1000.00 & 0.10 & 1.00 & 10.00 & 100.00 & 1000.00 & 0.10 & 1.00 & 10.00 & 100.00 & 1000.00 & 0.10 & 1.00 & 10.00 & 100.00 & 1000.00 \\
\hline
BNN & $ 98.6 $ &$ 50.2 $ &$ 24.4 $ &$ 19.7 $ &$ 38.5 $ & N/A & N/A & N/A & N/A & N/A & N/A & N/A & N/A & N/A & N/A &$ 85.6 $ &$ 96.2 $ &$ 93.6 $ &$ 94.0 $ &$ 96.5 $  \\
GFZ & $ 98.7 $ &$ 76.7 $ &$ 28.6 $ &$ 20.8 $ &$ 28.4 $ &$ 45.4 $ &$ 25.1 $ &$ 11.7 $ &$ 17.9 $ &$ 49.2 $ &$ 62.0 $ &$ 39.2 $ &$ 25.3 $ &$ 29.9 $ &$ 57.4 $ &$ 91.1 $ &$ 94.9 $ &$ 95.3 $ &$ 93.5 $ &$ 94.8 $  \\
GBZ & $ 97.3 $ &$ 95.3 $ &$ 81.5 $ &$ 80.2 $ &$ 66.6 $ &$ 35.1 $ &$ 29.0 $ &$ 10.8 $ &$ 14.0 $ &$ 38.2 $ &$ 56.3 $ &$ 50.6 $ &$ 26.2 $ &$ 29.2 $ &$ 55.8 $ &$ 85.3 $ &$ 89.9 $ &$ 90.8 $ &$ 91.8 $ &$ 91.7 $  \\
GFY & $ 98.7 $ &$ 70.4 $ &$ 28.6 $ &$ 25.0 $ &$ 33.7 $ &$ 52.7 $ &$ 37.5 $ &$ 21.2 $ &$ 29.0 $ &$ 59.7 $ &$ 68.5 $ &$ 51.1 $ &$ 33.5 $ &$ 39.4 $ &$ 67.2 $ &$ 91.4 $ &$ 96.3 $ &$ 96.0 $ &$ 94.4 $ &$ 95.3 $  \\
GBY & $ 98.5 $ &$ 77.7 $ &$ 32.7 $ &$ 26.8 $ &$ 38.1 $ &$ 44.7 $ &$ 26.4 $ &$ 12.1 $ &$ 18.8 $ &$ 52.8 $ &$ 56.8 $ &$ 39.7 $ &$ 22.3 $ &$ 27.1 $ &$ 59.6 $ &$ 90.0 $ &$ 95.5 $ &$ 94.5 $ &$ 92.9 $ &$ 94.8 $  \\
DFX & $ 96.5 $ &$ 64.7 $ &$ 20.3 $ &$ 11.8 $ &$ 2.6 $ & N/A & N/A & N/A & N/A & N/A & N/A & N/A & N/A & N/A & N/A &$ 96.7 $ &$ 99.9 $ &$ 100.0 $ &$ 99.7 $ &$ 97.8 $  \\
DBX & $ 97.1 $ &$ 83.3 $ &$ 30.2 $ &$ 10.9 $ &$ 29.2 $ & N/A & N/A & N/A & N/A & N/A & N/A & N/A & N/A & N/A & N/A &$ 90.0 $ &$ 93.4 $ &$ 96.7 $ &$ 96.1 $ &$ 97.5 $  \\
DFZ & $ 95.9 $ &$ 51.2 $ &$ 13.6 $ &$ 9.6 $ &$ 16.0 $ &$ 17.3 $ &$ 7.2 $ &$ 9.2 $ &$ 17.9 $ &$ 41.6 $ &$ 30.6 $ &$ 21.5 $ &$ 22.4 $ &$ 31.9 $ &$ 54.0 $ &$ 96.2 $ &$ 99.8 $ &$ 99.5 $ &$ 98.6 $ &$ 98.2 $  \\
\hline
\end{tabular}
\end{table}

%%%%% --- MNIST perfect Knowledge 
\begin{table}
\caption{WB+S+(M/L/K) attacks on MNIST. This is done using the PGD attack with $\epsilon=0.2$ and $\lambda_{\text{detect}} = 0.1$. `WB' stands for attacks against the classifier only, `WB+S' means the adversary has knowledge of the $K$ samples but not of the detection system. `WB+S+M', `WB+S+L', and `WB+S+K' are attacks where the adversary has knowledge of the K samples and of the marginal, logit and KL detection mechanisms respectively.}
\centering
\small
\setlength{\tabcolsep}{2pt}
\begin{tabular}{c|cccc|cccc|cccc|cccc|cccc}
\hline
&\multicolumn{4}{c|}{WB}&\multicolumn{4}{c|}{WB+S}&\multicolumn{4}{c|}{WB+S+M}&\multicolumn{4}{c|}{WB+S+L}&\multicolumn{4}{c|}{WB+S+K}\\
\hline
     &   acc. & TP marg.   & TP logit   &   TP KL &   acc. & TP marg.   & TP logit   &   TP KL & acc.   & TP marg.   & TP logit   & TP KL   & acc.   & TP marg.   & TP logit   & TP KL   &   acc. & TP marg.   & TP logit   &   TP KL \\
\hline
 DBX &   18.3 & N/A        & N/A        &    94   &   17   & N/A        & N/A        &    93.1 & N/A    & N/A        & N/A        & N/A     & N/A    & N/A        & N/A        & N/A     &   35.9 & N/A        & N/A        &    93.9 \\
 GBZ &   57.4 & 50.3       & 66.3       &    91.5 &   35.5 & 82.2       & 90.7       &    91.4 & 47.9   & 36.5       & 67.4       & 91.3    & 44.6   & 50.5       & 69.9       & 91.0    &   45   & 73.0       & 84.8       &    91.5 \\
 GBY &   35.9 & 76.1       & 79.3       &    92.1 &   22   & 90.8       & 91.9       &    92.1 & 43.8   & 47.0       & 61.9       & 91.1    & 52.0   & 56.4       & 66.9       & 90.6    &   32.8 & 89.2       & 91.8       &    93.3 \\
\hline
\end{tabular}
\end{table}

% plane-vs-frog
\begin{table}
\caption{FGSM white-box attack results on CIFAR plane-vs-frog binary classification. }
\centering
\small
\setlength{\tabcolsep}{3pt}
\begin{tabular}{c|ccccc|ccccc|ccccc|ccccc}\hline
 & \multicolumn{5}{c|}{acc.~(adv)} & \multicolumn{5}{c|}{TP marginal} & \multicolumn{5}{c|}{TP logit} & \multicolumn{5}{c}{TP KL} \\
\hline
$\epsilon$ & 0.01 & 0.02 & 0.05 & 0.10 & 0.20 & 0.01 & 0.02 & 0.05 & 0.10 & 0.20 & 0.01 & 0.02 & 0.05 & 0.10 & 0.20 & 0.01 & 0.02 & 0.05 & 0.10 & 0.20 \\
\hline
BNN & $ 98.2 $ &$ 93.2 $ &$ 58.5 $ &$ 14.5 $ &$ 6.3 $ & N/A & N/A & N/A & N/A & N/A & N/A & N/A & N/A & N/A & N/A &$ 58.6 $ &$ 55.1 $ &$ 65.5 $ &$ 54.0 $ &$ 58.5 $  \\
GFZ & $ 97.1 $ &$ 94.7 $ &$ 81.8 $ &$ 56.4 $ &$ 31.4 $ &$ 11.4 $ &$ 10.1 $ &$ 5.9 $ &$ 17.4 $ &$ 99.3 $ &$ 11.4 $ &$ 10.4 $ &$ 6.8 $ &$ 17.2 $ &$ 99.3 $ &$ 58.6 $ &$ 40.1 $ &$ 35.9 $ &$ 41.0 $ &$ 50.6 $  \\
GBZ & $ 95.1 $ &$ 93.5 $ &$ 87.1 $ &$ 74.9 $ &$ 62.0 $ &$ 26.0 $ &$ 18.6 $ &$ 15.5 $ &$ 38.6 $ &$ 99.6 $ &$ 26.3 $ &$ 20.3 $ &$ 19.1 $ &$ 41.9 $ &$ 99.6 $ &$ 35.5 $ &$ 41.5 $ &$ 47.0 $ &$ 43.7 $ &$ 45.6 $  \\
GFY & $ 96.5 $ &$ 94.2 $ &$ 80.7 $ &$ 56.7 $ &$ 32.0 $ &$ 9.6 $ &$ 9.2 $ &$ 6.9 $ &$ 18.8 $ &$ 99.0 $ &$ 17.6 $ &$ 13.1 $ &$ 8.1 $ &$ 20.8 $ &$ 99.1 $ &$ 46.6 $ &$ 43.7 $ &$ 36.1 $ &$ 39.5 $ &$ 49.3 $  \\
GBY & $ 96.1 $ &$ 92.9 $ &$ 82.0 $ &$ 60.5 $ &$ 36.3 $ &$ 17.2 $ &$ 15.1 $ &$ 8.1 $ &$ 28.5 $ &$ 99.2 $ &$ 20.2 $ &$ 18.1 $ &$ 10.0 $ &$ 31.1 $ &$ 99.2 $ &$ 49.7 $ &$ 48.0 $ &$ 39.2 $ &$ 38.7 $ &$ 47.3 $  \\
DFX & $ 83.8 $ &$ 42.2 $ &$ 0.7 $ &$ 0.0 $ &$ 0.0 $ & N/A & N/A & N/A & N/A & N/A & N/A & N/A & N/A & N/A & N/A &$ 63.3 $ &$ 66.6 $ &$ 48.4 $ &$ 97.7 $ &$ 100.0 $  \\
DBX & $ 90.9 $ &$ 78.6 $ &$ 50.7 $ &$ 31.7 $ &$ 18.3 $ & N/A & N/A & N/A & N/A & N/A & N/A & N/A & N/A & N/A & N/A &$ 50.7 $ &$ 58.7 $ &$ 56.8 $ &$ 59.7 $ &$ 57.2 $  \\
DFZ & $ 83.9 $ &$ 52.1 $ &$ 2.9 $ &$ 0.0 $ &$ 0.0 $ &$ 6.6 $ &$ 3.8 $ &$ 2.2 $ &$ 3.9 $ &$ 60.6 $ &$ 6.9 $ &$ 4.2 $ &$ 2.4 $ &$ 3.3 $ &$ 60.4 $ &$ 57.5 $ &$ 62.3 $ &$ 54.8 $ &$ 66.7 $ &$ 99.8 $  \\
\hline
\end{tabular}
\end{table}

\begin{table}
\caption{PGD white-box attack results on CIFAR plane-vs-frog binary classification. }
\centering
\small
\setlength{\tabcolsep}{3pt}
\begin{tabular}{c|ccccc|ccccc|ccccc|ccccc}\hline
 & \multicolumn{5}{c|}{acc.~(adv)} & \multicolumn{5}{c|}{TP marginal} & \multicolumn{5}{c|}{TP logit} & \multicolumn{5}{c}{TP KL} \\
\hline
$\epsilon$ & 0.01 & 0.02 & 0.05 & 0.10 & 0.20 & 0.01 & 0.02 & 0.05 & 0.10 & 0.20 & 0.01 & 0.02 & 0.05 & 0.10 & 0.20 & 0.01 & 0.02 & 0.05 & 0.10 & 0.20 \\
\hline
BNN & $ 97.9 $ &$ 86.7 $ &$ 19.7 $ &$ 1.0 $ &$ 0.0 $ & N/A & N/A & N/A & N/A & N/A & N/A & N/A & N/A & N/A & N/A &$ 41.7 $ &$ 59.4 $ &$ 55.7 $ &$ 68.4 $ &$ 98.9 $  \\
GFZ & $ 98.0 $ &$ 93.9 $ &$ 67.7 $ &$ 21.7 $ &$ 3.5 $ &$ 3.3 $ &$ 6.6 $ &$ 5.6 $ &$ 7.8 $ &$ 32.9 $ &$ 5.0 $ &$ 9.5 $ &$ 6.0 $ &$ 8.2 $ &$ 33.0 $ &$ 37.5 $ &$ 45.1 $ &$ 34.5 $ &$ 44.0 $ &$ 57.7 $  \\
GBZ & $ 94.6 $ &$ 93.7 $ &$ 83.9 $ &$ 67.4 $ &$ 52.8 $ &$ 19.8 $ &$ 17.9 $ &$ 12.8 $ &$ 14.3 $ &$ 43.1 $ &$ 22.8 $ &$ 19.5 $ &$ 17.4 $ &$ 17.0 $ &$ 44.3 $ &$ 32.1 $ &$ 39.3 $ &$ 40.2 $ &$ 37.1 $ &$ 33.2 $  \\
GFY & $ 98.4 $ &$ 95.0 $ &$ 67.9 $ &$ 25.8 $ &$ 4.1 $ &$ 4.2 $ &$ 6.7 $ &$ 6.4 $ &$ 7.8 $ &$ 35.7 $ &$ 3.1 $ &$ 8.2 $ &$ 7.6 $ &$ 7.8 $ &$ 33.8 $ &$ 31.2 $ &$ 43.6 $ &$ 32.9 $ &$ 39.6 $ &$ 53.3 $  \\
GBY & $ 96.4 $ &$ 92.9 $ &$ 67.3 $ &$ 25.7 $ &$ 6.9 $ &$ 11.1 $ &$ 8.4 $ &$ 7.9 $ &$ 11.6 $ &$ 41.4 $ &$ 13.1 $ &$ 11.5 $ &$ 9.2 $ &$ 12.2 $ &$ 40.3 $ &$ 43.0 $ &$ 43.0 $ &$ 33.6 $ &$ 39.1 $ &$ 52.2 $  \\
DFX & $ 82.7 $ &$ 35.7 $ &$ 0.3 $ &$ 0.0 $ &$ 0.0 $ & N/A & N/A & N/A & N/A & N/A & N/A & N/A & N/A & N/A & N/A &$ 64.1 $ &$ 64.7 $ &$ 69.6 $ &$ 100.0 $ &$ 100.0 $  \\
DBX & $ 83.3 $ &$ 34.6 $ &$ 4.2 $ &$ 0.7 $ &$ 0.0 $ & N/A & N/A & N/A & N/A & N/A & N/A & N/A & N/A & N/A & N/A &$ 59.8 $ &$ 59.5 $ &$ 65.1 $ &$ 78.7 $ &$ 93.4 $  \\
DFZ & $ 82.4 $ &$ 36.9 $ &$ 0.4 $ &$ 0.0 $ &$ 0.0 $ &$ 4.9 $ &$ 3.8 $ &$ 5.1 $ &$ 12.3 $ &$ 91.1 $ &$ 6.5 $ &$ 4.1 $ &$ 5.5 $ &$ 12.3 $ &$ 91.3 $ &$ 56.3 $ &$ 60.3 $ &$ 54.7 $ &$ 99.7 $ &$ 100.0 $  \\
\hline
\end{tabular}
\end{table}

\begin{table}
\caption{MIM white-box attack results on CIFAR plane-vs-frog binary classification. }
\centering
\small
\setlength{\tabcolsep}{3pt}
\begin{tabular}{c|ccccc|ccccc|ccccc|ccccc}\hline
 & \multicolumn{5}{c|}{acc.~(adv)} & \multicolumn{5}{c|}{TP marginal} & \multicolumn{5}{c|}{TP logit} & \multicolumn{5}{c}{TP KL} \\
\hline
$\epsilon$ & 0.01 & 0.02 & 0.05 & 0.10 & 0.20 & 0.01 & 0.02 & 0.05 & 0.10 & 0.20 & 0.01 & 0.02 & 0.05 & 0.10 & 0.20 & 0.01 & 0.02 & 0.05 & 0.10 & 0.20 \\
\hline
BNN & $ 96.9 $ &$ 84.6 $ &$ 18.7 $ &$ 0.9 $ &$ 0.0 $ & N/A & N/A & N/A & N/A & N/A & N/A & N/A & N/A & N/A & N/A &$ 35.9 $ &$ 54.7 $ &$ 56.3 $ &$ 62.4 $ &$ 78.8 $  \\
GFZ & $ 96.9 $ &$ 91.5 $ &$ 58.1 $ &$ 15.2 $ &$ 1.9 $ &$ 6.5 $ &$ 6.2 $ &$ 6.4 $ &$ 10.6 $ &$ 86.1 $ &$ 10.0 $ &$ 8.3 $ &$ 6.6 $ &$ 11.3 $ &$ 84.5 $ &$ 31.0 $ &$ 39.0 $ &$ 32.1 $ &$ 47.2 $ &$ 77.1 $  \\
GBZ & $ 92.5 $ &$ 89.4 $ &$ 71.5 $ &$ 40.4 $ &$ 17.9 $ &$ 15.0 $ &$ 13.2 $ &$ 11.0 $ &$ 24.0 $ &$ 95.9 $ &$ 16.4 $ &$ 14.8 $ &$ 12.8 $ &$ 25.1 $ &$ 96.0 $ &$ 42.5 $ &$ 37.5 $ &$ 31.0 $ &$ 37.4 $ &$ 55.0 $  \\
GFY & $ 97.6 $ &$ 92.3 $ &$ 59.5 $ &$ 16.5 $ &$ 2.3 $ &$ 5.6 $ &$ 6.9 $ &$ 6.3 $ &$ 13.3 $ &$ 88.1 $ &$ 5.7 $ &$ 6.2 $ &$ 6.8 $ &$ 13.8 $ &$ 87.6 $ &$ 23.8 $ &$ 41.4 $ &$ 33.8 $ &$ 43.9 $ &$ 74.6 $  \\
GBY & $ 95.1 $ &$ 88.8 $ &$ 56.5 $ &$ 14.6 $ &$ 1.3 $ &$ 15.1 $ &$ 9.5 $ &$ 8.6 $ &$ 19.2 $ &$ 97.3 $ &$ 15.3 $ &$ 10.7 $ &$ 9.9 $ &$ 18.5 $ &$ 97.0 $ &$ 51.1 $ &$ 43.2 $ &$ 31.7 $ &$ 46.7 $ &$ 87.6 $  \\
DFX & $ 82.5 $ &$ 35.2 $ &$ 0.3 $ &$ 0.0 $ &$ 0.0 $ & N/A & N/A & N/A & N/A & N/A & N/A & N/A & N/A & N/A & N/A &$ 65.5 $ &$ 67.1 $ &$ 69.6 $ &$ 100.0 $ &$ 100.0 $  \\
DBX & $ 82.5 $ &$ 38.8 $ &$ 6.1 $ &$ 1.2 $ &$ 0.1 $ & N/A & N/A & N/A & N/A & N/A & N/A & N/A & N/A & N/A & N/A &$ 54.6 $ &$ 60.0 $ &$ 61.5 $ &$ 78.8 $ &$ 88.5 $  \\
DFZ & $ 81.5 $ &$ 36.1 $ &$ 0.4 $ &$ 0.0 $ &$ 0.0 $ &$ 5.4 $ &$ 3.9 $ &$ 5.0 $ &$ 13.4 $ &$ 96.9 $ &$ 6.2 $ &$ 4.1 $ &$ 5.5 $ &$ 13.8 $ &$ 96.8 $ &$ 61.1 $ &$ 58.5 $ &$ 55.3 $ &$ 99.9 $ &$ 100.0 $  \\
\hline
\end{tabular}
\end{table}

\begin{table}
\caption{CW white-box attack results on CIFAR plane-vs-frog binary classification. }
\centering
\small
\setlength{\tabcolsep}{3pt}
\begin{tabular}{c|ccccc|ccccc|ccccc|ccccc}\hline
 & \multicolumn{5}{c|}{acc.~(adv)} & \multicolumn{5}{c|}{TP marginal} & \multicolumn{5}{c|}{TP logit} & \multicolumn{5}{c}{TP KL} \\
\hline
$\epsilon$ & 0.10 & 1.00 & 10.00 & 100.00 & 1000.00 & 0.10 & 1.00 & 10.00 & 100.00 & 1000.00 & 0.10 & 1.00 & 10.00 & 100.00 & 1000.00 & 0.10 & 1.00 & 10.00 & 100.00 & 1000.00 \\
\hline
BNN & $ 98.0 $ &$ 68.9 $ &$ 38.5 $ &$ 19.5 $ &$ 9.5 $ & N/A & N/A & N/A & N/A & N/A & N/A & N/A & N/A & N/A & N/A &$ 56.2 $ &$ 79.0 $ &$ 79.5 $ &$ 71.5 $ &$ 65.3 $  \\
GFZ & $ 99.5 $ &$ 95.6 $ &$ 76.5 $ &$ 56.7 $ &$ 43.4 $ &$ 0.0 $ &$ 4.5 $ &$ 2.6 $ &$ 5.9 $ &$ 14.7 $ &$ 0.0 $ &$ 4.6 $ &$ 3.0 $ &$ 6.4 $ &$ 14.0 $ &$ 29.2 $ &$ 30.4 $ &$ 25.6 $ &$ 28.1 $ &$ 33.2 $  \\
GBZ & $ 96.0 $ &$ 93.6 $ &$ 88.9 $ &$ 80.6 $ &$ 69.8 $ &$ 16.7 $ &$ 11.5 $ &$ 7.8 $ &$ 11.7 $ &$ 33.8 $ &$ 18.9 $ &$ 13.7 $ &$ 9.3 $ &$ 13.6 $ &$ 37.5 $ &$ 41.1 $ &$ 51.4 $ &$ 40.4 $ &$ 36.0 $ &$ 37.9 $  \\
GFY & $ 99.8 $ &$ 95.9 $ &$ 78.8 $ &$ 60.6 $ &$ 42.7 $ &$ 0.0 $ &$ 6.6 $ &$ 4.1 $ &$ 5.2 $ &$ 15.9 $ &$ 0.0 $ &$ 6.6 $ &$ 4.8 $ &$ 5.7 $ &$ 15.6 $ &$ 75.0 $ &$ 37.3 $ &$ 26.0 $ &$ 28.2 $ &$ 31.5 $  \\
GBY & $ 97.5 $ &$ 93.1 $ &$ 76.7 $ &$ 61.6 $ &$ 47.7 $ &$ 15.8 $ &$ 14.6 $ &$ 6.0 $ &$ 6.4 $ &$ 19.9 $ &$ 18.9 $ &$ 13.9 $ &$ 6.0 $ &$ 7.5 $ &$ 21.3 $ &$ 42.6 $ &$ 34.3 $ &$ 27.2 $ &$ 25.2 $ &$ 33.9 $  \\
DFX & $ 82.6 $ &$ 44.2 $ &$ 34.3 $ &$ 28.6 $ &$ 4.0 $ & N/A & N/A & N/A & N/A & N/A & N/A & N/A & N/A & N/A & N/A &$ 100.0 $ &$ 100.0 $ &$ 100.0 $ &$ 99.9 $ &$ 64.6 $  \\
DBX & $ 96.5 $ &$ 72.2 $ &$ 26.3 $ &$ 12.1 $ &$ 11.5 $ & N/A & N/A & N/A & N/A & N/A & N/A & N/A & N/A & N/A & N/A &$ 61.6 $ &$ 79.4 $ &$ 85.2 $ &$ 75.2 $ &$ 70.2 $  \\
DFZ & $ 94.5 $ &$ 72.3 $ &$ 29.9 $ &$ 12.9 $ &$ 4.8 $ &$ 7.3 $ &$ 5.8 $ &$ 4.3 $ &$ 4.6 $ &$ 5.7 $ &$ 14.1 $ &$ 6.9 $ &$ 5.0 $ &$ 4.9 $ &$ 6.3 $ &$ 82.6 $ &$ 97.3 $ &$ 97.1 $ &$ 90.0 $ &$ 68.8 $  \\
\hline
\end{tabular}
\end{table}

%%%%% --- P/F perfect Knowledge 
\begin{table}
\caption{WB+S+(M/L/K) attacks on CIFAR binary task. This is done using the PGD attack with $\epsilon=0.1$ and $\lambda_{\text{detect}} = 1.0$. `WB' stands for attacks against the classifier only, `WB+S' means the adversary has knowledge of the $K$ samples but not of the detection system. `WB+S+M', `WB+S+L', and `WB+S+K' are attacks where the adversary has knowledge of the K samples and of the marginal, logit and KL detection mechanisms respectively.}
\centering
\small
\setlength{\tabcolsep}{2pt}
\begin{tabular}{c|cccc|cccc|cccc|cccc|cccc}
\hline
&\multicolumn{4}{c|}{WB}&\multicolumn{4}{c|}{WB+S}&\multicolumn{4}{c|}{WB+S+M}&\multicolumn{4}{c|}{WB+S+L}&\multicolumn{4}{c|}{WB+S+K}\\
\hline
     &   acc. & TP marg.   & TP logit   &   TP KL &   acc. & TP marg.   & TP logit   &   TP KL & acc.   & TP marg.   & TP logit   & TP KL   & acc.   & TP marg.   & TP logit   & TP KL   &   acc. & TP marg.   & TP logit   &   TP KL \\
\hline
 DBX &    0.7 & N/A        & N/A        &    78.7 &    0.7 & N/A        & N/A        &    82.5 & N/A    & N/A        & N/A        & N/A     & N/A    & N/A        & N/A        & N/A     &    0.6 & N/A        & N/A        &    82.6 \\
 GBZ &   67.4 & 14.3       & 17.0       &    37.1 &   57   & 40.6       & 45.1       &    49   & 58.4   & 9.6        & 18.5       & 46.9    & 58.5   & 1.8        & 7.8        & 47.8    &   59.9 & 25.5       & 27.6       &    42.4 \\
 GBY &   25.7 & 11.6       & 12.2       &    39.1 &   28.5 & 22.0       & 22.6       &    49   & 29.5   & 5.6        & 7.6        & 45.9    & 29.2   & 0.8        & 1.8        & 45.4    &   27.9 & 15.6       & 15.9       &    45   \\
\hline
\end{tabular}
\end{table}

% grey box

% mnist
\begin{table}
\caption{Grey-box PGD attack results on MNIST. }
\centering
\small
\setlength{\tabcolsep}{3pt}
\begin{tabular}{c|ccccc|ccccc|ccccc}\hline
 & \multicolumn{5}{c|}{substitute acc} & \multicolumn{5}{c|}{victim acc} & \multicolumn{5}{c|}{TP logit} \\
\hline
$\epsilon$ & 0.10 & 0.20 & 0.30 & 0.40 & 0.50 & 0.10 & 0.20 & 0.30 & 0.40 & 0.50 & 0.10 & 0.20 & 0.30 & 0.40 & 0.50 \\
\hline
GFZ & $ 86.7 $ &$ 14.3 $ &$ 0.0 $ &$ 0.0 $ &$ 0.0 $ &$ 96.6 $ &$ 83.3 $ &$ 51.3 $ &$ 26.8 $ &$ 17.2 $ &$ 49.5 $ &$ 73.8 $ &$ 99.7 $ &$ 100.0 $ &$ 100.0 $  \\
GBZ & $ 81.2 $ &$ 9.1 $ &$ 0.0 $ &$ 0.0 $ &$ 0.0 $ &$ 93.5 $ &$ 80.1 $ &$ 55.3 $ &$ 35.0 $ &$ 25.2 $ &$ 46.7 $ &$ 73.3 $ &$ 99.6 $ &$ 100.0 $ &$ 100.0 $  \\
GFY & $ 84.8 $ &$ 6.4 $ &$ 0.0 $ &$ 0.0 $ &$ 0.0 $ &$ 96.7 $ &$ 82.3 $ &$ 55.2 $ &$ 33.5 $ &$ 24.0 $ &$ 58.7 $ &$ 88.8 $ &$ 100.0 $ &$ 100.0 $ &$ 100.0 $  \\
GBY & $ 86.7 $ &$ 15.0 $ &$ 0.0 $ &$ 0.0 $ &$ 0.0 $ &$ 95.8 $ &$ 81.6 $ &$ 51.0 $ &$ 27.5 $ &$ 18.1 $ &$ 50.1 $ &$ 78.5 $ &$ 99.9 $ &$ 100.0 $ &$ 100.0 $  \\
DFX & $ 74.2 $ &$ 5.2 $ &$ 0.5 $ &$ 0.0 $ &$ 0.0 $ &$ 91.7 $ &$ 57.7 $ &$ 19.6 $ &$ 4.3 $ &$ 1.4 $ & N/A & N/A & N/A & N/A & N/A  \\
DBX & $ 80.6 $ &$ 5.1 $ &$ 0.0 $ &$ 0.0 $ &$ 0.0 $ &$ 93.2 $ &$ 59.2 $ &$ 22.5 $ &$ 10.9 $ &$ 8.2 $ & N/A & N/A & N/A & N/A & N/A  \\
DFZ & $ 69.6 $ &$ 3.4 $ &$ 0.3 $ &$ 0.0 $ &$ 0.0 $ &$ 91.9 $ &$ 57.2 $ &$ 21.4 $ &$ 6.3 $ &$ 2.6 $ &$ 33.6 $ &$ 55.1 $ &$ 91.7 $ &$ 100.0 $ &$ 100.0 $  \\
\hline
\end{tabular}
\end{table}

\begin{table}
\caption{Grey-box MIM attack results on MNIST. }
\centering
\small
\setlength{\tabcolsep}{3pt}
\begin{tabular}{c|ccccc|ccccc|ccccc}\hline
 & \multicolumn{5}{c|}{substitute acc} & \multicolumn{5}{c|}{victim acc} & \multicolumn{5}{c|}{TP logit} \\
\hline
$\epsilon$ & 0.10 & 0.20 & 0.30 & 0.40 & 0.50 & 0.10 & 0.20 & 0.30 & 0.40 & 0.50 & 0.10 & 0.20 & 0.30 & 0.40 & 0.50 \\
\hline
GFZ & $ 87.2 $ &$ 16.1 $ &$ 0.0 $ &$ 0.0 $ &$ 0.0 $ &$ 96.5 $ &$ 82.4 $ &$ 42.8 $ &$ 14.2 $ &$ 3.7 $ &$ 50.6 $ &$ 82.5 $ &$ 100.0 $ &$ 100.0 $ &$ 100.0 $  \\
GBZ & $ 81.8 $ &$ 9.1 $ &$ 0.0 $ &$ 0.0 $ &$ 0.0 $ &$ 93.3 $ &$ 78.6 $ &$ 45.3 $ &$ 17.0 $ &$ 5.2 $ &$ 46.2 $ &$ 81.7 $ &$ 100.0 $ &$ 100.0 $ &$ 100.0 $  \\
GFY & $ 85.2 $ &$ 6.9 $ &$ 0.0 $ &$ 0.0 $ &$ 0.0 $ &$ 96.6 $ &$ 80.8 $ &$ 48.0 $ &$ 18.9 $ &$ 7.4 $ &$ 59.8 $ &$ 96.0 $ &$ 100.0 $ &$ 100.0 $ &$ 100.0 $  \\
GBY & $ 87.0 $ &$ 15.9 $ &$ 0.0 $ &$ 0.0 $ &$ 0.0 $ &$ 95.7 $ &$ 79.9 $ &$ 41.8 $ &$ 13.0 $ &$ 4.0 $ &$ 51.4 $ &$ 87.1 $ &$ 100.0 $ &$ 100.0 $ &$ 100.0 $  \\
DFX & $ 76.5 $ &$ 12.3 $ &$ 3.6 $ &$ 2.1 $ &$ 2.1 $ &$ 91.5 $ &$ 56.1 $ &$ 22.2 $ &$ 9.9 $ &$ 9.9 $ & N/A & N/A & N/A & N/A & N/A  \\
DBX & $ 85.1 $ &$ 19.9 $ &$ 0.6 $ &$ 0.1 $ &$ 0.1 $ &$ 93.4 $ &$ 57.0 $ &$ 14.1 $ &$ 6.3 $ &$ 6.3 $ & N/A & N/A & N/A & N/A & N/A  \\
DFZ & $ 70.9 $ &$ 3.2 $ &$ 0.0 $ &$ 0.0 $ &$ 0.0 $ &$ 91.8 $ &$ 54.6 $ &$ 16.4 $ &$ 2.7 $ &$ 0.3 $ &$ 34.5 $ &$ 63.8 $ &$ 98.9 $ &$ 100.0 $ &$ 100.0 $  \\
\hline
\end{tabular}
\end{table}

\begin{table}
\caption{Grey-box CW attack results on MNIST. }
\centering
\small
\setlength{\tabcolsep}{3pt}
\begin{tabular}{c|ccccc|ccccc|ccccc}\hline
 & \multicolumn{5}{c|}{substitute acc} & \multicolumn{5}{c|}{victim acc} & \multicolumn{5}{c|}{TP logit} \\
\hline
$\epsilon$ & 0.10 & 1.00 & 10.00 & 100.00 & 1000.00 & 0.10 & 1.00 & 10.00 & 100.00 & 1000.00 & 0.10 & 1.00 & 10.00 & 100.00 & 1000.00 \\
\hline
GFZ & $ 98.4 $ &$ 4.6 $ &$ 0.0 $ &$ 0.0 $ &$ 0.0 $ &$ 98.8 $ &$ 96.9 $ &$ 96.0 $ &$ 93.4 $ &$ 90.2 $ &$ 53.4 $ &$ 44.2 $ &$ 41.6 $ &$ 40.6 $ &$ 43.7 $  \\
GBZ & $ 98.4 $ &$ 63.2 $ &$ 0.0 $ &$ 0.0 $ &$ 0.0 $ &$ 97.4 $ &$ 95.4 $ &$ 92.9 $ &$ 88.6 $ &$ 84.5 $ &$ 45.9 $ &$ 39.7 $ &$ 33.0 $ &$ 31.2 $ &$ 38.6 $  \\
GFY & $ 98.1 $ &$ 0.9 $ &$ 0.0 $ &$ 0.0 $ &$ 0.0 $ &$ 99.0 $ &$ 97.6 $ &$ 97.0 $ &$ 95.7 $ &$ 94.0 $ &$ 61.0 $ &$ 55.2 $ &$ 49.2 $ &$ 49.5 $ &$ 52.7 $  \\
GBY & $ 98.3 $ &$ 5.3 $ &$ 0.0 $ &$ 0.0 $ &$ 0.0 $ &$ 98.7 $ &$ 96.8 $ &$ 96.0 $ &$ 93.8 $ &$ 91.2 $ &$ 54.2 $ &$ 41.5 $ &$ 37.6 $ &$ 38.7 $ &$ 43.6 $  \\
DFX & $ 85.1 $ &$ 0.0 $ &$ 0.0 $ &$ 0.0 $ &$ 0.0 $ &$ 97.6 $ &$ 93.3 $ &$ 91.2 $ &$ 90.4 $ &$ 89.8 $ & N/A & N/A & N/A & N/A & N/A  \\
DBX & $ 97.3 $ &$ 46.0 $ &$ 0.4 $ &$ 0.0 $ &$ 0.0 $ &$ 98.7 $ &$ 93.9 $ &$ 88.4 $ &$ 85.5 $ &$ 76.3 $ & N/A & N/A & N/A & N/A & N/A  \\
DFZ & $ 80.5 $ &$ 0.0 $ &$ 0.0 $ &$ 0.0 $ &$ 0.0 $ &$ 97.2 $ &$ 94.6 $ &$ 92.3 $ &$ 91.5 $ &$ 91.2 $ &$ 32.9 $ &$ 28.6 $ &$ 28.1 $ &$ 36.5 $ &$ 44.5 $  \\
\hline
\end{tabular}
\end{table}

% plane-vs-frog
\begin{table}
\caption{Grey-box PGD attack results on CIFAR plane-vs-frog binary classification. }
\centering
\small
\setlength{\tabcolsep}{3pt}
\begin{tabular}{c|ccccc|ccccc|ccccc}\hline
 & \multicolumn{5}{c|}{substitute acc} & \multicolumn{5}{c|}{victim acc} & \multicolumn{5}{c|}{TP logit} \\
\hline
$\epsilon$ & 0.01 & 0.02 & 0.05 & 0.10 & 0.20 & 0.01 & 0.02 & 0.05 & 0.10 & 0.20 & 0.01 & 0.02 & 0.05 & 0.10 & 0.20 \\
\hline
GFZ & $ 96.7 $ &$ 88.1 $ &$ 41.7 $ &$ 4.5 $ &$ 0.1 $ &$ 97.0 $ &$ 94.9 $ &$ 84.0 $ &$ 48.3 $ &$ 8.3 $ &$ 8.4 $ &$ 11.2 $ &$ 6.1 $ &$ 7.5 $ &$ 49.8 $  \\
GBZ & $ 92.9 $ &$ 83.1 $ &$ 37.8 $ &$ 5.5 $ &$ 0.0 $ &$ 95.3 $ &$ 94.8 $ &$ 91.8 $ &$ 83.1 $ &$ 65.6 $ &$ 15.5 $ &$ 12.8 $ &$ 10.9 $ &$ 19.9 $ &$ 79.8 $  \\
GFY & $ 96.9 $ &$ 88.1 $ &$ 33.7 $ &$ 3.3 $ &$ 0.1 $ &$ 96.9 $ &$ 95.9 $ &$ 85.3 $ &$ 54.2 $ &$ 12.7 $ &$ 2.6 $ &$ 2.0 $ &$ 5.0 $ &$ 8.5 $ &$ 61.2 $  \\
GBY & $ 95.2 $ &$ 85.6 $ &$ 32.4 $ &$ 2.8 $ &$ 0.1 $ &$ 96.8 $ &$ 95.3 $ &$ 86.4 $ &$ 59.8 $ &$ 18.5 $ &$ 21.8 $ &$ 14.3 $ &$ 10.5 $ &$ 12.5 $ &$ 75.2 $  \\
DFX & $ 95.7 $ &$ 80.8 $ &$ 10.1 $ &$ 0.2 $ &$ 0.0 $ &$ 99.1 $ &$ 96.7 $ &$ 79.4 $ &$ 28.3 $ &$ 0.9 $ & N/A & N/A & N/A & N/A & N/A  \\
DBX & $ 95.7 $ &$ 80.5 $ &$ 30.5 $ &$ 1.7 $ &$ 0.0 $ &$ 99.5 $ &$ 98.2 $ &$ 91.1 $ &$ 65.1 $ &$ 21.8 $ & N/A & N/A & N/A & N/A & N/A  \\
DFZ & $ 96.5 $ &$ 81.9 $ &$ 10.5 $ &$ 0.3 $ &$ 0.0 $ &$ 99.1 $ &$ 96.9 $ &$ 76.9 $ &$ 26.2 $ &$ 1.3 $ &$ 4.5 $ &$ 16.5 $ &$ 6.1 $ &$ 8.5 $ &$ 59.6 $  \\
\hline
\end{tabular}
\end{table}

\begin{table}
\caption{Grey-box MIM attack results on CIFAR plane-vs-frog binary classification. }
\centering
\small
\setlength{\tabcolsep}{3pt}
\begin{tabular}{c|ccccc|ccccc|ccccc}\hline
 & \multicolumn{5}{c|}{substitute acc} & \multicolumn{5}{c|}{victim acc} & \multicolumn{5}{c|}{TP logit} \\
\hline
$\epsilon$ & 0.01 & 0.02 & 0.05 & 0.10 & 0.20 & 0.01 & 0.02 & 0.05 & 0.10 & 0.20 & 0.01 & 0.02 & 0.05 & 0.10 & 0.20 \\
\hline
GFZ & $ 96.3 $ &$ 87.9 $ &$ 42.1 $ &$ 4.7 $ &$ 0.1 $ &$ 97.1 $ &$ 94.8 $ &$ 83.0 $ &$ 44.8 $ &$ 4.9 $ &$ 8.7 $ &$ 11.1 $ &$ 5.8 $ &$ 8.5 $ &$ 75.1 $  \\
GBZ & $ 92.9 $ &$ 82.5 $ &$ 36.9 $ &$ 4.9 $ &$ 0.0 $ &$ 95.5 $ &$ 94.7 $ &$ 91.4 $ &$ 81.7 $ &$ 49.5 $ &$ 14.8 $ &$ 12.6 $ &$ 13.9 $ &$ 21.9 $ &$ 97.8 $  \\
GFY & $ 96.8 $ &$ 87.9 $ &$ 34.1 $ &$ 3.5 $ &$ 0.1 $ &$ 96.9 $ &$ 95.6 $ &$ 84.6 $ &$ 49.1 $ &$ 6.9 $ &$ 2.6 $ &$ 2.0 $ &$ 6.7 $ &$ 10.3 $ &$ 81.1 $  \\
GBY & $ 95.1 $ &$ 85.5 $ &$ 32.0 $ &$ 2.6 $ &$ 0.1 $ &$ 96.8 $ &$ 95.5 $ &$ 85.7 $ &$ 56.2 $ &$ 10.9 $ &$ 21.8 $ &$ 14.8 $ &$ 9.2 $ &$ 13.4 $ &$ 97.6 $  \\
DFX & $ 95.5 $ &$ 80.5 $ &$ 10.5 $ &$ 0.2 $ &$ 0.0 $ &$ 99.1 $ &$ 96.5 $ &$ 77.3 $ &$ 24.0 $ &$ 0.2 $ & N/A & N/A & N/A & N/A & N/A  \\
DBX & $ 95.6 $ &$ 81.1 $ &$ 34.5 $ &$ 2.2 $ &$ 0.0 $ &$ 99.5 $ &$ 98.3 $ &$ 89.3 $ &$ 58.9 $ &$ 9.9 $ & N/A & N/A & N/A & N/A & N/A  \\
DFZ & $ 96.3 $ &$ 81.5 $ &$ 10.7 $ &$ 0.5 $ &$ 0.0 $ &$ 99.1 $ &$ 96.9 $ &$ 74.9 $ &$ 21.9 $ &$ 0.3 $ &$ 4.5 $ &$ 16.6 $ &$ 6.4 $ &$ 9.9 $ &$ 80.2 $  \\
\hline
\end{tabular}
\end{table}

\begin{table}
\caption{Grey-box CW attack results on CIFAR plane-vs-frog binary classification. }
\centering
\small
\setlength{\tabcolsep}{3pt}
\begin{tabular}{c|ccccc|ccccc|ccccc}\hline
 & \multicolumn{5}{c|}{substitute acc} & \multicolumn{5}{c|}{victim acc} & \multicolumn{5}{c|}{TP logit} \\
\hline
$\epsilon$ & 0.10 & 1.00 & 10.00 & 100.00 & 1000.00 & 0.10 & 1.00 & 10.00 & 100.00 & 1000.00 & 0.10 & 1.00 & 10.00 & 100.00 & 1000.00 \\
\hline
GFZ & $ 98.7 $ &$ 61.9 $ &$ 0.0 $ &$ 0.0 $ &$ 0.0 $ &$ 98.6 $ &$ 95.8 $ &$ 94.3 $ &$ 93.8 $ &$ 93.2 $ &$ 11.7 $ &$ 10.2 $ &$ 10.9 $ &$ 9.9 $ &$ 8.5 $  \\
GBZ & $ 96.3 $ &$ 69.8 $ &$ 3.3 $ &$ 0.0 $ &$ 0.0 $ &$ 95.9 $ &$ 95.7 $ &$ 94.7 $ &$ 94.7 $ &$ 94.5 $ &$ 13.5 $ &$ 12.7 $ &$ 12.3 $ &$ 13.7 $ &$ 14.4 $  \\
GFY & $ 98.4 $ &$ 54.4 $ &$ 0.0 $ &$ 0.0 $ &$ 0.0 $ &$ 98.1 $ &$ 96.1 $ &$ 95.3 $ &$ 95.1 $ &$ 94.3 $ &$ 0.0 $ &$ 2.2 $ &$ 3.7 $ &$ 4.6 $ &$ 7.6 $  \\
GBY & $ 97.6 $ &$ 52.9 $ &$ 2.2 $ &$ 0.0 $ &$ 0.0 $ &$ 97.7 $ &$ 95.6 $ &$ 94.5 $ &$ 94.3 $ &$ 93.7 $ &$ 25.8 $ &$ 15.7 $ &$ 14.2 $ &$ 13.7 $ &$ 18.0 $  \\
DFX & $ 64.4 $ &$ 0.0 $ &$ 0.0 $ &$ 0.0 $ &$ 0.0 $ &$ 98.5 $ &$ 97.2 $ &$ 97.1 $ &$ 96.7 $ &$ 95.5 $ & N/A & N/A & N/A & N/A & N/A  \\
DBX & $ 91.3 $ &$ 56.2 $ &$ 0.3 $ &$ 0.0 $ &$ 0.0 $ &$ 99.7 $ &$ 99.0 $ &$ 99.0 $ &$ 98.9 $ &$ 98.1 $ & N/A & N/A & N/A & N/A & N/A  \\
DFZ & $ 80.3 $ &$ 0.0 $ &$ 0.0 $ &$ 0.0 $ &$ 0.0 $ &$ 99.2 $ &$ 97.9 $ &$ 97.5 $ &$ 97.3 $ &$ 95.4 $ &$ 0.0 $ &$ 15.7 $ &$ 14.6 $ &$ 8.8 $ &$ 6.1 $  \\
\hline
\end{tabular}
\end{table}

% black box

% mnist
\begin{table}
\caption{Black-box PGD attack results on MNIST. }
\centering
\small
\setlength{\tabcolsep}{3pt}
\begin{tabular}{c|ccccc|ccccc|ccccc}\hline
 & \multicolumn{5}{c|}{substitute acc} & \multicolumn{5}{c|}{victim acc} & \multicolumn{5}{c|}{TP logit} \\
\hline
$\epsilon$ & 0.10 & 0.20 & 0.30 & 0.40 & 0.50 & 0.10 & 0.20 & 0.30 & 0.40 & 0.50 & 0.10 & 0.20 & 0.30 & 0.40 & 0.50 \\
\hline
GFZ & $ 44.2 $ &$ 0.7 $ &$ 0.0 $ &$ 0.0 $ &$ 0.0 $ &$ 97.3 $ &$ 88.4 $ &$ 62.7 $ &$ 33.8 $ &$ 21.8 $ &$ 48.7 $ &$ 61.0 $ &$ 92.3 $ &$ 99.8 $ &$ 100.0 $  \\
GBZ & $ 7.4 $ &$ 0.0 $ &$ 0.0 $ &$ 0.0 $ &$ 0.0 $ &$ 94.8 $ &$ 87.1 $ &$ 70.7 $ &$ 52.9 $ &$ 41.3 $ &$ 51.0 $ &$ 68.6 $ &$ 97.7 $ &$ 100.0 $ &$ 100.0 $  \\
GFY & $ 49.4 $ &$ 0.8 $ &$ 0.0 $ &$ 0.0 $ &$ 0.0 $ &$ 97.4 $ &$ 86.5 $ &$ 58.1 $ &$ 31.9 $ &$ 21.0 $ &$ 52.5 $ &$ 70.8 $ &$ 97.4 $ &$ 99.8 $ &$ 100.0 $  \\
GBY & $ 21.7 $ &$ 0.0 $ &$ 0.0 $ &$ 0.0 $ &$ 0.0 $ &$ 96.9 $ &$ 89.3 $ &$ 70.0 $ &$ 49.7 $ &$ 37.8 $ &$ 51.2 $ &$ 70.8 $ &$ 98.2 $ &$ 100.0 $ &$ 100.0 $  \\
DFX & $ 49.2 $ &$ 1.3 $ &$ 0.0 $ &$ 0.0 $ &$ 0.0 $ &$ 91.4 $ &$ 52.1 $ &$ 13.9 $ &$ 2.2 $ &$ 0.7 $ & N/A & N/A & N/A & N/A & N/A  \\
DBX & $ 43.1 $ &$ 0.7 $ &$ 0.0 $ &$ 0.0 $ &$ 0.0 $ &$ 95.0 $ &$ 67.0 $ &$ 26.2 $ &$ 9.7 $ &$ 6.5 $ & N/A & N/A & N/A & N/A & N/A  \\
DFZ & $ 53.8 $ &$ 1.7 $ &$ 0.0 $ &$ 0.0 $ &$ 0.0 $ &$ 92.9 $ &$ 56.3 $ &$ 15.2 $ &$ 3.4 $ &$ 1.5 $ &$ 34.7 $ &$ 52.8 $ &$ 94.0 $ &$ 99.9 $ &$ 100.0 $  \\
\hline
\end{tabular}
\end{table}

\begin{table}
\caption{Black-box MIM attack results on MNIST. }
\centering
\small
\setlength{\tabcolsep}{3pt}
\begin{tabular}{c|ccccc|ccccc|ccccc}\hline
 & \multicolumn{5}{c|}{substitute acc} & \multicolumn{5}{c|}{victim acc} & \multicolumn{5}{c|}{TP logit} \\
\hline
$\epsilon$ & 0.10 & 0.20 & 0.30 & 0.40 & 0.50 & 0.10 & 0.20 & 0.30 & 0.40 & 0.50 & 0.10 & 0.20 & 0.30 & 0.40 & 0.50 \\
\hline
GFZ & $ 45.5 $ &$ 2.8 $ &$ 0.0 $ &$ 0.0 $ &$ 0.0 $ &$ 97.2 $ &$ 86.6 $ &$ 50.3 $ &$ 17.0 $ &$ 17.0 $ &$ 51.1 $ &$ 68.0 $ &$ 97.5 $ &$ 99.5 $ &$ 99.5 $  \\
GBZ & $ 8.4 $ &$ 0.0 $ &$ 0.0 $ &$ 0.0 $ &$ 0.0 $ &$ 94.7 $ &$ 85.9 $ &$ 63.5 $ &$ 35.2 $ &$ 17.1 $ &$ 51.2 $ &$ 74.4 $ &$ 99.0 $ &$ 100.0 $ &$ 100.0 $  \\
GFY & $ 53.1 $ &$ 2.5 $ &$ 0.3 $ &$ 0.1 $ &$ 0.1 $ &$ 97.2 $ &$ 83.7 $ &$ 49.2 $ &$ 20.1 $ &$ 20.1 $ &$ 55.1 $ &$ 80.6 $ &$ 99.0 $ &$ 99.9 $ &$ 99.9 $  \\
GBY & $ 24.5 $ &$ 0.0 $ &$ 0.0 $ &$ 0.0 $ &$ 0.0 $ &$ 96.8 $ &$ 88.3 $ &$ 62.8 $ &$ 31.7 $ &$ 13.3 $ &$ 49.3 $ &$ 77.6 $ &$ 99.9 $ &$ 100.0 $ &$ 100.0 $  \\
DFX & $ 51.3 $ &$ 2.3 $ &$ 0.1 $ &$ 0.0 $ &$ 0.0 $ &$ 91.4 $ &$ 51.7 $ &$ 13.3 $ &$ 1.9 $ &$ 1.9 $ & N/A & N/A & N/A & N/A & N/A  \\
DBX & $ 47.8 $ &$ 1.6 $ &$ 0.1 $ &$ 0.0 $ &$ 0.0 $ &$ 94.8 $ &$ 67.0 $ &$ 23.2 $ &$ 7.6 $ &$ 7.6 $ & N/A & N/A & N/A & N/A & N/A  \\
DFZ & $ 56.1 $ &$ 2.9 $ &$ 0.1 $ &$ 0.0 $ &$ 0.0 $ &$ 92.7 $ &$ 56.0 $ &$ 13.1 $ &$ 2.8 $ &$ 2.8 $ &$ 35.2 $ &$ 59.9 $ &$ 97.6 $ &$ 100.0 $ &$ 100.0 $  \\
\hline
\end{tabular}
\end{table}

\begin{table}
\caption{Black-box CW attack results on MNIST. }
\centering
\small
\setlength{\tabcolsep}{3pt}
\begin{tabular}{c|ccccc|ccccc|ccccc}\hline
 & \multicolumn{5}{c|}{substitute acc} & \multicolumn{5}{c|}{victim acc} & \multicolumn{5}{c|}{TP logit} \\
\hline
$\epsilon$ & 0.10 & 1.00 & 10.00 & 100.00 & 1000.00 & 0.10 & 1.00 & 10.00 & 100.00 & 1000.00 & 0.10 & 1.00 & 10.00 & 100.00 & 1000.00 \\
\hline
GFZ & $ 65.2 $ &$ 0.3 $ &$ 0.0 $ &$ 0.0 $ &$ 0.0 $ &$ 98.8 $ &$ 98.7 $ &$ 97.4 $ &$ 94.3 $ &$ 92.0 $ &$ 52.3 $ &$ 50.8 $ &$ 39.6 $ &$ 44.7 $ &$ 55.3 $  \\
GBZ & $ 76.6 $ &$ 0.4 $ &$ 0.0 $ &$ 0.0 $ &$ 0.0 $ &$ 97.3 $ &$ 97.2 $ &$ 95.9 $ &$ 92.7 $ &$ 90.4 $ &$ 45.7 $ &$ 46.1 $ &$ 41.5 $ &$ 40.3 $ &$ 45.7 $  \\
GFY & $ 88.4 $ &$ 3.8 $ &$ 0.0 $ &$ 0.0 $ &$ 0.0 $ &$ 99.0 $ &$ 98.9 $ &$ 98.0 $ &$ 94.8 $ &$ 92.0 $ &$ 60.0 $ &$ 59.2 $ &$ 51.9 $ &$ 47.6 $ &$ 57.2 $  \\
GBY & $ 76.0 $ &$ 1.3 $ &$ 0.0 $ &$ 0.0 $ &$ 0.0 $ &$ 98.7 $ &$ 98.6 $ &$ 97.3 $ &$ 95.1 $ &$ 93.5 $ &$ 53.4 $ &$ 52.2 $ &$ 41.8 $ &$ 39.7 $ &$ 46.7 $  \\
DFX & $ 82.4 $ &$ 0.0 $ &$ 0.0 $ &$ 0.0 $ &$ 0.0 $ &$ 98.8 $ &$ 96.8 $ &$ 92.4 $ &$ 86.2 $ &$ 84.9 $ & N/A & N/A & N/A & N/A & N/A  \\
DBX & $ 82.5 $ &$ 0.6 $ &$ 0.0 $ &$ 0.0 $ &$ 0.0 $ &$ 98.9 $ &$ 98.4 $ &$ 95.5 $ &$ 85.2 $ &$ 83.1 $ & N/A & N/A & N/A & N/A & N/A  \\
DFZ & $ 86.5 $ &$ 0.2 $ &$ 0.0 $ &$ 0.0 $ &$ 0.0 $ &$ 98.8 $ &$ 94.5 $ &$ 89.3 $ &$ 81.5 $ &$ 79.3 $ &$ 42.9 $ &$ 25.5 $ &$ 22.3 $ &$ 27.7 $ &$ 38.2 $  \\
\hline
\end{tabular}
\end{table}

% plane-vs-frog
\begin{table}
\caption{Black-box PGD attack results on CIFAR plane-vs-frog binary classification. }
\centering
\small
\setlength{\tabcolsep}{3pt}
\begin{tabular}{c|ccccc|ccccc|ccccc}\hline
 & \multicolumn{5}{c|}{substitute acc} & \multicolumn{5}{c|}{victim acc} & \multicolumn{5}{c|}{TP logit} \\
\hline
$\epsilon$ & 0.01 & 0.02 & 0.05 & 0.10 & 0.20 & 0.01 & 0.02 & 0.05 & 0.10 & 0.20 & 0.01 & 0.02 & 0.05 & 0.10 & 0.20 \\
\hline
GFZ & $ 95.5 $ &$ 89.9 $ &$ 45.5 $ &$ 6.8 $ &$ 0.1 $ &$ 97.6 $ &$ 95.6 $ &$ 88.7 $ &$ 68.8 $ &$ 29.7 $ &$ 6.2 $ &$ 6.7 $ &$ 6.0 $ &$ 10.8 $ &$ 71.8 $  \\
GBZ & $ 92.0 $ &$ 84.1 $ &$ 33.8 $ &$ 3.5 $ &$ 0.0 $ &$ 95.6 $ &$ 94.9 $ &$ 92.7 $ &$ 85.0 $ &$ 65.9 $ &$ 16.7 $ &$ 14.6 $ &$ 12.8 $ &$ 17.9 $ &$ 76.8 $  \\
GFY & $ 95.3 $ &$ 89.1 $ &$ 38.7 $ &$ 2.5 $ &$ 0.0 $ &$ 97.3 $ &$ 96.6 $ &$ 90.0 $ &$ 72.9 $ &$ 35.5 $ &$ 0.0 $ &$ 2.3 $ &$ 6.8 $ &$ 10.4 $ &$ 72.3 $  \\
GBY & $ 91.8 $ &$ 80.5 $ &$ 29.7 $ &$ 6.1 $ &$ 0.2 $ &$ 97.2 $ &$ 95.9 $ &$ 90.2 $ &$ 70.8 $ &$ 30.3 $ &$ 21.8 $ &$ 16.4 $ &$ 9.6 $ &$ 13.6 $ &$ 77.3 $  \\
DFX & $ 89.1 $ &$ 74.5 $ &$ 19.1 $ &$ 0.9 $ &$ 0.0 $ &$ 99.7 $ &$ 98.7 $ &$ 92.6 $ &$ 64.8 $ &$ 9.9 $ & N/A & N/A & N/A & N/A & N/A  \\
DBX & $ 85.5 $ &$ 76.4 $ &$ 42.5 $ &$ 4.5 $ &$ 0.0 $ &$ 99.6 $ &$ 99.0 $ &$ 94.8 $ &$ 77.5 $ &$ 24.5 $ & N/A & N/A & N/A & N/A & N/A  \\
DFZ & $ 85.5 $ &$ 73.5 $ &$ 21.1 $ &$ 1.5 $ &$ 0.0 $ &$ 99.5 $ &$ 99.2 $ &$ 93.8 $ &$ 70.1 $ &$ 15.1 $ &$ 0.0 $ &$ 5.0 $ &$ 8.9 $ &$ 15.9 $ &$ 85.1 $  \\
\hline
\end{tabular}
\end{table}

\begin{table}
\caption{Black-box MIM attack results on CIFAR plane-vs-frog binary classification. }
\centering
\small
\setlength{\tabcolsep}{3pt}
\begin{tabular}{c|ccccc|ccccc|ccccc}\hline
 & \multicolumn{5}{c|}{substitute acc} & \multicolumn{5}{c|}{victim acc} & \multicolumn{5}{c|}{TP logit} \\
\hline
$\epsilon$ & 0.01 & 0.02 & 0.05 & 0.10 & 0.20 & 0.01 & 0.02 & 0.05 & 0.10 & 0.20 & 0.01 & 0.02 & 0.05 & 0.10 & 0.20 \\
\hline
GFZ & $ 95.5 $ &$ 89.7 $ &$ 46.4 $ &$ 7.7 $ &$ 0.2 $ &$ 97.5 $ &$ 95.7 $ &$ 88.5 $ &$ 66.5 $ &$ 20.5 $ &$ 6.2 $ &$ 6.7 $ &$ 5.9 $ &$ 12.7 $ &$ 82.7 $  \\
GBZ & $ 91.8 $ &$ 84.0 $ &$ 33.1 $ &$ 3.4 $ &$ 0.0 $ &$ 95.6 $ &$ 94.9 $ &$ 92.3 $ &$ 85.1 $ &$ 60.7 $ &$ 16.7 $ &$ 15.5 $ &$ 14.2 $ &$ 21.6 $ &$ 97.9 $  \\
GFY & $ 95.3 $ &$ 88.9 $ &$ 40.5 $ &$ 3.1 $ &$ 0.0 $ &$ 97.3 $ &$ 96.7 $ &$ 89.5 $ &$ 70.6 $ &$ 31.9 $ &$ 0.0 $ &$ 2.4 $ &$ 7.6 $ &$ 12.5 $ &$ 88.0 $  \\
GBY & $ 91.7 $ &$ 80.2 $ &$ 29.5 $ &$ 5.9 $ &$ 0.1 $ &$ 97.3 $ &$ 95.9 $ &$ 89.6 $ &$ 69.0 $ &$ 26.1 $ &$ 22.2 $ &$ 16.1 $ &$ 9.7 $ &$ 14.0 $ &$ 95.9 $  \\
DFX & $ 89.0 $ &$ 74.5 $ &$ 19.4 $ &$ 0.9 $ &$ 0.0 $ &$ 99.6 $ &$ 98.7 $ &$ 92.1 $ &$ 63.0 $ &$ 10.9 $ & N/A & N/A & N/A & N/A & N/A  \\
DBX & $ 85.5 $ &$ 76.3 $ &$ 42.7 $ &$ 4.5 $ &$ 0.0 $ &$ 99.6 $ &$ 99.0 $ &$ 95.1 $ &$ 76.5 $ &$ 27.2 $ & N/A & N/A & N/A & N/A & N/A  \\
DFZ & $ 85.3 $ &$ 73.4 $ &$ 21.4 $ &$ 1.6 $ &$ 0.0 $ &$ 99.6 $ &$ 99.1 $ &$ 92.9 $ &$ 67.7 $ &$ 14.9 $ &$ 0.0 $ &$ 4.5 $ &$ 7.5 $ &$ 14.2 $ &$ 91.9 $  \\
\hline
\end{tabular}
\end{table}

\begin{table}
\caption{Black-box CW attack results on CIFAR plane-vs-frog binary classification. }
\centering
\small
\setlength{\tabcolsep}{3pt}
\begin{tabular}{c|ccccc|ccccc|ccccc}\hline
 & \multicolumn{5}{c|}{substitute acc} & \multicolumn{5}{c|}{victim acc} & \multicolumn{5}{c|}{TP logit} \\
\hline
$\epsilon$ & 0.10 & 1.00 & 10.00 & 100.00 & 1000.00 & 0.10 & 1.00 & 10.00 & 100.00 & 1000.00 & 0.10 & 1.00 & 10.00 & 100.00 & 1000.00 \\
\hline
GFZ & $ 96.3 $ &$ 41.5 $ &$ 0.5 $ &$ 0.0 $ &$ 0.0 $ &$ 98.7 $ &$ 94.7 $ &$ 93.4 $ &$ 93.3 $ &$ 92.3 $ &$ 13.3 $ &$ 8.6 $ &$ 7.7 $ &$ 7.7 $ &$ 8.9 $  \\
GBZ & $ 94.9 $ &$ 73.3 $ &$ 1.4 $ &$ 0.0 $ &$ 0.0 $ &$ 95.9 $ &$ 95.3 $ &$ 94.6 $ &$ 94.5 $ &$ 94.5 $ &$ 13.5 $ &$ 11.8 $ &$ 11.8 $ &$ 13.2 $ &$ 13.1 $  \\
GFY & $ 96.4 $ &$ 62.0 $ &$ 1.3 $ &$ 0.0 $ &$ 0.0 $ &$ 98.1 $ &$ 96.7 $ &$ 94.8 $ &$ 94.7 $ &$ 94.1 $ &$ 0.0 $ &$ 2.8 $ &$ 6.5 $ &$ 6.1 $ &$ 6.8 $  \\
GBY & $ 93.5 $ &$ 33.6 $ &$ 1.7 $ &$ 0.0 $ &$ 0.0 $ &$ 97.7 $ &$ 95.9 $ &$ 95.4 $ &$ 95.4 $ &$ 95.0 $ &$ 25.8 $ &$ 16.4 $ &$ 15.9 $ &$ 18.6 $ &$ 19.8 $  \\
DFX & $ 90.3 $ &$ 10.9 $ &$ 0.0 $ &$ 0.0 $ &$ 0.0 $ &$ 99.9 $ &$ 98.8 $ &$ 98.5 $ &$ 98.5 $ &$ 97.9 $ & N/A & N/A & N/A & N/A & N/A  \\
DBX & $ 87.2 $ &$ 31.8 $ &$ 0.0 $ &$ 0.0 $ &$ 0.0 $ &$ 99.9 $ &$ 97.1 $ &$ 94.5 $ &$ 93.9 $ &$ 94.7 $ & N/A & N/A & N/A & N/A & N/A  \\
DFZ & $ 84.9 $ &$ 22.7 $ &$ 0.0 $ &$ 0.0 $ &$ 0.0 $ &$ 99.9 $ &$ 99.0 $ &$ 98.3 $ &$ 98.3 $ &$ 97.8 $ &$ 0.0 $ &$ 3.8 $ &$ 2.3 $ &$ 2.3 $ &$ 11.0 $  \\
\hline
\end{tabular}
\end{table}

% cifar-10 VGG

\begin{table}
\caption{FGSM white-box attack results on CIFAR-10. }
\centering
\small
\setlength{\tabcolsep}{3pt}
\begin{tabular}{c|ccccc|ccccc|ccccc|ccccc}\hline
 & \multicolumn{5}{c|}{acc.~(adv)} & \multicolumn{5}{c|}{TP marginal} & \multicolumn{5}{c|}{TP logit} & \multicolumn{5}{c}{TP KL} \\
\hline
$\epsilon$ & 0.01 & 0.02 & 0.05 & 0.10 & 0.20 & 0.01 & 0.02 & 0.05 & 0.10 & 0.20 & 0.01 & 0.02 & 0.05 & 0.10 & 0.20 & 0.01 & 0.02 & 0.05 & 0.10 & 0.20 \\
\hline
VGG16 & $ 44.5 $ &$ 25.8 $ &$ 16.6 $ &$ 11.9 $ &$ 10.1 $ & N/A & N/A & N/A & N/A & N/A & N/A & N/A & N/A & N/A & N/A &$ 95.1 $ &$ 95.7 $ &$ 97.0 $ &$ 97.5 $ &$ 97.1 $  \\
GBZ-FC1 & $ 63.7 $ &$ 49.7 $ &$ 34.3 $ &$ 18.9 $ &$ 12.7 $ &$ 60.0 $ &$ 62.4 $ &$ 80.6 $ &$ 90.9 $ &$ 94.1 $ &$ 63.7 $ &$ 65.0 $ &$ 80.6 $ &$ 95.0 $ &$ 98.4 $ &$ 90.1 $ &$ 90.8 $ &$ 92.3 $ &$ 92.6 $ &$ 92.6 $  \\
GBY-FC1 & $ 64.9 $ &$ 50.0 $ &$ 36.1 $ &$ 19.5 $ &$ 11.3 $ &$ 52.2 $ &$ 59.0 $ &$ 80.4 $ &$ 90.1 $ &$ 92.5 $ &$ 54.6 $ &$ 61.9 $ &$ 80.3 $ &$ 95.6 $ &$ 98.4 $ &$ 90.7 $ &$ 91.3 $ &$ 92.1 $ &$ 92.4 $ &$ 90.3 $  \\
DBX-FC1 & $ 57.3 $ &$ 45.7 $ &$ 31.8 $ &$ 16.3 $ &$ 10.8 $ & N/A & N/A & N/A & N/A & N/A & N/A & N/A & N/A & N/A & N/A &$ 92.8 $ &$ 93.2 $ &$ 94.0 $ &$ 94.3 $ &$ 92.8 $  \\
GBZ-CONV9 & $ 81.6 $ &$ 74.1 $ &$ 58.5 $ &$ 48.9 $ &$ 45.8 $ &$ 13.7 $ &$ 17.0 $ &$ 28.4 $ &$ 76.0 $ &$ 90.2 $ &$ 18.3 $ &$ 20.6 $ &$ 31.7 $ &$ 64.3 $ &$ 71.6 $ &$ 91.7 $ &$ 91.5 $ &$ 91.7 $ &$ 93.3 $ &$ 86.4 $  \\
GBY-CONV9 & $ 75.8 $ &$ 66.0 $ &$ 39.7 $ &$ 22.5 $ &$ 17.5 $ &$ 14.0 $ &$ 18.2 $ &$ 29.4 $ &$ 74.4 $ &$ 88.9 $ &$ 17.7 $ &$ 20.8 $ &$ 25.8 $ &$ 55.2 $ &$ 68.9 $ &$ 89.6 $ &$ 89.9 $ &$ 90.0 $ &$ 89.1 $ &$ 89.2 $  \\
DBX-CONV9 & $ 59.0 $ &$ 45.7 $ &$ 31.8 $ &$ 17.8 $ &$ 11.3 $ & N/A & N/A & N/A & N/A & N/A & N/A & N/A & N/A & N/A & N/A &$ 93.7 $ &$ 94.1 $ &$ 94.9 $ &$ 95.2 $ &$ 93.7 $  \\
\hline
\end{tabular}
\end{table}

\begin{table}
\caption{PGD white-box attack results on CIFAR-10. }
\centering
\small
\setlength{\tabcolsep}{3pt}
\begin{tabular}{c|ccccc|ccccc|ccccc|ccccc}\hline
 & \multicolumn{5}{c|}{acc.~(adv)} & \multicolumn{5}{c|}{TP marginal} & \multicolumn{5}{c|}{TP logit} & \multicolumn{5}{c}{TP KL} \\
\hline
$\epsilon$ & 0.01 & 0.02 & 0.05 & 0.10 & 0.20 & 0.01 & 0.02 & 0.05 & 0.10 & 0.20 & 0.01 & 0.02 & 0.05 & 0.10 & 0.20 & 0.01 & 0.02 & 0.05 & 0.10 & 0.20 \\
\hline
VGG16 & $ 18.8 $ &$ 0.6 $ &$ 0.0 $ &$ 0.0 $ &$ 0.0 $ & N/A & N/A & N/A & N/A & N/A & N/A & N/A & N/A & N/A & N/A &$ 93.9 $ &$ 91.8 $ &$ 67.4 $ &$ 69.9 $ &$ 66.4 $  \\
GBZ-FC1 & $ 37.7 $ &$ 6.1 $ &$ 0.1 $ &$ 0.0 $ &$ 0.0 $ &$ 30.7 $ &$ 22.2 $ &$ 76.0 $ &$ 92.0 $ &$ 94.1 $ &$ 33.5 $ &$ 24.9 $ &$ 77.0 $ &$ 91.7 $ &$ 90.2 $ &$ 90.8 $ &$ 93.7 $ &$ 97.3 $ &$ 98.5 $ &$ 95.9 $  \\
GBY-FC1 & $ 31.2 $ &$ 8.5 $ &$ 2.6 $ &$ 1.5 $ &$ 0.6 $ &$ 32.0 $ &$ 20.3 $ &$ 82.3 $ &$ 95.1 $ &$ 97.6 $ &$ 36.6 $ &$ 29.5 $ &$ 84.1 $ &$ 93.7 $ &$ 90.4 $ &$ 90.3 $ &$ 94.1 $ &$ 97.7 $ &$ 97.3 $ &$ 96.9 $  \\
DBX-FC1 & $ 23.4 $ &$ 1.7 $ &$ 0.0 $ &$ 0.0 $ &$ 0.0 $ & N/A & N/A & N/A & N/A & N/A & N/A & N/A & N/A & N/A & N/A &$ 91.9 $ &$ 91.0 $ &$ 5.6 $ &$ 7.4 $ &$ 5.9 $  \\
GBZ-CONV9 & $ 61.1 $ &$ 38.3 $ &$ 24.2 $ &$ 10.2 $ &$ 2.3 $ &$ 19.2 $ &$ 60.3 $ &$ 96.7 $ &$ 98.9 $ &$ 99.6 $ &$ 20.1 $ &$ 60.3 $ &$ 96.8 $ &$ 98.3 $ &$ 98.7 $ &$ 90.3 $ &$ 93.1 $ &$ 97.7 $ &$ 98.7 $ &$ 99.6 $  \\
GBY-CONV9 & $ 66.5 $ &$ 26.6 $ &$ 3.2 $ &$ 0.4 $ &$ 0.0 $ &$ 17.0 $ &$ 55.6 $ &$ 95.7 $ &$ 99.2 $ &$ 99.6 $ &$ 17.8 $ &$ 52.7 $ &$ 95.3 $ &$ 98.6 $ &$ 97.6 $ &$ 89.6 $ &$ 91.8 $ &$ 98.7 $ &$ 99.4 $ &$ 96.9 $  \\
DBX-CONV9 & $ 24.0 $ &$ 3.0 $ &$ 0.3 $ &$ 0.1 $ &$ 0.0 $ & N/A & N/A & N/A & N/A & N/A & N/A & N/A & N/A & N/A & N/A &$ 91.3 $ &$ 93.1 $ &$ 99.3 $ &$ 99.6 $ &$ 99.2 $  \\
\hline
\end{tabular}
\end{table}

\begin{table}
\caption{MIM white-box attack results on CIFAR-10. }
\centering
\small
\setlength{\tabcolsep}{3pt}
\begin{tabular}{c|ccccc|ccccc|ccccc|ccccc}\hline
 & \multicolumn{5}{c|}{acc.~(adv)} & \multicolumn{5}{c|}{TP marginal} & \multicolumn{5}{c|}{TP logit} & \multicolumn{5}{c}{TP KL} \\
\hline
$\epsilon$ & 0.01 & 0.02 & 0.05 & 0.10 & 0.20 & 0.01 & 0.02 & 0.05 & 0.10 & 0.20 & 0.01 & 0.02 & 0.05 & 0.10 & 0.20 & 0.01 & 0.02 & 0.05 & 0.10 & 0.20 \\
\hline
VGG16 & $ 13.1 $ &$ 0.4 $ &$ 0.0 $ &$ 0.0 $ &$ 0.0 $ & N/A & N/A & N/A & N/A & N/A & N/A & N/A & N/A & N/A & N/A &$ 92.2 $ &$ 92.4 $ &$ 72.3 $ &$ 77.1 $ &$ 79.0 $  \\
GBZ-FC1 & $ 24.5 $ &$ 2.7 $ &$ 0.1 $ &$ 0.0 $ &$ 0.0 $ &$ 34.8 $ &$ 40.9 $ &$ 88.5 $ &$ 95.8 $ &$ 96.0 $ &$ 38.8 $ &$ 44.0 $ &$ 90.4 $ &$ 96.4 $ &$ 96.6 $ &$ 91.3 $ &$ 95.0 $ &$ 95.7 $ &$ 97.9 $ &$ 96.8 $  \\
GBY-FC1 & $ 33.0 $ &$ 19.6 $ &$ 11.8 $ &$ 7.2 $ &$ 4.9 $ &$ 29.1 $ &$ 38.7 $ &$ 95.7 $ &$ 99.1 $ &$ 99.3 $ &$ 38.4 $ &$ 49.2 $ &$ 96.3 $ &$ 99.3 $ &$ 99.4 $ &$ 90.7 $ &$ 95.0 $ &$ 97.8 $ &$ 97.9 $ &$ 97.9 $  \\
DBX-FC1 & $ 17.3 $ &$ 0.9 $ &$ 0.0 $ &$ 0.0 $ &$ 0.0 $ & N/A & N/A & N/A & N/A & N/A & N/A & N/A & N/A & N/A & N/A &$ 91.1 $ &$ 89.6 $ &$ 9.6 $ &$ 14.5 $ &$ 16.0 $  \\
GBZ-CONV9 & $ 57.6 $ &$ 43.5 $ &$ 29.1 $ &$ 24.1 $ &$ 22.6 $ &$ 27.4 $ &$ 78.5 $ &$ 99.4 $ &$ 99.6 $ &$ 99.7 $ &$ 29.1 $ &$ 78.3 $ &$ 99.6 $ &$ 99.7 $ &$ 99.8 $ &$ 90.2 $ &$ 94.4 $ &$ 97.5 $ &$ 98.0 $ &$ 98.2 $  \\
GBY-CONV9 & $ 39.1 $ &$ 18.1 $ &$ 4.3 $ &$ 1.9 $ &$ 1.5 $ &$ 25.6 $ &$ 70.4 $ &$ 98.6 $ &$ 99.6 $ &$ 99.8 $ &$ 28.2 $ &$ 71.1 $ &$ 98.6 $ &$ 99.7 $ &$ 99.8 $ &$ 89.6 $ &$ 94.5 $ &$ 99.4 $ &$ 99.8 $ &$ 99.7 $  \\
DBX-CONV9 & $ 19.2 $ &$ 3.2 $ &$ 0.5 $ &$ 0.4 $ &$ 0.4 $ & N/A & N/A & N/A & N/A & N/A & N/A & N/A & N/A & N/A & N/A &$ 92.1 $ &$ 96.7 $ &$ 99.9 $ &$ 100.0 $ &$ 100.0 $  \\
\hline
\end{tabular}
\end{table}

%
% tables for the bottleneck test
\begin{table}
\caption{FGSM white-box attack results on MNIST (with varied bottleneck layer sizes). }
\label{tab:mnist_bottleneck_fgsm}
\centering
\small
\setlength{\tabcolsep}{3pt}
\begin{tabular}{c|ccccc|ccccc|ccccc|ccccc}\hline
 & \multicolumn{5}{c|}{acc.~(adv)} & \multicolumn{5}{c|}{TP marginal} & \multicolumn{5}{c|}{TP logit} & \multicolumn{5}{c}{TP KL} \\
\hline
$\epsilon$ & 0.10 & 0.20 & 0.30 & 0.40 & 0.50 & 0.10 & 0.20 & 0.30 & 0.40 & 0.50 & 0.10 & 0.20 & 0.30 & 0.40 & 0.50 & 0.10 & 0.20 & 0.30 & 0.40 & 0.50 \\
\hline
DBX-16 & $ 92.6 $ &$ 85.6 $ &$ 76.4 $ &$ 64.7 $ &$ 52.3 $ & N/A & N/A & N/A & N/A & N/A & N/A & N/A & N/A & N/A & N/A &$ 90.3 $ &$ 92.4 $ &$ 92.3 $ &$ 92.7 $ &$ 94.2 $  \\
DBX-32 & $ 92.6 $ &$ 84.4 $ &$ 71.1 $ &$ 57.2 $ &$ 46.0 $ & N/A & N/A & N/A & N/A & N/A & N/A & N/A & N/A & N/A & N/A &$ 92.2 $ &$ 92.3 $ &$ 92.8 $ &$ 94.9 $ &$ 95.0 $  \\
DBX-64 & $ 91.6 $ &$ 77.8 $ &$ 58.1 $ &$ 44.5 $ &$ 36.0 $ & N/A & N/A & N/A & N/A & N/A & N/A & N/A & N/A & N/A & N/A &$ 92.6 $ &$ 93.6 $ &$ 95.2 $ &$ 96.3 $ &$ 97.0 $  \\
DBX-128 & $ 87.8 $ &$ 47.6 $ &$ 20.1 $ &$ 12.7 $ &$ 10.3 $ & N/A & N/A & N/A & N/A & N/A & N/A & N/A & N/A & N/A & N/A &$ 94.4 $ &$ 97.6 $ &$ 97.8 $ &$ 97.1 $ &$ 96.4 $  \\
\hline
GBZ-16 & $ 92.0 $ &$ 75.8 $ &$ 52.5 $ &$ 34.0 $ &$ 21.7 $ &$ 28.7 $ &$ 71.5 $ &$ 99.9 $ &$ 100.0 $ &$ 100.0 $ &$ 49.2 $ &$ 86.3 $ &$ 99.9 $ &$ 100.0 $ &$ 100.0 $ &$ 90.8 $ &$ 90.6 $ &$ 92.0 $ &$ 91.9 $ &$ 90.9 $  \\
GBZ-32 & $ 91.1 $ &$ 74.4 $ &$ 52.4 $ &$ 33.4 $ &$ 21.6 $ &$ 26.5 $ &$ 73.5 $ &$ 99.7 $ &$ 100.0 $ &$ 100.0 $ &$ 49.4 $ &$ 87.2 $ &$ 99.9 $ &$ 100.0 $ &$ 100.0 $ &$ 91.0 $ &$ 90.6 $ &$ 91.8 $ &$ 91.4 $ &$ 91.6 $  \\
\hline
\end{tabular}
\end{table}

\begin{table}
\caption{PGD white-box attack results on MNIST (with varied bottleneck layer sizes). }
\label{tab:mnist_bottleneck_pgd}
\centering
\small
\setlength{\tabcolsep}{3pt}
\begin{tabular}{c|ccccc|ccccc|ccccc|ccccc}\hline
 & \multicolumn{5}{c|}{acc.~(adv)} & \multicolumn{5}{c|}{TP marginal} & \multicolumn{5}{c|}{TP logit} & \multicolumn{5}{c}{TP KL} \\
\hline
$\epsilon$ & 0.10 & 0.20 & 0.30 & 0.40 & 0.50 & 0.10 & 0.20 & 0.30 & 0.40 & 0.50 & 0.10 & 0.20 & 0.30 & 0.40 & 0.50 & 0.10 & 0.20 & 0.30 & 0.40 & 0.50 \\
\hline
DBX-16 & $ 69.8 $ &$ 36.6 $ &$ 16.0 $ &$ 5.9 $ &$ 1.8 $ & N/A & N/A & N/A & N/A & N/A & N/A & N/A & N/A & N/A & N/A &$ 91.2 $ &$ 91.9 $ &$ 92.8 $ &$ 93.3 $ &$ 93.7 $  \\
DBX-32 & $ 63.5 $ &$ 26.8 $ &$ 12.5 $ &$ 6.2 $ &$ 2.7 $ & N/A & N/A & N/A & N/A & N/A & N/A & N/A & N/A & N/A & N/A &$ 91.5 $ &$ 92.1 $ &$ 92.7 $ &$ 92.9 $ &$ 93.5 $  \\
DBX-64 & $ 58.0 $ &$ 18.3 $ &$ 6.0 $ &$ 1.3 $ &$ 0.2 $ & N/A & N/A & N/A & N/A & N/A & N/A & N/A & N/A & N/A & N/A &$ 92.7 $ &$ 94.0 $ &$ 94.1 $ &$ 93.7 $ &$ 84.8 $  \\
DBX-128 & $ 42.3 $ &$ 3.0 $ &$ 1.1 $ &$ 0.3 $ &$ 0.2 $ & N/A & N/A & N/A & N/A & N/A & N/A & N/A & N/A & N/A & N/A &$ 97.1 $ &$ 97.1 $ &$ 98.4 $ &$ 98.4 $ &$ 98.6 $  \\
\hline
GBZ-16 & $ 88.3 $ &$ 62.8 $ &$ 37.4 $ &$ 23.1 $ &$ 15.5 $ &$ 27.1 $ &$ 35.8 $ &$ 71.9 $ &$ 98.9 $ &$ 100.0 $ &$ 46.3 $ &$ 53.0 $ &$ 82.3 $ &$ 99.3 $ &$ 100.0 $ &$ 89.6 $ &$ 91.7 $ &$ 92.3 $ &$ 91.4 $ &$ 91.8 $  \\
GBZ-32 & $ 87.1 $ &$ 61.9 $ &$ 40.2 $ &$ 28.0 $ &$ 20.4 $ &$ 24.0 $ &$ 39.1 $ &$ 76.7 $ &$ 98.9 $ &$ 100.0 $ &$ 41.5 $ &$ 56.4 $ &$ 85.4 $ &$ 99.7 $ &$ 100.0 $ &$ 91.3 $ &$ 91.0 $ &$ 91.1 $ &$ 91.3 $ &$ 91.1 $  \\
\hline
\end{tabular}
\end{table}

\begin{table}
\caption{MIM white-box attack results on MNIST (with varied bottleneck layer sizes). }
\label{tab:mnist_bottleneck_mim}
\centering
\small
\setlength{\tabcolsep}{3pt}
\begin{tabular}{c|ccccc|ccccc|ccccc|ccccc}\hline
 & \multicolumn{5}{c|}{acc.~(adv)} & \multicolumn{5}{c|}{TP marginal} & \multicolumn{5}{c|}{TP logit} & \multicolumn{5}{c}{TP KL} \\
\hline
$\epsilon$ & 0.10 & 0.20 & 0.30 & 0.40 & 0.50 & 0.10 & 0.20 & 0.30 & 0.40 & 0.50 & 0.10 & 0.20 & 0.30 & 0.40 & 0.50 & 0.10 & 0.20 & 0.30 & 0.40 & 0.50 \\
\hline
DBX-16 & $ 72.2 $ &$ 37.1 $ &$ 20.7 $ &$ 14.7 $ &$ 11.7 $ & N/A & N/A & N/A & N/A & N/A & N/A & N/A & N/A & N/A & N/A &$ 91.8 $ &$ 92.3 $ &$ 92.8 $ &$ 93.4 $ &$ 93.9 $  \\
DBX-32 & $ 66.7 $ &$ 27.6 $ &$ 16.8 $ &$ 12.5 $ &$ 10.6 $ & N/A & N/A & N/A & N/A & N/A & N/A & N/A & N/A & N/A & N/A &$ 91.8 $ &$ 92.5 $ &$ 92.4 $ &$ 93.2 $ &$ 93.5 $  \\
DBX-64 & $ 66.7 $ &$ 28.7 $ &$ 19.7 $ &$ 17.2 $ &$ 17.2 $ & N/A & N/A & N/A & N/A & N/A & N/A & N/A & N/A & N/A & N/A &$ 93.5 $ &$ 94.3 $ &$ 94.6 $ &$ 94.7 $ &$ 94.7 $  \\
DBX-128 & $ 42.3 $ &$ 1.3 $ &$ 0.2 $ &$ 0.1 $ &$ 0.0 $ & N/A & N/A & N/A & N/A & N/A & N/A & N/A & N/A & N/A & N/A &$ 97.2 $ &$ 97.6 $ &$ 98.6 $ &$ 98.9 $ &$ 99.1 $  \\
\hline
GBZ-16 & $ 83.8 $ &$ 36.4 $ &$ 8.9 $ &$ 2.3 $ &$ 0.9 $ &$ 26.9 $ &$ 79.8 $ &$ 99.7 $ &$ 100.0 $ &$ 100.0 $ &$ 43.3 $ &$ 88.0 $ &$ 99.8 $ &$ 100.0 $ &$ 100.0 $ &$ 91.6 $ &$ 92.1 $ &$ 91.2 $ &$ 91.6 $ &$ 92.6 $  \\
GBZ-32 & $ 82.7 $ &$ 35.5 $ &$ 9.5 $ &$ 3.0 $ &$ 1.1 $ &$ 26.6 $ &$ 83.2 $ &$ 99.7 $ &$ 100.0 $ &$ 100.0 $ &$ 45.8 $ &$ 90.0 $ &$ 99.8 $ &$ 100.0 $ &$ 100.0 $ &$ 90.3 $ &$ 91.7 $ &$ 91.4 $ &$ 92.5 $ &$ 92.6 $  \\
\hline
\end{tabular}
\end{table}

\end{landscape}

%%%%%%%%%%%%%%%%%%%%%%%%%%%%%%%%%%%%%%%%%%%%%%%%%%%%%%%%%%%%%%%%%%%%%%%%%%%%%%%
%%%%%%%%%%%%%%%%%%%%%%%%%%%%%%%%%%%%%%%%%%%%%%%%%%%%%%%%%%%%%%%%%%%%%%%%%%%%%%%

\end{document}